\def\ARXIV{}
\ifdefined\ARXIV
  \documentclass[doc,floatsintext,natbib]{apa7}
\else
  \documentclass[man,floatsintext,natbib]{apa7}
\fi

\ifdefined\ARXIV\fi

\usepackage{amsmath}

\usepackage{longtable}
\usepackage{array}

\newcommand{\valNHuman}{874{,}434}
\newcommand{\valNHumanRaw}{1{,}015{,}341}
\newcommand{\valNHumanRemovedPct}{13.9}
\newcommand{\valNPairs}{1{,}225}
\newcommand{\valNRev}{18}
\newcommand{\valRangeCosMin}{.394}
\newcommand{\valRangeCosMax}{.938}
\newcommand{\valRangeArMin}{-.777}
\newcommand{\valRangeSqMin}{-.267}
\newcommand{\valNliOpp}{-1.00}
\newcommand{\valArOpp}{-.63}
\newcommand{\valAtOpp}{.63}
\newcommand{\valEigFirst}{29.63}
\newcommand{\valEigFirstPct}{59}
\newcommand{\valPaKWord}{five}
\newcommand{\valKaiserKWord}{six}
\newcommand{\valTefiKWord}{two}
\newcommand{\valConsensusKWord}{five}
\newcommand{\valDepthOpt}{1{,}431}
\newcommand{\valDepthFull}{4{,}096}
\newcommand{\valDepthNmi}{.704}
\newcommand{\valDepthNDimWord}{five}
\newcommand{\valPaKSmallWord}{three}
\newcommand{\valPaKMediumWord}{four}
\newcommand{\valEgaKSmallWord}{six}
\newcommand{\valEgaKMediumWord}{six}
\newcommand{\valConsSmallWord}{six}
\newcommand{\valConsMediumWord}{six}
\newcommand{\valEkcKWord}{six}
\newcommand{\valEkcKLargeWord}{six}
\newcommand{\valMapK}{10}
\newcommand{\valCdHumanN}{2{,}500}
\newcommand{\valCdHumanImpFive}{57.2}
\newcommand{\valCdHumanImpSix}{19.7}
\newcommand{\valCdHumanNormFive}{16\%}
\newcommand{\valCdEmbNormFiveRange}{47--49\%}
\newcommand{\valCdEmbImpLow}{9}
\newcommand{\valCdEmbImpHigh}{22}
\newcommand{\valCdCrossTwoFiftyWord}{seven}
\newcommand{\valCdCrossSatWord}{ten}
\newcommand{\valCdEmbVerdictWord}{ten}
\newcommand{\valKmo}{.971}
\newcommand{\valTefiMain}{-27.71}
\newcommand{\valRmsr}{.032}
\newcommand{\valCaf}{.412}
\newcommand{\valPhiLo}{.40}
\newcommand{\valPhiHi}{.69}
\newcommand{\valCommMin}{.566}
\newcommand{\valCommMinItem}{A23}
\newcommand{\valCommMax}{.905}
\newcommand{\valCommMaxItem}{N16}
\newcommand{\valMrA}{MR2}
\newcommand{\valMrE}{MR5}
\newcommand{\valMrN}{MR3}
\newcommand{\valMrC}{MR1}
\newcommand{\valMrO}{MR4}
\newcommand{\valOmegaAsgnA}{.830}
\newcommand{\valOmegaTheoA}{.560}
\newcommand{\valOmegaAsgnE}{.709}
\newcommand{\valOmegaTheoE}{.583}
\newcommand{\valCalRmsrMin}{.0119}
\newcommand{\valCalRmsrMax}{.0134}
\newcommand{\valCalCafMin}{.501}
\newcommand{\valCalCafMax}{.532}
\newcommand{\valCalTefiMin}{-21.66}
\newcommand{\valCalTefiMax}{-17.82}
\newcommand{\valAnchorNTwOwn}{.69}
\newcommand{\valAnchorNTwRival}{.72}
\newcommand{\valAnchorATwThreeGap}{.006}
\newcommand{\valAnchorATwFiveOwn}{.36}
\newcommand{\valAnchorATwFiveNeur}{.37}
\newcommand{\valProjNTwCalm}{.82}
\newcommand{\valNmiTheory}{.527}
\newcommand{\valAriTheory}{.402}
\newcommand{\valUvaPairsWord}{eleven}
\newcommand{\valUvaClustersWord}{ten}
\newcommand{\valWtoTop}{.462}
\newcommand{\valCosPairsWord}{seven}
\newcommand{\valCosClustersWord}{five}
\newcommand{\valCosTopA}{.938}
\newcommand{\valCosTopB}{.933}
\newcommand{\valSfNmiFull}{.527}
\newcommand{\valSfNmiRed}{.810}
\newcommand{\valSfAriFull}{.402}
\newcommand{\valSfAriRed}{.571}
\newcommand{\valSfKRedWord}{three}
\newcommand{\valCosInUvaWord}{five}
\newcommand{\valSfRedKeptWord}{four}
\newcommand{\valIfGap}{.04}
\newcommand{\valIfNearest}{.95}
\newcommand{\valIfGoodGap}{.09}
\newcommand{\valHumanPaKWord}{ten}
\newcommand{\valPhiO}{.92}
\newcommand{\valPhiC}{.84}
\newcommand{\valPhiN}{.80}
\newcommand{\valPhiE}{.64}
\newcommand{\valPhiA}{.52}
\newcommand{\valPhiEA}{.52}
\newcommand{\valPhiMean}{.744}
\newcommand{\valFrob}{.664}
\newcommand{\valNliMinEig}{-14.1}
\newcommand{\valPairR}{.46}
\newcommand{\valMcpRRaw}{.456}
\newcommand{\valMcpRKeyed}{.434}
\newcommand{\valPairRAbs}{.53}
\newcommand{\valPairRsqLin}{.21}
\newcommand{\valPairRsqQuad}{.40}
\newcommand{\valPairNegN}{120}
\newcommand{\valPairNegOppN}{97}
\newcommand{\valPairSimNeg}{.58}
\newcommand{\valPairSimZero}{.56}
\newcommand{\valAtRKeyed}{.434}
\newcommand{\valArRKeyed}{.137}
\newcommand{\valArRRaw}{.444}
\newcommand{\valNliNmi}{.548}
\newcommand{\valNliRRaw}{.561}
\newcommand{\valNliRKeyed}{.063}
\newcommand{\valNliPhi}{.286}
\newcommand{\valSqRKeyed}{.567}
\newcommand{\valSqRRaw}{.490}
\newcommand{\valSqDis}{1.000}
\newcommand{\valSqKmo}{.022}
\newcommand{\valNmiGapNli}{.021}
\newcommand{\valSepSmall}{.398}
\newcommand{\valSepMedium}{.458}
\newcommand{\valSepLarge}{.416}
\newcommand{\valSepGapSmall}{.077}
\newcommand{\valSepGapMedium}{.077}
\newcommand{\valSepGapLarge}{.079}
\newcommand{\valSepCiSmallLo}{.286}
\newcommand{\valSepCiSmallHi}{.515}
\newcommand{\valSepCiMediumLo}{.372}
\newcommand{\valSepCiMediumHi}{.548}
\newcommand{\valSepCiLargeLo}{.316}
\newcommand{\valSepCiLargeHi}{.535}
\newcommand{\valDomGapNSmall}{.127}
\newcommand{\valDomGapNMedium}{.104}
\newcommand{\valDomGapNLarge}{.131}
\newcommand{\valDomGapOtherMax}{.077}
\newcommand{\valCalReps}{100}
\newcommand{\valCalRotWarnN}{20}
\newcommand{\valAnchorCloserN}{six}
\newcommand{\valAnchorOwnNameN}{29}
\newcommand{\valNmiNullMax}{.292}
\newcommand{\valAriNullMax}{.149}
\newcommand{\valJjAEDiv}{.17}
\newcommand{\valJjAEContent}{.89}
\newcommand{\valDepthFullNmi}{.454}
\newcommand{\valDepthFullNDimWord}{five}
\newcommand{\valNliNegEig}{18}
\newcommand{\valNliRepairR}{.93}
\newcommand{\valAsgnOSize}{seven}
\newcommand{\valAsgnOOwn}{seven}
\newcommand{\valAsgnCOwn}{ten}
\newcommand{\valAsgnCRev}{six}
\newcommand{\valAsgnESize}{seven}
\newcommand{\valAsgnASize}{ten}
\newcommand{\valAsgnARev}{ten}
\newcommand{\valAsgnNOwn}{eight}
\newcommand{\valAsgnADomainsWord}{four}
\newcommand{\valAsgnRevOff}{eight}
\newcommand{\valAsgnEOnBlend}{nine}
\newcommand{\valAsgnAOnBlend}{five}
\newcommand{\valAsgnAOnN}{four}
\newcommand{\valAsgnBlendOther}{three}
\newcommand{\valMsFloor}{0.3}
\newcommand{\valEncMaxDiff}{.0010}
\newcommand{\valEncR}{1.000}
\newcommand{\valPaThrSmall}{1.50}
\newcommand{\valPaThrMedium}{1.31}
\newcommand{\valPaThrLarge}{1.23}
\newcommand{\valPaThrEighthLarge}{1.14}
\newcommand{\valCalConvN}{99}
\newcommand{\valCalNonConvWord}{one}
\newcommand{\valCalRmsrMean}{.0127}
\newcommand{\valCalCafMean}{.507}
\newcommand{\valCalCafSd}{.003}
\newcommand{\valCalTefiMean}{-20.85}
\newcommand{\valCalTefiSd}{0.53}
\newcommand{\valCalTefiConvMin}{-21.66}
\newcommand{\valCalTefiConvMax}{-18.80}
\newcommand{\valCalCafConvMax}{.512}
\newcommand{\valPairRLo}{.34}
\newcommand{\valPairRHi}{.58}
\newcommand{\valPairSimPos}{.69}
\newcommand{\valPairSimNegOther}{.56}
\newcommand{\valPairNegOtherN}{23}
\newcommand{\valSfFiveNmi}{.810}
\newcommand{\valSfFiveAri}{.707}
\newcommand{\valSfRandN}{200}
\newcommand{\valSfRandNmiMed}{.600}
\newcommand{\valSfRandNmiHi}{.734}
\newcommand{\valSfRandNmiMax}{.785}
\newcommand{\valSfRandAriMed}{.367}
\newcommand{\valSfRandAriHi}{.559}
\newcommand{\valSfRandAriMax}{.656}
\newcommand{\valNliOrdersR}{.57}
\newcommand{\valNliSameKey}{.20}
\newcommand{\valNliOppKey}{-.79}
\newcommand{\valNliCrossNegPct}{71}
\newcommand{\valNliCrossStrongPct}{26}
\newcommand{\valArOppPct}{4}
\newcommand{\valArOppMean}{-.58}
\newcommand{\valCdRepLowMin}{seven}
\newcommand{\valCdRepLowMax}{nine}
\newcommand{\valCdRepMidMin}{eight}
\newcommand{\valCdRepMidMax}{ten}
\newcommand{\valDepthNEval}{148}
\newcommand{\valDepthStep}{28}
\newcommand{\valDepthNmiMin}{.193}
\newcommand{\valDepthNmiMax}{.704}
\newcommand{\valDepthMin}{3}
\newcommand{\valDepthGridN}{147}
\newcommand{\valDepthGridLast}{4{,}091}
\newcommand{\valCalWarnReps}{one}
\newcommand{\valCalWarnTotal}{20}
\newcommand{\valPairSameKeyN}{115}
\newcommand{\valPairSameKeySim}{.71}
\newcommand{\valPairSameKeyR}{.39}
\newcommand{\valPairOppKeyN}{110}
\newcommand{\valPairOppKeySim}{.58}
\newcommand{\valPairOppKeyR}{-.34}
\newcommand{\valPairOppKeyNegPct}{100}
\newcommand{\valPairCrossN}{1{,}000}
\newcommand{\valPairCrossSim}{.57}
\newcommand{\valPairCrossR}{.02}
\newcommand{\valPairGain}{.20}
\newcommand{\valPairGainLo}{.12}
\newcommand{\valPairGainHi}{.28}
\newcommand{\valPairGap}{.12}
\newcommand{\valPairGapLo}{.09}
\newcommand{\valPairGapHi}{.16}
\newcommand{\valPairOppCrossGap}{.016}
\newcommand{\valPairOppCrossLo}{-.005}
\newcommand{\valPairOppCrossHi}{.037}
\newcommand{\valPaSmallFourObs}{1.370}
\newcommand{\valPaSmallFourThr}{1.372}
\newcommand{\valPcaVarOne}{9}
\newcommand{\valPcaVarTwo}{9}
\newcommand{\valPurityTsne}{.72}
\newcommand{\valPurityUmap}{.68}
\newcommand{\valPurityLin}{.42}
\newcommand{\valPurityNTsne}{.80}
\newcommand{\valPurityNUmap}{.78}
\newcommand{\valPurityNLin}{.70}
\newcommand{\valPurityChance}{.18}
\newcommand{\valOmegaAsgnO}{.704}
\newcommand{\valOmegaTheoO}{.732}
\newcommand{\valOmegaAsgnN}{.908}
\newcommand{\valOmegaTheoN}{.902}
\newcommand{\valOmegaAsgnC}{.863}
\newcommand{\valOmegaTheoC}{.850}
\newcommand{\valAsgnNearTieN}{four}
\newcommand{\valLabelMarginA}{.045}
\newcommand{\valLabelMarginE}{.049}
\newcommand{\valPyPython}{3.11.5}
\newcommand{\valPyNumpy}{2.4.2}
\newcommand{\valPyTorch}{2.12.1}
\newcommand{\valPyTransformers}{5.12.1}
\newcommand{\valPySentTrans}{5.2.0}
\newcommand{\valNliRevision}{6c749ce3425cd33b46d187e45b92bbf96ee12ec7}
\newcommand{\valRevSmall}{97b0c614be4d77ee51c0cef4e5f07c00f9eb65b3}
\newcommand{\valRevMedium}{5cf2132abc99cad020ac570b19d031efec650f2b}
\newcommand{\valRevLarge}{1d8ad4ca9b3dd8059ad90a75d4983776a23d44af}
\newcommand{\valEmbCheckMinCos}{.999998}
\newcommand{\valDassRpbSmall}{.481}
\newcommand{\valDassGapSmall}{.099}
\newcommand{\valDassNmiSmall}{.412}
\newcommand{\valDassAriSmall}{.271}
\newcommand{\valBigNmiSmall}{.497}
\newcommand{\valBigAriSmall}{.393}
\newcommand{\valDassDepGapSmall}{.118}
\newcommand{\valDassAnxGapSmall}{.053}
\newcommand{\valDassStrGapSmall}{.126}
\newcommand{\valDassConsSmallWord}{five}
\newcommand{\valDassSmallAnxModal}{eight}
\newcommand{\valDassRpbMedium}{.505}
\newcommand{\valDassGapMedium}{.106}
\newcommand{\valDassNmiMedium}{.700}
\newcommand{\valDassAriMedium}{.650}
\newcommand{\valBigNmiMedium}{.693}
\newcommand{\valBigAriMedium}{.574}
\newcommand{\valDassDepGapMedium}{.129}
\newcommand{\valDassAnxGapMedium}{.044}
\newcommand{\valDassStrGapMedium}{.144}
\newcommand{\valDassConsMediumWord}{four}
\newcommand{\valDassRpbLarge}{.510}
\newcommand{\valDassGapLarge}{.124}
\newcommand{\valDassNmiLarge}{.834}
\newcommand{\valDassAriLarge}{.801}
\newcommand{\valBigNmiLarge}{.527}
\newcommand{\valBigAriLarge}{.402}
\newcommand{\valDassDepGapLarge}{.180}
\newcommand{\valDassAnxGapLarge}{.034}
\newcommand{\valDassStrGapLarge}{.158}
\newcommand{\valDassPaKLargeWord}{four}
\newcommand{\valDassConsLargeWord}{four}
\newcommand{\valDassLargeDepModal}{14}
\newcommand{\valDassLargeAnxModal}{11}
\newcommand{\valDassLargeStrModal}{14}
\newcommand{\valDassGapDiffLo}{-.001}
\newcommand{\valDassGapDiffHi}{.053}
\newcommand{\valBigGapDiffLo}{-.012}
\newcommand{\valBigGapDiffHi}{.013}
\newcommand{\valDassPermNmi}{.249}
\newcommand{\valDassPermAri}{.195}
\newcommand{\valDassFloor}{0.2}
\newcommand{\valDassNPairs}{861}
\newcommand{\valDassSmallStrOnAnx}{14}
\newcommand{\valDassSeedN}{40}
\newcommand{\valDassSmallSeedsRep}{16}
\newcommand{\valDassSmallSeedsAlt}{24}
\newcommand{\valDassCritSmall}{0.564}
\newcommand{\valDassCritAlt}{0.567}
\newcommand{\valDassAltNmi}{.665}
\newcommand{\valDassAltAri}{.514}
\newcommand{\valDassAltAnxModal}{14}
\newcommand{\valDassAltStrOnAnx}{13}
\newcommand{\valDassEmbCheckCos}{1.000000}

\title{semanticfa: An R Package for Response-Free Semantic Factor Analysis of Psychometric Scales}
\shorttitle{SEMANTIC FACTOR ANALYSIS IN R}

\authorsnames{Devon Yanitski, Chris Westbury}
\ifdefined\ARXIV
\authorsaffiliations{{Department of Psychology, University of Alberta, Edmonton, Alberta, Canada\\
  \href{mailto:dyanitsk@ualberta.ca}{dyanitsk@ualberta.ca}\\[1.2em]
  \today\\[0.6em]\mbox{}}}
\else
\authorsaffiliations{{Department of Psychology, University of Alberta, Edmonton, Alberta, Canada}}
\fi

\authornote{%
\addORCIDlink{Devon Yanitski}{0009-0006-1568-3387}

\addORCIDlink{Chris Westbury}{0000-0001-7788-0278}

This work received no external funding. Computing resources were provided
by the Digital Research Alliance of Canada. The authors have no competing
interests to declare.

Correspondence concerning this article should be addressed to Devon Yanitski,
Department of Psychology, University of Alberta, P-217 Biological Sciences
Building, Edmonton, Alberta T6G 2E9, Canada.

E-mail: dyanitsk@ualberta.ca}

\shortcites{bowman-etal-2015,grand-etal-2022,christensen-etal-2023,pellert-etal-2026,garrido-etal-2025,wang-etal-2026,guenole-etal-2025,%
arnulf-etal-2014,arnulf-etal-2018,arnulf-etal-2021,rosenbusch-etal-2020,nimon-etal-2016,%
borsboom-etal-2004,stanghellini-etal-2024,%
milano-etal-2025a,milano-etal-2025b,casella-etal-2024,schoenegger-etal-2025,ravenda-etal-2025,maharjan-etal-2025,lara-etal-1992,%
huang-etal-2025,kambeitz-etal-2025,kojima-etal-2026,russell-lasalandra-etal-2026,%
mikolov-etal-2013,pennington-etal-2014,vaswani-etal-2017,devlin-etal-2019,karimimahabadi-etal-2019,gunther-etal-2019,mandera-etal-2017,buchanan-etal-2001,kmetty-etal-2021,hussain-etal-2024,low-etal-preprint,golino-etal-2021,%
liu-etal-2025,lim-etal-2026,han-etal-2025,peters-etal-2025,bunt-etal-2025,eberhardt-etal-2025,suarez-alvarez-etal-2026,grobelny-etal-2025,goldberg-etal-2006,ushey-etal-2025,lorenzo-seva-etal-2011,massara-etal-2016,mcinnes-etal-2018,zhang-etal-2025,carlini-etal-2021,friedman-etal-2008,kusupati-etal-2022,he-etal-2023,williams-etal-2018,%
crede-etal-2017,harris-etal-2020,paszke-etal-2019,wolf-etal-2020,yang-etal-2025,zinbarg-etal-2005}

\abstract{%
The items of a reflective psychometric scale can be read as a proposal
about a construct's structure. Language models make that proposal
measurable by converting each item into an embedding, a list of numbers,
similar for similar meanings. We introduce \texttt{semanticfa}, an R
package for semantic factor analysis, which factors item-embedding
similarities. It also interprets factors, flags redundant items, builds short
forms, and screens new items. A worked example
analyzes the 50-item International Personality Item Pool Big-Five Factor
Markers, embedded with Qwen3-Embedding-8B, and compares the results with
a factor analysis of \valNHuman{} respondents. Retention criteria tied
between five and six factors, and the tie rule kept \valConsensusKWord.
Three corresponded mainly to single domains, and two blended most
Extraversion items with \valAsgnAOnBlend{} Agreeableness items. Each
factor's congruence with its best-matching human factor (1 = identical)
ranged from
\valPhiO{} for Openness to \valPhiA{} for Agreeableness. Item pairs more
similar in meaning tended to be answered more alike ($r = \valPairR$ with
raw responses), but same-domain reverse-keyed opposites, answered oppositely, were
about as similar as cross-domain pairs (\valPairOppKeySim{} vs.\
\valPairCrossSim). Sign-flipping reverse-keyed items' embeddings
did not improve agreement with raw responses and worsened it with
reverse-scored ones.
With five factors, a 25-item short form matched the domains better than
all \valSfRandN{} random five-per-domain sets, but parallel analysis kept only
\valSfKRedWord. We propose semantic--behavioral coherence, the agreement
between semantic and response structures, as a measurable property of a
scale, given an embedding model. Semantic factor analysis complements,
and does not replace, response-based factor analysis.
}

\newcommand{\tighttable}{\linespread{1}\small\selectfont}

\begin{document}
\maketitle


\ifdefined\ARXIV\section{Introduction}\fi

Much of what psychologists measure, from personality traits and
attitudes to symptoms of anxiety and depression, is measured through
language \citep{hommel-arslan-2025,kambeitz-etal-2025}. The instruments are scales: sets of short statements, called
items, that respondents rate for how well each one describes them, such
as ``I am the life of the party.'' for extraversion. Because the
attributes these scales target cannot be observed directly, psychologists
treat them as constructs. A construct is a theoretical attribute whose
standing is inferred from the pattern of people's answers rather than
observed. The standard tool for studying that pattern is factor
analysis. It identifies clusters of items whose responses correlate more
strongly with one another than with the rest of the scale, and it treats
each cluster as the expression of one underlying dimension, called a
factor. Each item's loading is a number that expresses how strongly the
item reflects a factor. A scale is considered well built when its items
load on the factors that correspond to the constructs they were written
to measure.

Factor analysis is almost always run on responses. Before anyone answers
a scale, however, its items already exist as sentences, and what those
sentences mean shapes much of the structure that factor analysis later
finds in the answers. The structure of a scale, in this sense, is the
number of factors it has and which items load on each. Items that
describe similar behavior in similar words tend to be answered alike, in
part because a respondent who endorses one statement tends to endorse
another that says much the same thing
\citep{arnulf-etal-2014,hommel-arslan-2025}. \emph{Semantic factor analysis}
(SFA) builds on that observation by estimating the structure of a scale
from its wording alone. The R package \texttt{semanticfa}
\citep{yanitski-westbury-2026} implements SFA, from
the item texts to the interpreted factors.

What makes the wording measurable is a kind of language model called a
sentence-embedding model. Trained on very large collections of text,
such a model learns to represent any sentence as a long list of numbers
(4,096 for the model used in this paper), which are the coordinates of a
point in a space with that many dimensions. That list is the sentence's
embedding. Training arranges the space so that sentences with similar meanings end
up close together, whatever their exact wording
\citep{reimers-gurevych-2019}. The
embeddings of ``I am the life of the party.'' and ``I talk to a lot of
different people at parties.'' therefore lie near each other, while the
embeddings of sentences on unrelated topics lie far apart. The
similarity of two items can then be expressed as the cosine similarity
of their embeddings. If each embedding is drawn as an arrow from the
origin of the space to its point, cosine similarity is the cosine of the
angle between the two arrows. It approaches 1 when two statements say
nearly the same thing and approaches 0 when their meanings are
unrelated. In practice, the embeddings from such models all share some similarity,
as the internal word representations of contextual language models
commonly do \citep{ethayarajh-2019}, so even unrelated statements score well above 0.
This point returns in the subsection on reverse-keyed items. Computing it for every pair of items yields a matrix with the
same form as the correlation matrix on which factor analysis normally
operates. SFA is factor analysis of that similarity matrix, so no
respondent is needed at any stage.

The \texttt{semanticfa} package organizes the analysis into four stages: building
the similarity matrix from item embeddings, deciding how many factors it
supports, extracting the factors and computing their diagnostics, and
interpreting each factor by relating the embeddings of its items to the
embeddings of candidate construct names. Beyond these stages, the
package offers refinement tools that use a fitted model to flag
redundant items, propose shorter forms, and screen newly written items
before they are administered.

SFA does not replace factor analysis of responses. It recovers a second
structure over the same items, the semantic structure, which is built
from what the items mean rather than from how people answer them. Where
the semantic structure and the structure found in responses agree, the
items that are similar in meaning are also the items that people answer
alike. Where the two diverge, the divergence identifies the items that
wording and responses group differently and can suggest why. We call the
degree of agreement between the two structures \emph{semantic--behavioral
coherence}, because the structure found in responses reflects how people
behave when they answer the items. In the worked example, that
behavioral structure, which we also call the response structure, is the
factor structure of the human responses, and the Results call its
factors the human factors.

The Introduction develops these ideas in six subsections. It begins with
the evidence that item wording predicts how people respond, and then
explains how language models turn the meaning of a sentence into
numbers, the step on which both that evidence and SFA depend. Once
embeddings have been explained, the third subsection considers what the
semantic structure of a scale represents and how it relates to the
structure found in responses. The fourth addresses the first practical
difficulty any user meets, which concerns items worded so that agreeing
with them indicates less of the construct, because how an analysis
handles these items changes its results. The last two subsections turn
from the method to its uses. The fifth describes what semantic structure
can do before any data are collected, and the sixth introduces the
package together with the worked example that the rest of the paper
follows.

\subsection{Item Wording Predicts How People Respond}

This subsection reviews the evidence that a scale's wording predicts how
people respond to it. The evidence runs from single pairs of items to
whole scales and then to its limits: how well wording predicts the
correlation between two items, how well grouping items by meaning
recovers a scale's intended factors, and what the disagreements between
semantic and response-based results reveal.

The wording of items predicts, in part, how responses to them correlate
\citep{arnulf-etal-2014,arnulf-etal-2018,gefen-larsen-2017,hommel-arslan-2025,hussain-etal-2024,kambeitz-etal-2025,nimon-etal-2016,ravenda-etal-2025,schoenegger-etal-2025,wulff-mata-2025},
and in the strongest studies it predicts a substantial share
\citep{arnulf-etal-2014,hommel-arslan-2025}.
For example, \citet{arnulf-etal-2014} measured the semantic similarity of
item pairs in organizational-behavior surveys with methods such as
latent semantic analysis. Latent semantic analysis is an early technique
that represents each word by the texts in which it occurs and then
reduces these representations to a few hundred dimensions
\citep{landauer-dumais-1997}. Across three samples, semantic similarity explained 47\% to 79\% of the
variance in the correlations observed in responses (the in-sample $R^2$ of
regressions over item pairs), and 68\% to 87\% once the scale each item
belonged to was also taken into account. \citet{arnulf-etal-2014} explained
the result with what they called the semantic theory of survey response,
which holds that the relations among survey items are largely determined
by the items' semantic properties. The program of work that followed is
summarized in an editorial by \citet{arnulf-etal-2021}. The early methods in this program represented words rather than whole
sentences, and on one kind of scale they predicted almost nothing. Item
meaning predicted almost nothing about responses to a five-factor
personality inventory \citep{arnulf-etal-2014}, and a reanalysis of the
same data agreed \citep{arnulf-etal-2018}. The authors tentatively
attributed this to how personality inventories are written. They
suggested that the items of inventories such as the one they studied
sample a wide range of past behaviors, so the answers carry information
that the wording does not supply \citep{arnulf-etal-2018}.
The worked example in this paper also uses a five-factor personality
inventory, and the studies with sentence embeddings described below have
since recovered structure in such inventories. Separately, \citet{hommel-arslan-2025} trained a sentence-embedding model
to predict item correlations and tested it on new items in a preregistered
study. After adjustment for sampling
error, its predictions correlated $r = .59$ with the observed
correlations between single items and $r = .84$ with the correlations
between whole scales, so the model predicted how whole scales relate more
accurately than how single items relate. For the scale-level
predictions, the embeddings of items worded against the construct were
multiplied by $-1$ before the embeddings of each scale's items were
averaged, which requires knowing in advance which items are worded that
way.

Wording also predicts some properties of single items. In educational
testing, the difficulty of an item can be predicted in part from its text
\citep{peters-etal-2025}. Difficulty is commonly measured either as the proportion
of people who answer the item correctly or as a corresponding parameter
of a statistical model of the responses. Item
discrimination refers to how sharply an item separates people who are
high on the construct from people who are low on it. In one large bank
of English language arts items, discrimination was harder to predict
from text alone \citep{han-etal-2025}, although language models
fine-tuned to answer the items as simulated respondents recovered it
better on the items of a Grade 6 English language arts assessment
\citep{ormerod-2026}.

A stronger version of the finding concerns whole factor structures.
Grouping items, or facets of items, by the similarity of their meaning,
whether by principal component analysis, factor analysis, or clustering,
largely recovers the factors that established scales were designed to
measure
\citep{casella-etal-2024,guenole-etal-2025,kambeitz-etal-2025,milano-etal-2025a}.
For one clinical scale, a proof of concept found the same factor
structure in the semantic similarities as in the response correlations
\citep{ravenda-etal-2025}.
\citet{guenole-etal-2025} formalized the approach as \emph{pseudo-factor
analysis}, in which the matrix of cosine similarities between embeddings is
factor-analyzed as a correlation matrix would be. The embeddings
were those of the items, pooled over the items of each facet, or of each
facet's items embedded together as one text. Analyzed at the level of their
facets, the five- and six-factor structures of two large
personality inventories from the International Personality Item Pool
(IPIP), the IPIP-NEO-300 and the IPIP version of the HEXACO-240, were
largely recovered, with the theoretically expected number of factors and
an oblique target rotation toward the theoretical structure. The
loadings themselves correlated substantially with those from responses
but were often not equivalent to them, with Tucker congruence often
below .85.
\citet{milano-etal-2025a} found the same pattern across four scales,
including a set of Big Five markers and the Depression Anxiety Stress
Scales. In each scale, more than 70\% of a construct's items, on average,
had their highest loading on the same principal component, and the
items' semantic loadings correlated with the loadings estimated from
responses. The finding extends, in part, to clinical symptom scales
\citep{kambeitz-etal-2025}. A cognitive-network analysis of the
Depression Anxiety Stress Scales found a related overlap: communities of
the items' words overlapped with the factors found in responses more
than random allocations of the same words would
\citep{stanghellini-etal-2024}.

The match is strong but not complete. Studies that compared semantic
structures with structures derived from responses or surveys have
located specific places where the two diverge \citep{feraco-toffalini-2025,kmetty-etal-2021,kojima-etal-2026,milano-etal-2025a}.
\citet{feraco-toffalini-2025} proposed fitting confirmatory factor models
to a matrix of cosine similarities between item embeddings, so that a
model can be evaluated before any data are collected. They fitted such
models both to the similarities among the items of a character-strengths
inventory and to the responses of 31,697 people. A confirmatory factor model specifies in
advance which items load on which factors and tests whether that
structure fits the data. The semantic similarities correlated only
moderately with the response correlations ($r = .67$). Every model that
fit the semantic matrix also fit the responses, but about half of the
models that the semantic matrix rejected fit the responses as well. In that study, then, a model that fit the semantic matrix also fit the
responses, whereas a model that failed to fit the semantic matrix often
fit the responses anyway. In one unpublished study of a single sample and nine dimensions from
four scales, the strength of the relation between wording and responses
also appeared to vary across dimensions \citep{zhang-qiu-2026}. \citet{kojima-etal-2026} extracted factors
from both embedding similarities and response correlations for 145
psychiatric symptom items in Japanese. Both analyses produced a general factor on which all the items loaded. This factor was nearly identical in the
two (congruence $= .97$), whereas the narrower factors diverged.
Congruence measures how similar the loadings of two factors are. A value of 1 means that the two factors have the same pattern of loadings.
Insomnia and physical symptoms, for example, formed one factor in the
responses but not in the wording, plausibly because they occur together
in people, perhaps through bodily or physiological causes, without being
described alike. Semantic analyses of personality items also find agreeableness and
extraversion items overlapping in meaning
\citep{casella-etal-2024,milano-etal-2025a}. One possible reason is that
items of both domains describe warmth and sociability. Both studies
note, however, that the two constructs are also associated in human
responses \citep{casella-etal-2024,milano-etal-2025a}. Such divergences are
informative, because they can mark the places where the way a construct
is described differs from the way people experience it. For that reason, this paper presents SFA as a
complement to response data rather than a replacement.

Together, these studies suggest that much of a scale's structure is already
present in its wording. All of them depend on a step that the next
subsection explains, turning the meaning of a sentence into numbers.

\subsection{Language Models Turn Item Meaning Into Numbers}

This subsection explains how the meaning of a sentence is turned into
numbers, in four steps that build on one another: the distributional
hypothesis about meaning, the word vectors that first applied it, the
sentence embeddings that SFA uses, and the cosine similarity that
compares them.

The underlying idea is old. \citet[\S43]{wittgenstein-1953} held that,
for a large class of cases, ``the meaning of a word is its use in the
language.'' In the same years, linguists stated a related claim, now called the
distributional hypothesis. The hypothesis holds
that words that occur in similar contexts tend to have similar meanings
and, conversely, that words that differ more in meaning tend to occur in
more different contexts \citep{harris-1954,joos-1950}.

Computational models of meaning apply this idea by learning, for each
word, a vector of numbers from the contexts in which the word appears
across a large body of text. \citet{mikolov-etal-2013} trained a small
neural network to predict which words appear near each other and used
the network's internal weights as the word vectors, whereas
\citet{pennington-etal-2014} built vectors directly from counts of how
often each pair of words occurs near each other. In both kinds of space,
words with similar meanings receive similar vectors. Differences between vectors often capture relations between words as
well. For example, the vector
for \emph{king}, minus the vector for \emph{man}, plus the vector for
\emph{woman}, lies close to the vector for \emph{queen}, because the
difference between \emph{king} and \emph{man} is roughly the same as the
difference between \emph{queen} and \emph{woman}
\citep{mikolov-yih-zweig-2013}.

Although such spaces are built only from where words occur, distances
in them predict how people judge and process words
\citep{buchanan-etal-2001,gunther-etal-2019}.
Similarities between word vectors predict how similar people judge two
words to be, nearly as well as people agree with one another
\citep{gunther-etal-2019}, and the density of a word's neighborhood in such a space predicts how
quickly people recognize it \citep{buchanan-etal-2001}. The spaces even preserve
graded features of meaning, such as size. \citet{grand-etal-2022} located each word's vector along a line running
between words of opposite meaning, such as \emph{small} and
\emph{large}, and the order of the words along that line tracked
people's ratings of everyday words on many, though not all, such
features. How closely the order tracked the ratings varied widely from
one combination of word category and feature to another.

Scale items are sentences rather than single words, so their meaning
depends on how their words combine. Modern language models represent
this combination with the transformer architecture, in which the
representation of each word can draw on the other words in its sentence
\citep{devlin-etal-2019,vaswani-etal-2017}. A simple way to obtain one vector per sentence is to average the
vectors of its words. Averaged in this way, the vectors from the
transformer model BERT agreed less with human judgments of sentence
similarity than averaged vectors from an older word-level method, GloVe
\citep{reimers-gurevych-2019}.
\citet{reimers-gurevych-2019} addressed the problem by training the
model on pairs of sentences, so that sentences with similar meanings
receive similar vectors. The sentence-embedding models described at
the start of the paper are likewise trained on pairs of texts, although
many differ from this model in architecture and training objective. Their output for a sentence is the embedding that SFA uses. The embeddings in
this paper come from one such model, Qwen3-Embedding-8B, which is built
on a Qwen3 large language model \citep{qwen-2025,zhang-etal-2025}.

SFA compares embeddings with the cosine similarity introduced at the
start of the paper. The embeddings of two items with similar meanings,
such as ``I am the life of the party.'' and ``I talk to a lot of
different people at parties.'', point in similar directions and have a
high cosine similarity. The embeddings of items with less related
meanings point in more different directions and have a lower one.
Embedding every item and computing the cosine similarity of every pair
therefore turns the wording of a scale into a matrix that factor
analysis can use. What the factors of that matrix represent is the
question the next subsection takes up.

\subsection{A Scale Has Both a Semantic and a Behavioral Structure}

The evidence reviewed so far shows that wording predicts responses, but
it does not say what the semantic structure of a scale is evidence of.
This subsection argues that the semantic structure is part of what
defines a construct, and that its agreement with the structure found in
responses is evidence about that construct. The argument has three
steps. First, a construct is defined by its relations to behavior and to
other constructs, so the relations among the meanings of its items are
part of what defines it. Second, a scale's items are written
deliberately to sample the construct, so their meanings are a designed
proposal about its structure. Third, the same factor-analytic procedure
can be run on either the meanings or the responses, so the agreement
between the two results is evidence about the construct. This subsection then states what that agreement can and cannot show and closes with two
consequences of the argument, one for naming factors and one for a
current critique of rating scales.

The first step concerns what a construct is. \citet{cronbach-meehl-1955}
argued that a construct is defined not by any single operation or criterion but by
its place in a \emph{nomological network}, the set of lawful relations that
connect a construct to behavior, to observations, to theory, and to
other constructs. Two psychological scales therefore measure the same
thing when they relate in the same way to the same behaviors,
observations, and other constructs. The same relational view holds for
word meaning, as the previous subsection showed. Distributional theory
defines the meaning of a word by the contexts in which it occurs, and
nomological theory defines the meaning of a construct by its lawful
relations, so a construct stands to its nomological network as a word
stands to its distribution. Scale items fall under both theories at
once, because each item is both a piece of language and an indicator of
a construct. Validity theory applies the relational definition to
test scores. What a score means is established in large part by its relations to other
measures
\citep{cronbach-meehl-1955,messick-1995}. Convergent relations are those
with measures of the same construct made by independent methods, and discriminant relations are those with measures of different constructs
\citep{campbell-fiske-1959}. \citet{messick-1995}
distinguishes six aspects of validity, two of which bear on SFA. The content aspect concerns whether the items are relevant to, and
representative of, the construct's domain, and the structural aspect
concerns whether the way the items are scored mirrors the structure of
that domain. SFA bears on the first and, in part, on the second, because it examines
what the items say and how their meanings group.

The second step concerns how items are written. Scale developers do not
choose items at random. They write an initial pool that deliberately
covers every kind of content that might belong to the construct, and
they keep the items that perform best when the pool is tried out on
respondents \citep{clark-watson-2019}. The wording they choose has
consequences of its own. \citet{clark-watson-2019} note, for example,
that responses to an item containing almost any negative mood term will
partly reflect neuroticism, so the exact phrasing of an item can change
what it measures. A scale's items are therefore a \emph{semantic proposal}, a
deliberate attempt to set out in ordinary language what a construct
contains and how its parts relate, and SFA measures the structure of
that proposal. This view assumes a particular measurement model, which
states how a construct relates to the responses to its items. In a
reflective model, the construct causes the responses to its items, so
the items should correlate with one another and factor analysis is
appropriate. This expectation that the items of a construct correlate is
the dominant view in psychology \citep{bollen-lennox-1991}. In a formative model, by contrast, the
items jointly determine the construct and need not correlate at all
\citep{bollen-lennox-1991}. SFA is designed for
reflective scales, which include most personality and attitude scales.
For formative item sets, the measurement model does not constrain how
responses correlate \citep{bollen-lennox-1991}, so agreement between the
semantic and response structures cannot be assumed, and weak agreement
need not be a sign that something is wrong.

The third step is that one procedure recovers both structures. Factor
analysis starts from a matrix of associations between every pair of
items and finds the few dimensions that best account for it. When the
associations are correlations between responses, the result is the
scale's behavioral structure. When the associations are similarities
between item meanings, the result is its semantic structure. Both uses
have roots in the beginnings of personality measurement. \citet{galton-1884}
proposed measuring character by systematically recording what people do,
and he estimated that English has at least a thousand words that
describe character, observing that each has its own
shade of meaning while sharing much of it with others. Twenty years
later, \citet{spearman-1904} supplied the inference on which factor
analysis rests, that correlated measurements reflect a common underlying
factor. Later lexical studies, which
\citet{goldberg-1990} reviews, joined the two lines by factoring people's
ratings on sets of trait adjectives, and Goldberg showed, with the most
comprehensive pools of English trait adjectives, that this recovers the
Big Five. Because only the source of the matrix changes,
the rest of the factor-analytic toolkit carries over, with some tools
adapted to embeddings. This includes deciding how many factors to keep
\citep{horn-1965,kaiser-1960}, rotating the factors to make them easier to interpret, and testing whether a hypothesized
structure reproduces the observed associations \citep{joreskog-1969}.
Rotation re-expresses a factor solution so that each item loads strongly
on as few factors as possible, without changing how well the solution
fits. A rotation is either orthogonal or oblique
\citep{bernaards-jennrich-2005}. An orthogonal rotation, such as varimax
\citep{kaiser-1958}, keeps the factors uncorrelated, and an oblique
rotation lets them correlate. Testing a hypothesized structure carries
over only in part. The structure can be fitted to a similarity matrix,
but many tests of its fit cannot be computed, because they require a
sample size, which a similarity matrix lacks
\citep{suarez-alvarez-etal-2026}.

The agreement between the two structures is the semantic--behavioral
coherence introduced at the start of the paper, and agreement and
disagreement each have a specific interpretation. When the two
structures agree, the items that are similar in meaning are also the
items that many people answer alike, so two very different kinds of
evidence group the items into the same factors. In the terms of
\citet{cronbach-meehl-1955}, this match can be read as one of the lawful relations that make up the
construct's nomological network. When the
two structures disagree, the disagreement identifies the items that
wording and responses group differently and raises the question of why.
A researcher can then examine three explanations: (1) the items may be
worded ambiguously, (2) people may respond to them for reasons their
wording does not state, or (3) the construct may be organized
differently in behavior than in language.

Coherence is evidence about a construct, not proof that it exists.
\citet{borsboom-etal-2004} argue that a test is valid for an attribute
only if the attribute exists and causally produces the test's outcomes.
They also reject the view, adopted in the first step above, that a
construct is defined by its nomological network. On their account, a match between a test's relations and a theory
corroborates the theory, and a mismatch signals a problem, but the match
is not what makes the test valid. A close
semantic--behavioral match therefore supports, but cannot establish, the
claim that a scale measures what it was written to measure.

Coherence would be expected to be lower, moreover, where items influence one another
directly instead of through a common cause. \citet{borsboom-2008} proposes that some psychiatric symptoms relate in
this way. In his example, a panic
attack leads to worry about further attacks, and the worry in turn
leads to avoidance. The symptoms would then correlate because each causes the next,
not because their descriptions share meaning, so their response
structure need not match their semantic structure.

Embeddings can also help with the step of factor analysis that has
depended most on the analyst's judgment, naming the factors.
\citet{lara-etal-1992} identified factor interpretation as a major
problem in factor analysis, because many rotated solutions fit the data
equally well and the analyst chooses among the meanings they suggest
without a formal check. They proposed stating a candidate meaning for each factor as a
hypothesis and testing whether it is compatible with the variables that
load on that factor, and they anticipated that automatic interpretation
of factors might become necessary for very large problems. SFA can run an analogous test,
because a candidate name for a factor, such as \emph{extraversion}, can
be embedded in the same space as the items. The cosine similarity
between the embedding of the name and the embeddings of the items that
load on the factor then measures how well the name fits.

One recent critique of rating scales motivates studying their semantic
structure separately. \citet{uher-2025} argues that rating-scale data are people's
judgments of verbal statements rather than measurements of the phenomena
the statements describe, and calls the confusion of the two psychology's
\emph{cardinal error}. If this is right, much of the correlation among
ratings is fixed by what the statements mean before anyone answers them,
and the semantic structure of the items is a real object that can be
studied separately. Uher also argues that language-model methods cannot
by themselves show how responses relate to the phenomena they describe. SFA estimates that structure from the wording alone,
before any data exist, and comparing it with the structure found in
responses shows how closely the responses follow the wording. Every
analysis of this kind, however, has difficulty with items worded
against the construct, and the next subsection explains why.

\subsection{Embeddings Do Not Represent Reverse-Keyed Items as Opposites}

Most scales mix two kinds of items. A forward-keyed item is worded so
that agreeing with it indicates more of the construct, as ``I am the life
of the party.'' does for extraversion. A reverse-keyed item is worded so
that agreeing with it indicates less of the construct, as ``I don't talk
a lot.'' does. Before responses are analyzed, the answers to
reverse-keyed items are reversed, so that a high score always means more
of the construct. In the raw responses, the two kinds of items correlate
negatively, because people who agree with one tend to disagree with the
other. This subsection explains why embeddings miss that negative
relation and describes three families of solutions that have been
proposed.

The cosine similarity between two sentence embeddings almost never falls
below zero. Sentence embeddings share many features of language use that
are common to nearly all sentences, and these shared features keep every
pair of embeddings pointing in broadly the same direction
\citep{hommel-arslan-2025}. In one of their examples, a widely used
sentence-embedding model rated ``I am the life of the party.'' about as
similar to its opposite, ``I keep in the background.'' (cosine $= .35$),
as to the compatible ``I make friends easily.'' (cosine $= .32$).
Factoring such a matrix does not yet separate reverse-keyed items from their
forward-keyed opposites \citep{suarez-alvarez-etal-2026}, and the
missing negative correlations are a major obstacle to reproducing the
correlation patterns of human data
\citep{hommel-arslan-2025,pellert-etal-2026}. A loosely related observation appears in an early attempt to formalize factor
interpretation. \citet{lara-etal-1992} noted that properties studied in
sociological domains cannot be tied exactly to continuous scales, so
their meanings ``are not unequivocally distinguishable from their
opposites'' (p.~14). Their remark concerns the blurred boundaries of such
concepts rather than how embeddings represent reverse-keyed items, so
the parallel is only an analogy.

The proposed solutions differ in where they get the negative sign. The
first family takes it from the scoring key, which records which items
are reverse-keyed. \citet{guenole-etal-2025} multiplied each
reverse-keyed item's embedding by $-1$ before computing cosines, as an
embedding analogue of reversing a response. This requires knowing the
key in advance, and it also rests on an assumption that the embedding
space does not meet. Reversing a response keeps the item's content and
inverts only the scale on which it is answered. Multiplying an embedding
by $-1$ instead flips every one of its numbers, including the many that
encode features all sentences share, so the result points away from the
item's whole topic rather than toward its opposite claim. The space is not organized so that negating a sentence corresponds to
inverting its vector. \citet{guenole-etal-2025} applied the flip before
averaging the item embeddings into one embedding per facet. With
that design, sign-flipping raised factor recovery for the IPIP-NEO-300.
For the HEXACO, however, it recovered the factors less often than the
other encodings tested and consistently failed to recover the
Emotionality factor.

The second family needs no scoring key and works from the embeddings
alone. SQuID (Survey and Questionnaire Item Embeddings Differentials)
takes the sign from the other items in the scale. It subtracts the
average of all the item embeddings, called
the questionnaire embedding, from every item's embedding
\citep{pellert-etal-2026}. Because this removes what all the items
share, the similarities that remain can fall below zero. The authors of
the method computed these similarities as correlations, whereas the
package computes them as cosines. They evaluated the method mainly on a
questionnaire of human values, and they noted that questionnaires with
reverse-keyed items still require methodological refinement.
Mean-centering, a second correction that needs no scoring key, subtracts
each embedding's own mean value from all of its entries. After this
centering, the cosine between two embeddings equals their Pearson
correlation \citep{pokropek-2026}, so a constant offset shared by all of
an embedding's entries no longer adds to the similarity. On the
embeddings of the worked example this made no practical difference, and
mean-centering does not by itself give opposite items negative
similarities. It is nevertheless the package's default,
for reasons given in the Method section.

The third family takes the sign from a model trained to recognize when
two sentences contradict each other. Natural language inference (NLI) is
the task of judging whether one sentence entails, contradicts, or is
neutral toward another, where one sentence entails another when the
truth of the first implies the truth of the second.
\citet{hommel-arslan-2025} retrained a sentence-embedding model on a
corpus of about 570,000 sentence pairs that people had labeled in this
way \citep{bowman-etal-2015}. Contradicting pairs were given negative
target similarities, so the model learned to place sentences that
contradict each other far apart. Their final model, which was also trained further on correlations
between survey items, still predicted positive correlations for more
item pairs than the response data showed, 67\% against 59\%
\citep{hommel-arslan-2025}. A cross-encoder
trained for NLI departs from embeddings entirely, because it produces no
item vectors at all. Instead, it reads the two items of a pair together
and estimates two probabilities, that the first item entails the second
and that it contradicts the second. Subtracting the probability of
contradiction from the probability of entailment gives a similarity
between $-1$ and 1 that is negative when the two items assert opposite
things \citep[cf.][]{hommel-arslan-2025}.

A different workaround, outside these three families, changes the items
rather than the similarities. \citet{jung-seo-2025} reworded each
reverse-keyed item into positive form before embedding it, so that ``I
don't talk a lot.'' became ``I talk a lot.'', and they noted that the
rewording may have changed what the items mean. The \texttt{semanticfa}
package implements at least one method from each family:
sign-flipping from the first, SQuID and mean-centering from the second,
and the NLI cross-encoder from the third. The worked example compares
them against human responses, so that a user can see what each choice
does to the results.

\subsection{Semantic Structure Can Guide Scale Development Before Any Data Are Collected}

Because SFA needs no respondents, it can be used at the stage of scale
development where changes are cheapest, before any data are collected.
Response data are costly to gather and have also become easier to
corrupt. \citet{westwood-2025} built an autonomous software agent that
passed 99.8\% of the attention checks used to screen online survey respondents. An analysis of item wording cannot be corrupted in this
way because it uses no respondents. This subsection describes four uses
of semantic structure before data collection: finding redundant items,
comparing a scale with existing scales, choosing the number of factors,
and building short forms.

First, semantic structure can find redundant items. Internal consistency is the degree to which a scale's items correlate with one another, and two items that say nearly the same thing raise it without adding content \citep{christensen-etal-2023}. \citet{christensen-etal-2023} developed
Unique Variable Analysis to detect such pairs in response data, and the
same screen can be applied to a matrix of item similarities. Redundancy
has also been flagged directly from embeddings. In a preliminary study,
\citet{huang-etal-2025} clustered item embeddings. The clusters kept
the gratitude items apart from the others, but they grouped the items of
one grit facet, perseverance of effort, with the conscientiousness items.
This grouping matches the strong correlation between grit and
conscientiousness found in responses \citep{crede-etal-2017}. Prompted chatbots given the same task
recovered the broad structure but not consistent finer structure.

Second, it can check whether scales with similar names measure similar
things. Psychology has many scales for overlapping constructs, built in
research traditions that use different words for the same concept and
the same word for different concepts. Borrowing a term from the philosophy of
science \citep{sankey-1998}, \citet{wulff-mata-2025} call the
resulting difficulty of comparing theories and findings \emph{taxonomic
incommensurability}. The two naming practices behind it produce two
named errors. A jingle fallacy occurs when two scales share a label but
measure different things, and a jangle fallacy is the reverse error, in which two scales have different labels but measure the same thing
\citep{larsen-bong-2016}.
Semantic similarity makes both errors detectable. A jingle appears as
two scales whose labels are similar but whose item content is dissimilar
\citep{larsen-bong-2016,wulff-mata-2025,wulff-mata-2026}, whereas a
jangle appears as two scales whose labels are dissimilar but whose item
content is similar \citep{larsen-bong-2016,rosenbusch-etal-2020,wulff-mata-2025}.
\citet{wulff-mata-2025} embedded 4,452 IPIP items, 459 scales, and 277
construct labels in one space and flagged 504 candidate
fallacies under a quantile-based criterion. They also
argue that automated screening of this kind should inform expert
judgment rather than replace it \citep{wulff-mata-2026}.

Third, it can suggest how many factors a scale has.
\citet{garrido-etal-2025} compared methods for this decision on
similarity matrices built from the embeddings of items measuring six
personality facets. They applied two of the most common traditional
rules, parallel analysis \citep{horn-1965} and the Kaiser rule
\citep{kaiser-1960}, to the principal components of these matrices. Both
rules overestimated the number of dimensions and recovered it in none of
the 500 simulated item pools, whereas Exploratory Graph Analysis (EGA)
recovered it in 43.4\% of them. EGA treats the items as a
network and finds dimensions as communities of closely connected items
\citep{golino-epskamp-2017}. Its recovery
became almost perfect once redundant and unstable items were removed
\citep{garrido-etal-2025}. The items in that study were written by one
language model, GPT-4o, rather than taken from a published inventory,
and they were embedded with one embedding model. A related but separate
decision is how many of each embedding's leading coordinates to keep.
Later work makes this decision with a dynamic version of EGA, by
weighting agreement with a known structure against an entropy-based
measure of fit \citep{golino-2026}.

Fourth, it can build shorter forms of a scale without any data
\citep{jung-seo-2025,wang-etal-2026}. \citet{jung-seo-2025}, for
example, reworded the reverse-keyed IPIP markers into positive form,
grouped the items' embeddings into five clusters that matched the five
domains, and kept the items closest to each cluster's center. The
resulting 30-item form performed comparably to short forms built from
responses. \citet{wang-etal-2026}, who also rewrote the reverse-keyed
IPIP items in positive form, applied semantic item selection to three
inventories and removed about 60\% of their items while retaining their
factor structure in confirmatory models. The five-factor model, however,
fitted their IPIP short form only moderately well.

All four uses analyze items that already exist. Other work uses language
models at different steps of measurement: to write new items and remove
redundant ones \citep{russell-lasalandra-etal-2026}, to adapt existing scales
\citep{grobelny-etal-2025}, to stand in for human respondents
\citep{lim-etal-2026,liu-etal-2025}, to rate or code text
\citep{bunt-etal-2025,eberhardt-etal-2025}, and to measure traits
directly from what people write \citep{maharjan-etal-2025}. SFA can be
combined with these approaches, for example by screening items that a
language model has written before anyone answers them.

\subsection{The R Package \texttt{semanticfa} Brings the Whole Workflow Together}

The methods reviewed above are scattered across separate papers,
scripts, and software, and each covers only one step of the analysis.
For example, \citet{pellert-etal-2026} shared code that reproduces their
SQuID analyses, and \citet{hommel-arslan-2025} released a web
application that estimates item correlations from item texts.
Even a recent framework for evaluating text-based psychological
measures, which applies natural language processing to the texts people
write, mentioned factor analysis of text embeddings in place of
rating-scale responses only briefly, as a possible direction for future
work on text data rather than on scale items
\citep{low-etal-preprint}. The \texttt{semanticfa} package
\citep{yanitski-westbury-2026} brings the whole workflow together in R
\citep{rcoreteam-2025}. It uses the \texttt{psych} package for
factor analysis \citep{revelle-2026} and, optionally, the \texttt{EGAnet}
package \citep{golino-christensen-2026} for network analyses such as EGA
\citep{golino-epskamp-2017},
so its results can be examined with tools that researchers already use.
The package follows the four stages previewed at the start of the paper.

\begin{enumerate}
\item \emph{Similarity construction} computes the similarity between
every pair of items, either as the cosine of their embeddings, raw or
after one of the corrections described in the subsection on
reverse-keyed items, or with the NLI cross-encoder, which uses no
embeddings. The package calls the four embedding-based rules
\emph{encodings} and the cross-encoder's output the NLI matrix.
\item \emph{Retention} decides how many factors to keep. It can apply five criteria, some of them adapted to embeddings,
including parallel analysis and EGA, and report their consensus, the
number of factors they choose most often, with a tie going to the
smaller number.
\item \emph{Extraction and diagnostics} fits the factor model and reports
five diagnostics. Three of them are fit statistics, which describe how
well the fitted model accounts for the similarity matrix. An optional
test compares these three fit statistics with the values that random
embeddings would produce.
\item \emph{Interpretation} computes how similar each item is to each
construct, represented by the average embedding of the construct's items
or by the embedding of its name. It also places items on axes defined by
pairs of opposite words and compares the semantic structure with a
theoretical or response-based one.
\end{enumerate}

\noindent Beyond the four stages, optional refinement tools use a fitted
model to flag redundant items, propose shorter forms, and screen newly
written items before they are administered.

The name \emph{semantic factor analysis} has, to our knowledge, one
earlier, independent use. \citet{muller-2026} used it for an analysis that recovered much of the
Big Five structure from trait adjectives with word vectors, in which each adjective's
vector was weighted by its mean response. SFA as defined here differs in
two ways, because it needs no responses at all and it applies to any
scale items rather than only to adjectives.

The Method and Results sections form a worked example. We apply every
stage and each refinement tool to the 50-item IPIP Big-Five Factor
Markers, embedded with Qwen3-Embedding-8B, and then compare the results
with a conventional factor analysis of \valNHuman{} people's responses to
the same items \citep{openpsychometrics-2018}. The Method describes the
materials, the embeddings, and the choices a user makes at each stage,
and the Results follow the analysis step by step, from the first call to
the package to the comparison with human data. The Discussion summarizes
what the example shows, gives recommendations for users, and states the
limits of the method.

\newpage
\section{Method}

The Method describes what a user needs and the choices a user makes. It
first describes the inventory and the human data used in the worked
example, then the embeddings, then the workflow the package implements,
and finally the steps of the analysis reported in the Results.

\subsection{Materials}

The worked example uses the 50-item IPIP Big-Five Factor Markers, the
IPIP's sentence-format counterpart \citep{goldberg-etal-2006} of the
adjective factor markers of \citet{goldberg-1992}. The inventory has ten short, first-person items
for each of the five domains, such as ``I am the life of the party.''
for Extraversion. The IPIP is a public-domain collection of personality
items \citep{goldberg-etal-2006}, so these items are free to use and
reproduce without permission. It measures the five-factor structure that
\citet{goldberg-1990} recovered across nearly complete sets of English
trait adjectives and across different raters and procedures, which makes
it a good test case for a semantic method. The semantic solution can be
checked against a well-established structure, and the items are short,
concrete statements whose meaning a language model can plausibly
capture. The inventory ships with the package as \texttt{data(big5)},
which holds the item texts, the item codes (E1 to O50), precomputed item
embeddings (see Embeddings below), a scoring key of $+1$ for
forward-keyed and $-1$ for reverse-keyed items, and each item's domain
label.

The human benchmark is the Open-Source Psychometrics Project's IPIP-FFM
data set of responses to this inventory \citep{openpsychometrics-2018},
the file \texttt{IPIP-FFM-data-8Nov2018.zip}, released on November 8,
2018. Only the 50 item columns of its data file were kept, in their
original order, and renamed to the package's item codes. The README of
the reproduction archive gives the command that does this. The
respondents are a self-selected online sample who rated each item on a
5-point Likert scale, and the data file codes
a skipped item as 0. Of its \valNHumanRaw{} response records, we removed
every record with a missing response or a response outside the 1--5
range (\valNHumanRemovedPct\% of the records), which left $N =
\valNHuman$ respondents. No further screening was applied, and the
ratings were analyzed through their Pearson correlations. Data from the same project have served as a validation corpus in earlier
embedding work \citep{jung-seo-2025,wulff-mata-2025}, and
\citet{wulff-mata-2025} also used them to fine-tune their embedding
model. The package's domain labels and
scoring key follow this administration, which names the domains
Extraversion, Neuroticism, Agreeableness, Conscientiousness, and
Openness. The domain that \citet{goldberg-1992} calls Emotional Stability is
therefore labeled Neuroticism and keyed in that direction, so that ``I
get stressed out easily.'' is forward-keyed. Under this key,
\valNRev{} of the 50 items are reverse-keyed.

The key itself is used in only two places. The \texttt{atomic\_reversed}
encoding uses it to multiply the embedding of each reverse-keyed item by
$-1$, one proposed way of handling reverse-keyed items
\citep{guenole-etal-2025}. The human comparison uses it to reverse-score
the responses to those items before they are factor-analyzed, so that
the human factor solution can be compared with the semantic one.

\subsection{Embeddings}

The main analyses use precomputed item embeddings from
Qwen3-Embedding-8B \citep{qwen-2025,zhang-etal-2025}, a sentence-embedding model whose
embeddings have 4,096 dimensions. Each of the 50 item texts was embedded
once, offline. An item's embedding was taken from the model's internal
representation of the item's final token (last-token pooling), no task
instruction was placed before the item text, and the model's final step
rescaled each vector to unit length. The resulting $50 \times 4{,}096$
matrix is the \texttt{embeddings} element of \texttt{data(big5)}.

Two interpretation tools described below also need embeddings of words
that are not items. Comparing items with construct names needs the five
construct names (\emph{extraversion}, \emph{neuroticism},
\emph{agreeableness}, \emph{conscientiousness}, \emph{openness}), and
semantic projection needs the four words that define its axes
(\emph{anxious}, \emph{calm}, \emph{solitary}, \emph{sociable}). The vectors for these nine words were taken from a precomputed lexicon of
single-word embeddings made with the same Qwen3-Embedding-8B model, with
the same pooling and rescaling and stored at half precision. The three
candidate items checked with the refinement tools were embedded with the
same model in the same way as the items. Every vector the analyses
compare therefore comes from one model, and all of the comparisons use
cosines, which do not depend on vector length. These auxiliary vectors
are also part of the reproduction archive.

The model-size comparison uses the same items embedded with the two
smaller members of the family, Qwen3-Embedding-0.6B (1,024 dimensions)
and 4B (2,560 dimensions), from archives generated with the same pooling,
no instruction, and the same rescaling. The three models were taken from
their Hugging Face repositories, \texttt{Qwen/Qwen3-Embedding-0.6B},
\texttt{Qwen/Qwen3-Embedding-4B}, and \texttt{Qwen/Qwen3-Embedding-8B}
\citep{qwen-2025}. The same three models also embedded the 42 items of a
second inventory, the Depression Anxiety Stress Scales, for a repetition
of the comparison that the Supplementary Materials report. The
0.6B model is the package's default for embedding new text on a user's
own computer, because its weights need roughly 1 GB of memory at 16-bit precision,
two bytes per parameter of its nominal size, against roughly 8 GB for
the 4B model and 16 GB for the 8B model. We nevertheless recommend the 8B model when memory allows. In the
model-size comparison reported in the Results, it was the only model
whose embeddings brought the two main retention criteria, parallel
analysis and EGA, into agreement on the five theoretical factors for
this inventory, and for that reason all main analyses here use it. Model
size and embedding dimension vary together in this family, so the
comparison cannot separate the two.

Finally, one similarity matrix is not built from embeddings.
The function \texttt{sfa\_nli\_matrix()} scores all item pairs with
\texttt{cross-encoder/nli-deberta-v3-base}, an NLI cross-encoder. It is a
DeBERTa-v3 model \citep{he-etal-2023} trained on natural language
inference data such as the Stanford Natural Language Inference (SNLI)
corpus \citep{bowman-etal-2015} and the Multi-Genre Natural Language
Inference (MultiNLI) corpus \citep{williams-etal-2018}. Both corpora
consist of sentence pairs labeled for whether the first sentence entails
the second, contradicts it, or does neither. Unlike the embedding models above, which
encode each item on its own, the cross-encoder reads the two items of a
pair together and scores whether one entails or contradicts the other.
The scoring runs in Python, called from R through the \texttt{reticulate}
package \citep{ushey-etal-2025} with the Python library
\texttt{sentence-transformers}. It is
the only step of the worked example that needs network access, because
the cross-encoder is downloaded from the Hugging Face Hub on first use.

\subsection{The \texttt{semanticfa} Workflow}

The package \citep{yanitski-westbury-2026} carries out the four stages
described in the Introduction, along with the optional refinement tools,
as documented R functions \citep{rcoreteam-2025}. For each stage, this
subsection says what the stage does, what choice the user makes, and
what the package does by default. The statistical details behind every
function are collected in the Technical Reference in the Supplementary
Materials.

\textbf{Similarity construction.} An encoding is the rule that turns item
embeddings into a similarity matrix. The function \texttt{sfa\_similarity()} offers
four encodings:

\begin{enumerate}
\item The \texttt{atomic} encoding takes the cosine similarity between
the raw embeddings. \citet{guenole-etal-2025} applied this encoding and
the next to embeddings pooled over the items of each facet, and the
package applies them to single items.
\item The \texttt{atomic\_reversed} encoding first multiplies each reverse-keyed
item's embedding by $-1$, then takes cosines \citep{guenole-etal-2025}.
It requires the scoring key.
\item The \texttt{squid} (Survey and Questionnaire Item Embeddings
Differentials) encoding first subtracts the questionnaire embedding, the average
of all the item embeddings, from every item's embedding
\citep{pellert-etal-2026}. The package then takes cosines, where the
original method correlated the results.
\item The \texttt{mean\_centered\_pearson} encoding, the default, first
subtracts each
embedding's own mean value from all of its entries, then takes cosines.
The cosine of two centered vectors equals their Pearson correlation, so
the result is a Pearson correlation matrix \citep{pokropek-2026}.
\end{enumerate}

\noindent The default encoding needs no scoring key, and the matrix it
produces is a Pearson correlation matrix, which is the input that the
factor-analytic methods of the later stages expect. The function
\texttt{sfa\_nli\_matrix()} offers one alternative that uses no
embeddings. It scores each pair of items with the NLI cross-encoder as
the probability of entailment minus the probability of contradiction,
and the result is called the NLI matrix. If the NLI matrix is not a
valid correlation matrix, \texttt{sfa()} prints a message when it
factors the matrix and replaces it with a valid one, as described in the
Technical Reference.

\textbf{Retention.} The user decides how many factors to keep. Each
factor has an eigenvalue, which measures how much of the total variation
in the similarity matrix the factor accounts for. The function \texttt{sfa\_parallel()}
runs parallel analysis \citep{horn-1965}, which compares each eigenvalue
with a threshold computed from the corresponding eigenvalues of matrices
built from random data. By default the threshold is the 95th percentile
of the corresponding eigenvalues of 100 such matrices. Following
\citet{horn-1965}, the function counts the leading eigenvalues that
exceed their thresholds and stops at the first that does not, so an
eigenvalue that exceeds its threshold after that point is not counted.
In SFA the random data are random embeddings
with the same number of items and dimensions as the real ones, so a
factor that passes this comparison accounts for more variation than
chance similarity among random vectors would produce.
The function \texttt{sfa\_nfactors()} adds four more criteria and
reports their consensus, the number of factors they choose most often,
with ties going to the smaller number. When the user does not fix the
number of factors, \texttt{sfa()} uses parallel analysis.

\textbf{Extraction and diagnostics.} The main function, \texttt{sfa()},
runs the whole
workflow in one call, from item text, embeddings, or a similarity
matrix. Through the psych package \citep{revelle-2026}, it extracts the
factors by minimum residual, a method that chooses the loadings so that
the similarities the factors reproduce differ as little as possible from
the observed ones. It then rotates them by oblimin, a rotation that
allows the factors to correlate, computed with the GPArotation package
\citep{bernaards-jennrich-2005}. It reports five diagnostics, described
in Table~\ref{tab:diagnostics}: the Kaiser--Meyer--Olkin measure of
sampling adequacy (KMO) \citep{kaiser-1970,kaiser-rice-1974}, the root
mean square residual (RMSR), the common part accounted for (CAF) of
\citet{lorenzo-seva-etal-2011}, the total entropy fit index (TEFI) of
\citet{golino-etal-2021}, and McDonald's omega total for each factor,
computed from the factor's loadings \citep{revelle-zinbarg-2009}. This is
a one-factor omega for each factor's items, not the hierarchical omega,
which \citet{zinbarg-etal-2005} distinguished from omega total and
which \citet{revelle-zinbarg-2009} recommend for judging whether a test
measures one construct. The
KMO takes the normalized form of \citet{kaiser-rice-1974}, which runs
from 0 to 1, and the printout labels its value with their terms, such as \emph{marvelous} for values of .90 or more. The TEFI is
the version of the index based on Von Neumann entropy. RMSR, CAF, and
TEFI are the fit statistics, because each describes how well the factor
solution fits the similarity matrix. The cutoff conventionally applied
to the RMSR, a standardized root mean square residual of about .08 or
below, was derived for confirmatory models of response data by
\citet{hu-bentler-1999}, who recommended supplementing it with a second
fit index, because in their simulations it was most sensitive to
misspecified factor covariances and less sensitive than other indices
to misspecified loadings. Values produced
by chance can pass it, as \citet{pokropek-2026}
showed for the standardized root mean square residual of confirmatory
models. The option \texttt{calibrate = TRUE} therefore compares the
three fit statistics with the values that the same model produces on
random embeddings.
The function \texttt{as\_psych()} converts a fit into the format of the psych package,
so that the psych package's own functions can be applied to the semantic
solution.

\begin{table}[tbp]
\centering
\captionsetup{justification=raggedright,singlelinecheck=false}
\caption{Diagnostics reported by \texttt{sfa()} and how to read them}
\label{tab:diagnostics}
\tighttable
\begin{tabular}{>{\raggedright\arraybackslash}p{0.2\textwidth}>{\raggedright\arraybackslash}p{0.38\textwidth}>{\raggedright\arraybackslash}p{0.33\textwidth}}
\toprule
Diagnostic & What it measures & How to read it \\
\midrule
Kaiser--Meyer--Olkin (KMO) & How large the correlations among items are
compared with the partial correlations between pairs of items once all
other items are held constant &
From 0 to 1. Higher values mean the matrix is better suited to factor
analysis. \\
Root mean square residual (RMSR) & The typical size of the differences
between the observed similarities and those the factors reproduce &
Lower is better. The range from random embeddings (see Note) shows how
little the conventional cutoff discriminates for embeddings. A value
above that range is not evidence of misfit. \\
Common part accounted for (CAF) & How much of the common variance the
factors capture. It is one minus the KMO of what the factors leave
unexplained & From 0 to 1. Higher is better. A low value means
factorable structure remains after the factors are removed. \\
Total entropy fit index (TEFI) & How cleanly the factors divide the
items, measured with Von Neumann entropy \citep{golino-etal-2021} & Lower
is better. It has no conventional cutoff and is read against other
solutions. The range from random embeddings (see Note) shows only
whether the solution can be told apart from one fitted to random
embeddings. \\
McDonald's omega total & The share of the variance of the summed score
of a factor's items that the factor accounts for, computed from the
factor's loadings. It indexes internal consistency & From 0 to 1.
Higher is better. It is computed twice, over the items whose largest
loading is on the factor and over the items of
the theoretical domain the factor matches best (ten items in the worked
example). \\
\bottomrule
\end{tabular}
\caption*{\textit{Note}. The option \texttt{calibrate = TRUE} adds, for
the three fit statistics, RMSR, CAF, and TEFI, the range of values
obtained from random embeddings
of the same size (see the Technical Reference in the Supplementary
Materials).}
\end{table}

\textbf{Interpretation.} Four functions help the user understand what
each factor means. Two of them represent a construct by its centroid,
the average of the embeddings of the construct's items, or by its name
embedding, the embedding of the construct's name, such as
\emph{extraversion}.

\begin{itemize}
\item The function \texttt{sfa\_anchor()} adapts the approach of
\citet{wulff-mata-2025} for aligning items with construct labels, which
the package calls anchoring. It computes the cosine similarity
between each item's raw embedding and each construct, leaving the item
out of its own construct's centroid, and flags items that are more
similar to another construct than to their own.
\item The function \texttt{sfa\_project()} adapts the semantic
projection of \citet{grand-etal-2022}, scoring items by a cosine rather
than an inner product and rescaling the scores. It subtracts the embedding of one word from the
embedding of its opposite, such as \emph{anxious} from \emph{calm}, and
scores every item by the cosine between its embedding and that
difference vector. The scores are rescaled so that 0 marks the first
word and 1 the second.
\item The function \texttt{sfa\_congruence()} measures how closely the semantic
structure agrees with a theoretical assignment of items, a factor
solution from responses, or a correlation matrix. Its agreement metrics
are defined where the Results use them.
\item The function \texttt{sfa\_jinglejangle()} adapts the
label-versus-content comparison of
\citet{wulff-mata-2025,wulff-mata-2026}. For every pair of scales, it
compares the similarity of their names with the similarity of their
items and flags pairs where the two differ by 0.20 or more. This single
threshold is the package's own, whereas \citet{wulff-mata-2025} set
separate cutoffs for the two similarities from quantiles of their
distributions.
\end{itemize}

\noindent For display, \texttt{sfa\_corplot()} draws the similarity matrix
as a heatmap, and \texttt{sfa\_itemplot()} draws two-dimensional maps of
the items. The \texttt{plot()} methods draw the loadings and the scree
plot, which shows the eigenvalues in decreasing order.

\textbf{Refinement tools.} Three optional functions help the user improve
the scale:

\begin{itemize}
\item The function \texttt{sfa\_redundancy()} flags pairs of items that say nearly the
same thing, using Unique Variable Analysis \citep{christensen-etal-2023},
and suggests which item of each redundant cluster to drop.
\item The function \texttt{sfa\_simplify()} builds a candidate short form from the
items most similar to their construct's centroid, in the spirit of
earlier methods that select representative items without responses
\citep{jung-seo-2025,wang-etal-2026}.
\item The function \texttt{sfa\_item\_fit()} checks newly written items against an
existing fit, extending the anchoring approach of
\citet{wulff-mata-2025}. It reports the construct whose centroid each
item is most similar to and, as additions of the package, how much more
similar it is to that construct than to the next one and whether it
nearly duplicates an existing item.
\end{itemize}

\subsection{Analysis Plan}

A single script, \texttt{reproduce.R}, ran every analysis. Every random
step of the main analyses used the seed 42, set either by the script or
by the package function itself. Three steps used other seeds or none. The five further human
subsamples of the comparison-data check used the seeds 1 to 5, the
parallel-analysis check across model sizes used the seeds 1 to 10, and
psych's parallel analysis of the human responses was not seeded. It
runs its simulations in parallel processes (two worker processes, the
default of R's \texttt{mclapply()}, because the script does not set the
\texttt{mc.cores} option), so their last digits can differ between runs
of the script. The script is meant to be run from
top to bottom. Its output transcript, saved by \texttt{R CMD BATCH}, supplies every
number in the Results. The excerpts shown there are copied from it,
reflowed to fit the page and with omissions marked. The script runs the steps in
the order in which they were built, so its transcript does not follow
the list below. The list follows the order in which the Results report
the steps.

\begin{enumerate}
\item \emph{Fit the model.} The function \texttt{sfa()} fit the model to the
\texttt{mean\_centered\_pearson} similarity matrix of the 8B embeddings.
This encoding was chosen before any factor was interpreted, because it
needs no scoring key and produces a Pearson correlation matrix. Parallel analysis chose the number of factors, and that number, five,
was then fixed for every later step. RMSR, CAF, and TEFI were finally compared
with their values in fits to 100 sets of random embeddings of the same
size.
\item \emph{Compare the encodings.} All four embedding encodings were
built from the same embedding matrix and compared by the range of their
similarities, and atomic and mean-centered Pearson were also compared
entry by entry. To see how each encoding treats reverse-keyed items, E1
was also compared with its reverse-keyed counterpart E2 and with E5,
which is keyed the same way as E1.
\item \emph{Choose the number of factors.} Parallel analysis compared
each eigenvalue with the 95th percentile of the corresponding
eigenvalues from 100 sets of random embeddings. It counted the leading
eigenvalues that exceeded their thresholds and stopped at the first that
did not. Its verdict was set against the consensus of the five criteria
in \texttt{sfa\_nfactors()}, the number they chose most often, with ties
going to the smaller number. Velicer's
minimum average partial test \citep{velicer-1976} was also run but not
counted in the consensus. The function \texttt{sfa\_dimselect()} then
chose how many leading coordinates of each embedding to keep, the depth.
It ran EGA on the first $d$ coordinates at \valDepthNEval{} depths:
\valDepthGridN{} depths spaced \valDepthStep{} apart, from
\valDepthMin{} to \valDepthGridLast{} leading coordinates, plus the full
embedding of \valDepthFull{} coordinates. At each depth, EGA estimated
the item network with the triangulated maximally filtered graph
\citep{massara-etal-2016} and found its communities with the Walktrap
algorithm \citep{pons-latapy-2006}. Two quantities were computed at each
depth. The first was the normalized mutual information (NMI) between the
EGA communities and the theoretical domains, which is defined in the
Results. The second was the TEFI of the EGA solution. Each was rescaled across the
depths to run from 0 to 1, and the chosen depth maximized $0.70 \times
\mathrm{NMI} - 0.30 \times \mathrm{TEFI}$ on the rescaled values, so
agreement with the theoretical domains counted for more than entropy
fit (see the Technical Reference in the Supplementary Materials). The
Qwen3-Embedding models support embeddings of a chosen, shorter size
\citep{zhang-etal-2025}. This support rests on Matryoshka representation
learning \citep{kusupati-etal-2022}, which trains the leading
coordinates of an embedding to work on their own, so keeping the first
$d$ coordinates gives a usable shorter embedding. The depth search was a
separate analysis. The main fit used the full embedding, because
\texttt{sfa()} keeps every coordinate unless its \texttt{dim\_select}
argument requests the search.
\item \emph{Choose an embedding model.} The similarity matrix and the
retention criteria, with each function's default settings, were
recomputed from the 0.6B and 4B embeddings, and the first eight
parallel-analysis thresholds were recorded for each model. To see how
far each verdict of parallel analysis depends on the random embeddings
drawn, parallel analysis was also repeated for each of the three models
with the seeds 1 to 10. To
measure how clearly each model separates the five domains, we computed a
point-biserial correlation over all \valNPairs{} item pairs. A point-biserial
correlation is a Pearson correlation between a two-valued variable and a
continuous one. Here the two-valued variable records whether the two
items of a pair belong to the same domain, and the continuous variable is
their similarity, so the correlation is high when same-domain pairs are
consistently more similar than cross-domain pairs. Its uncertainty at
each size was described by the item-level bootstrap described below for
the item pairs, and the within- minus between-domain gap in similarity
was also computed for each domain separately.
\item \emph{Interpret the factors.} Each item was compared by cosine
similarity with every construct centroid and every construct name. The items were also placed by an adaptation of the semantic projection
of \citet{grand-etal-2022} on two axes, \emph{anxious}--\emph{calm} and
\emph{solitary}--\emph{sociable}, each defined by a single pair of words.
They were then mapped in two dimensions with the package's default
settings, a perplexity of 16 for $t$-distributed stochastic neighbor
embedding \citep{vandermaaten-hinton-2008} and 15 neighbors for uniform
manifold approximation and projection \citep{mcinnes-etal-2018}, and
compared with their theoretical assignment. Both maps used the seed 42,
with the other arguments of the Rtsne \citep{krijthe-2023} and uwot
\citep{melville-2025} functions at their defaults. How well each map kept the
domains apart was summarized by the share of each item's five nearest
neighbors in the map that came from its own domain. Agreement between the
factors and the theoretical domains was judged against 1,000 random
permutations of the domain labels. The five
10-item subscales were screened for jingle--jangle fallacies, with any
pair of subscales flagged when the similarity of their names and the
similarity of their items differed by 0.20 or more. This threshold is
the package's default, a heuristic rather than a published criterion.
\item \emph{Refine the items.} Redundant pairs were flagged by Unique
Variable Analysis, using its published cutoff of 0.25 on the weighted
topological overlap defined in the Results \citep{christensen-etal-2023}.
Because a similarity matrix has no respondents, the package estimates
the underlying network with a nominal sample size of 1,000, a fixed
convention described in the Technical Reference, and the number of
flagged pairs was also counted with nominal sizes of 500 and 2,000.
As a check, pairs were also flagged when their similarity in the fitted
matrix, which here matches the raw cosine, was 0.85 or higher, a
stricter cutoff than the package default of 0.80. A 25-item short form kept the five items per domain most similar to the
centroid of their domain's other items, and it was refit with the same
encoding, with its number of factors chosen by the same parallel
analysis (100 random embeddings, seed 42). To see how much of its
agreement with the domains selection alone provides, the short form and
\valSfRandN{} random sets of five items per domain were also fitted with
five factors and compared with the domains. Three newly written candidate
items were then checked against the fitted model. They were written to
illustrate the three verdicts the check can give: an item that fits one
construct cleanly, an item that cross-loads, and a near-duplicate of an
existing item. By the package's rules, a candidate cross-loads when its
similarity to its best construct exceeds its similarity to the next by
less than 0.07, and it is a near-duplicate when its similarity to an
existing item reaches the cutoff, set here to 0.85 to match the pair
screen (the function's default of 0.90 gives the same verdicts here).
\item \emph{Compare with human responses.} The responses to the
\valNRev{} reverse-keyed items were reverse-scored by replacing each
response $x$ with $6 - x$, so that 1 became 5 and 5 became 1. The
correlation matrix of these reverse-scored responses was factored with
\texttt{psych::fa} \citep{revelle-2026}, using minimum residual
extraction, oblimin rotation, and five factors fixed at the theoretical five-factor
structure \citep{goldberg-1990}. The correlation matrix of the raw responses, without reverse-scoring,
was kept for comparisons with encodings that ignore the scoring key, and
it is the comparison for the default encoding, because neither uses the
key. A
conventional parallel analysis \citep{horn-1965} of the reverse-scored
correlations, the common-factor version in \texttt{psych::fa.parallel}
with that function's default of 20 simulated data sets, was run for
comparison. It compares each common-factor eigenvalue of the responses
with the mean of the corresponding eigenvalues of random data. The Supplementary Materials add comparison-data profiles
\citep{ruscio-roche-2012} of the responses and the embeddings, with their
settings. Semantic--behavioral coherence was measured at
three levels: the factors, the item pairs, and the comparison across
encodings. At the level of factors, each semantic factor was matched to
the human factor with which its loadings had the highest Tucker
congruence \citep{lorenzo-seva-tenberge-2006}, defined in the Results,
in absolute value, and the mean of these matched values is the mean
matched congruence. The absolute value was used because a factor whose
loadings are all multiplied by $-1$ fits the data equally well, so the
sign of a factor carries no meaning. At the level of item pairs, agreement was the Pearson correlation
between each similarity matrix and each human correlation matrix over
the \valNPairs{} item pairs. For the default encoding, which ignores the
scoring key, the headline comparison is with the correlations among raw
responses. Its uncertainty was described by an item-level bootstrap,
which resampled the 50 items with replacement 1,000 times and recomputed
the correlation over the pairs of distinct items, and the 2.5th and
97.5th percentiles of the resampled values give a 95\% percentile
interval. After seeing the scatterplot, we also described the shape of
the relation, which was not planned in advance. An ordinary least squares
regression of the similarity on the human correlation and its square,
both uncentered, measured the curvature. Mean similarity was computed
for the pairs whose human correlation was above .20, within .10 of zero,
and below $-.20$, cutoffs chosen after the scatterplot was seen. The
similarities were also correlated with the absolute values of the human
correlations. Finally, the pairs were grouped into three kinds: two
items of the same domain with the same scoring key, two items of the
same domain with opposite scoring keys (an item and a reverse-keyed item
of its domain), and two items of different domains. The same item-level
bootstrap gave 95\% intervals for the curvature, measured as the
quadratic minus the linear $R^2$, for the difference in mean
similarity between the first two kinds of pairs, and for the difference
between the second and third kinds. Across encodings, a five-factor model
was fit under every encoding and to the NLI matrix and scored on the same
agreement metrics (Table~\ref{tab:s-encodings} in the Supplementary
Materials).
\end{enumerate}

\newpage
\section{Results}

Applied to the 50 IPIP Big-Five Factor Markers, the workflow produced
three factors that corresponded mainly to Neuroticism,
Conscientiousness, and Openness. Most of the Extraversion items and
\valAsgnAOnBlend{} of the Agreeableness items were divided between two
blended factors, which also took \valAsgnBlendOther{} items from other
domains, and \valAsgnAOnN{} Agreeableness items joined the Neuroticism
factor (Table~\ref{tab:assignment}). The similarities that the workflow
produced correlated $r = \valPairR$ with the correlations among the raw
responses of \valNHuman{} respondents. The Results
follow the analysis from beginning to end, in the order a user would run
it. The first step fits the model in a single call and reads its output.
The next three steps examine the choices on which this call depends: the
encoding, the number of factors, and the embedding model. The fifth and
sixth steps use the fitted model to interpret the factors and to refine
the items, and the last step compares the semantic structure with the
responses of \valNHuman{} people.

\subsection{Fitting the Model}

Verbatim transcript excerpts (abridged where noted) trace a typical
workflow, starting with the bundled dataset and its
Qwen3-Embedding-8B vectors:

{\scriptsize
\begin{verbatim}
> data(big5)   # 50 IPIP Big-Five Factor Markers + bundled Qwen3-Embedding-8B
> str(big5, max.level = 1)
List of 5
 $ items     : chr [1:50] "I am the life of the party." "I don't talk a lot."
   "I feel comfortable around people." "I keep in the background." ...
 $ codes     : chr [1:50] "E1" "E2" "E3" "E4" ...
 $ factors   : chr [1:50] "Extraversion" "Extraversion" "Extraversion"
   "Extraversion" ...
 $ scoring   : num [1:50] 1 -1 1 -1 1 -1 1 -1 1 -1 ...
 $ embeddings: num [1:50, 1:4096] 0.0361 0.0193 0.0184 0.0269 0.0081 0.0227
   0.0204 0.0234 0.0214 0.0189 ...
  ..- attr(*, "dimnames")=List of 2
> big5_df <- data.frame(code = big5$codes, item = big5$items,
+                       factor = big5$factors, scoring = big5$scoring)
> E8 <- big5$embeddings    # 50 items x 4096 embedding dimensions
\end{verbatim}
}

The package's main function, \texttt{sfa()}, runs the whole workflow in
one call. It builds the similarity matrix, chooses the number of factors,
extracts and rotates the factors, and reports the five diagnostics. Each
of these steps is also available as its own function
(\texttt{sfa\_similarity()} for the encoding, \texttt{sfa\_parallel()} for
retention) and can be run separately, as the next three steps do to
examine the choices that the call depends on. Automatic retention chose
five factors, which we will fix as \texttt{nfactors~=~5} for the rest of
the analysis. The encoding method is set to the default mean-centered
Pearson (see Method). Because this encoding ignores the scoring key, the
warning confirms that the scoring column in \texttt{big5\_df} is not
used.

{\scriptsize
\begin{verbatim}
> fit_auto <- sfa(big5_df, embeddings = E8, encoding = "mean_centered_pearson")
Warning message:
'scoring' has reverse-keyed items but is ignored for encoding =
"mean_centered_pearson": this method is keying-free by design. Use
"atomic_reversed" for keyed sign-flipping.
> cat("Auto-retained factors:", fit_auto$factors, "\n")
Auto-retained factors: 5
> fit <- sfa(big5_df, embeddings = E8, encoding = "mean_centered_pearson",
+            nfactors = 5)
> fit
Semantic Factor Analysis
  Encoding: mean_centered_pearson
  Embedding dim: 4096
  Factors:5 (minres + oblimin)

Diagnostics:
  KMO:  0.971 (marvelous - higher is better)
  TEFI: -27.7059 (lower is better)
  RMSR: 0.0323 (good - lower is better)
  CAF:  0.4120 (marginal - higher is better)

[item-by-factor loadings omitted: reported in full in the supplement]

                 MR3   MR1   MR2   MR5   MR4
SS loadings    6.421 5.330 4.111 3.193 2.736
Proportion Var 0.128 0.107 0.082 0.064 0.055
Cumulative Var 0.128 0.235 0.317 0.381 0.436

Factor correlations (Phi):
      MR3   MR1   MR2   MR5   MR4
MR3 1.000 0.685 0.606 0.577 0.539
MR1 0.685 1.000 0.538 0.582 0.555
MR2 0.606 0.538 1.000 0.465 0.446
MR5 0.577 0.582 0.465 1.000 0.398
MR4 0.539 0.555 0.446 0.398 1.000
\end{verbatim}
}

The labels MR1 to MR5 in the printout are the automatic factor names of
the psych R package \citep{revelle-2026}. MR stands for minimum residual,
the extraction method, and the numbers carry no substantive meaning.
Below, each factor is matched to the domain on whose items its absolute
loadings are largest on average.

KMO was \valKmo, which means that the correlations among the items are
large compared with the partial correlations that remain between pairs of
items once all other items are held constant, so the matrix is well
suited to factoring. RMSR was \valRmsr. This means that the five-factor
model reproduces the observed similarities with a small average error,
and by the cutoff conventionally applied to response data, a
standardized root mean square residual of about .08 or below
\citep{hu-bentler-1999}, the value counts as good. The
comparison with random embeddings at the end of this subsection shows how
little that reading means for embeddings. TEFI was $\valTefiMain$. TEFI
has no conventional cutoff, so the value is read against other solutions
and against fits to random embeddings. CAF was \valCaf. This means that
the similarities the five factors leave unexplained still share structure
that additional factors could capture, so five factors account for only
part of what the items have in common. The factor correlations ranged
from \valPhiLo{} to \valPhiHi. These high values are consistent with the
positive component of similarity that every pair of items shares, whose
size the first eigenvalue in Choosing the Number of Factors shows.
Calling \texttt{summary(fit)} adds McDonald's omega total for each
factor, a one-factor omega computed separately for each factor from its
loadings on two sets of items \citep{revelle-zinbarg-2009}:

{\scriptsize
\begin{verbatim}
McDonald's omega per factor:
 factor omega_assigned omega_theoretical matched_theoretical
    MR3          0.908             0.902         Neuroticism
    MR1          0.863             0.850   Conscientiousness
    MR2          0.830             0.560       Agreeableness
    MR5          0.709             0.583        Extraversion
    MR4          0.704             0.732            Openness
\end{verbatim}
}

\begin{figure}[tbp]
\centering
\captionsetup{justification=raggedright,singlelinecheck=false}
\includegraphics[width=0.6\textwidth]{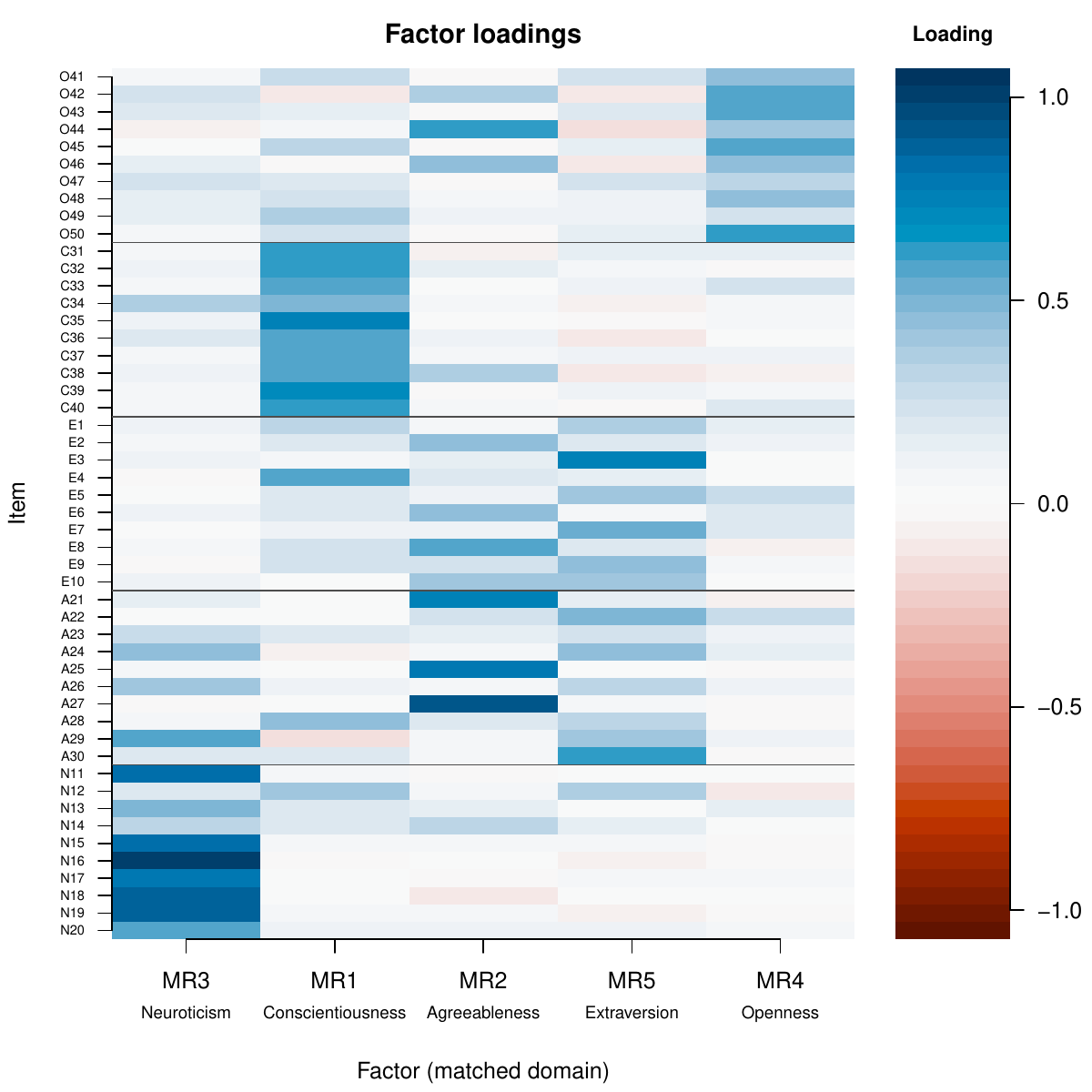}
\caption{Item-by-factor loading heatmap for the five-factor semantic
solution (mean-centered Pearson encoding of the Qwen3-Embedding-8B
embeddings). Rows are items, ordered by theoretical domain (Openness,
Conscientiousness, Extraversion, Agreeableness, Neuroticism) and
separated by horizontal lines, and columns are factors, ordered by
explained variance. Below each factor label is the domain the factor
matches best, the domain on whose items its mean absolute loading is
largest, as in Table~\ref{tab:assignment}. Two of the factors are
blended: \valMrE{} gathers forward-keyed sociability and warmth items and
\valMrA{} reverse-keyed items describing the absence of a disposition.
Color shows the oblimin pattern loading, from red (negative) through
white to blue (positive), on the scale at the right. The call
\texttt{plot(fit, type = "loadings")} draws the same map with items
sorted by their dominant factor. Item codes give the domain (O, C, E, A, N)
and the item's position in the inventory. MR = minimum residual, the
prefix of the psych package's automatic factor labels.}
\label{fig:loadings}
\end{figure}

Table~\ref{tab:assignment} assigns each item to the factor on which it
loads most strongly and cross-tabulates these assignments with the
items' theoretical domains. Three factors each correspond mainly to one
domain. \valMrN{} takes \valAsgnNOwn{} of the ten Neuroticism items,
together with \valAsgnAOnN{} Agreeableness items such as ``I feel others'
emotions.'' (A29). \valMrC{} takes \valAsgnCOwn{} of the ten
Conscientiousness items and a few items from other domains, and \valMrO{}
takes \valAsgnOOwn{} of the ten Openness items. The two remaining factors
are blended factors, which means that each mixes items from several
domains. Between them they take \valAsgnEOnBlend{} of the ten
Extraversion items, \valAsgnAOnBlend{} of the ten Agreeableness items,
and \valAsgnBlendOther{} items from other domains. Their composition is
consistent with a grouping by topic and phrasing rather than by domain. \valMrE{} gathers \valAsgnESize{}
forward-keyed items about sociability and warmth from both domains, such
as ``I feel comfortable around people.'' (E3) and ``I make people feel
at ease.'' (A30). \valMrA{} gathers \valAsgnASize{} items, all of them
reverse-keyed, that describe the absence of a disposition, such as ``I
am not really interested in others.'' (A27), ``I don't talk a lot.''
(E2), and ``I am not interested in abstract ideas.'' (O44). These items come
from \valAsgnADomainsWord{} domains, so the embeddings appear to group them by their
negated phrasing as well as by their topic (Figure~\ref{fig:loadings} and
Table~\ref{tab:s-semantic-loadings}). Not every reverse-keyed item joined
this factor, however. The other \valAsgnRevOff{} of the inventory's
\valNRev{} reverse-keyed items load most on other factors, \valAsgnCRev{}
of them on \valMrC. Some of these counts rest on near ties, items whose
two largest loadings differ by less than .03 (Table~\ref{tab:assignment}).
A24 decides whether the Neuroticism factor takes \valAsgnAOnN{}
Agreeableness items or one fewer, and N14 and O46 decide how many items
from other domains the blended factors take, so these counts could
change under a slightly different solution. An item's communality is the share
of its variance that the five factors jointly reproduce. Communalities
ranged from \valCommMin{} (\valCommMinItem, ``I insult people.'') to
\valCommMax{} (\valCommMaxItem, ``I get upset easily.''). This means that
the five factors reproduce more than half of the variance of even the
least well explained item. The loading of \valCommMaxItem{} on \valMrN{}
slightly exceeds 1, which an oblique solution permits.

\begin{table}[tbp]
\centering
\captionsetup{justification=raggedright,singlelinecheck=false}
\caption{Assignment of the 50 items to the five semantic factors, by
theoretical domain}
\label{tab:assignment}
\tighttable
{\footnotesize
\begin{tabular}{lrrrrrrr}
\toprule
& \multicolumn{5}{c}{Theoretical domain} & & \\
\cmidrule(lr){2-6}
Factor (matched domain) & O & C & E & A & N & Items & Reverse-keyed \\
\midrule
MR4 (O) & \textbf{7} & 0 & 0 & 0 & 0 & 7 & 1 \\
MR1 (C) & 1 & \textbf{10} & 1 & 1 & 1 & 14 & 6 \\
MR5 (E) & 0 & 0 & \textbf{5} & 2 & 0 & 7 & 0 \\
MR2 (A) & 2 & 0 & 4 & \textbf{3} & 1 & 10 & 10 \\
MR3 (N) & 0 & 0 & 0 & 4 & \textbf{8} & 12 & 1 \\
\bottomrule
\end{tabular}

}
\caption*{\textit{Note}. Each item is assigned to the factor on which its
absolute oblimin loading is largest (mean-centered Pearson encoding of
the Qwen3-Embedding-8B embeddings). Rows are factors, each labeled with
the domain on whose items its mean absolute loading is largest, and bold
counts are those of the matched domain. For the two blended factors this
label rests on a small margin. \valMrA{}'s mean absolute loading on the
Agreeableness items exceeds that on the Extraversion items by only
\valLabelMarginA, and \valMrE{}'s on the Extraversion items exceeds that
on the Agreeableness items by only \valLabelMarginE. O = Openness, C =
Conscientiousness, E = Extraversion, A = Agreeableness, N = Neuroticism.
Items = number of items assigned to the factor. Reverse-keyed = how many
of them are reverse-keyed. For \valAsgnNearTieN{} items, the two largest
absolute loadings differ by less than .03, so their assignments are near
ties. MR = minimum residual, the prefix of the psych package's automatic
factor labels.}
\end{table}

The two omega columns in the printout give a second view of the same
pattern. For each factor, omega is computed twice, over the items whose
largest loading is on that factor (the assigned omega) and over the ten
items of the theoretical domain the factor matches best (the theoretical
omega). The assigned item sets are chosen by their loadings, so at equal
size their omega would be expected to be the higher of the two. The sets
differ in size, however, and omega rises with the number of items. For
\valMrN{} and \valMrC{} the two values are close (\valOmegaAsgnN{}
against \valOmegaTheoN, and \valOmegaAsgnC{} against \valOmegaTheoC),
because those factors' item sets overlap largely with their domains. For
the Openness-matched factor, \valMrO, the ten theoretical items reached
\valOmegaTheoO, slightly above the \valOmegaAsgnO{} of its
\valAsgnOSize{} assigned items. For \valMrA{} the assigned omega is
\valOmegaAsgnA{} against a theoretical Agreeableness omega of
\valOmegaTheoA, and for \valMrE{} it is \valOmegaAsgnE{} against
\valOmegaTheoE. The gap is much larger for these two factors than for the
other three, which is consistent with how far their item sets depart from
the theoretical domains (Table~\ref{tab:assignment}). The comparison is
descriptive only.

Conventional cutoffs for fit statistics need not transfer to
embedding-derived matrices. In confirmatory models with word
embeddings, \citet{pokropek-2026} found that values produced by chance
often passed the cutoff for the standardized root mean square residual
in models with four to eight indicators, although less often as the
number of indicators grew. That
cutoff was derived from fits to simulated samples of respondents of
several sizes \citep{hu-bentler-1999}, so it was set with the sampling
error of response correlations in mind, that is, with how much a
correlation varies from one sample of respondents to the next. An
embedding similarity matrix has no respondents. Adding
\texttt{calibrate~=~TRUE} to the call compares RMSR, CAF, and TEFI with
their values in \valCalReps{} fits to random embeddings of the same size.
The random embeddings are random vectors of unit length, and the cosine
similarities among them are close to zero. These cosines are computed
without centering the random vectors, whatever encoding the real fit
used, whereas the default encoding of the observed fit centers each
embedding. These random fits passed the
conventional RMSR cutoff easily, with RMSR between \valCalRmsrMin{} and
\valCalRmsrMax, because a matrix of near-zero similarities leaves the
factors almost nothing to reproduce. Passing that cutoff therefore says
little about fit. For the same reason the observed RMSR (\valRmsr),
although higher than in every random fit, cannot be read as misfit,
because RMSR grows with the size of the similarities being reproduced.
TEFI behaves similarly. The observed value ($\valTefiMain$) was lower than in every random fit
($\valCalTefiMin$ to $\valCalTefiMax$, mean $\valCalTefiMean$), so
none of the \valCalReps{} random fits was as low, which shows only that the real items are
not random. The oblimin rotation did not converge in \valCalNonConvWord{} of the
random fits, and over the \valCalConvN{} fits in which it converged TEFI
ranged from $\valCalTefiConvMin$ to $\valCalTefiConvMax$. CAF has a different reference value.
Residuals with no shared structure give a CAF near .5, as the random fits
did (\valCalCafMin{} to \valCalCafMax, mean \valCalCafMean, $SD =
\valCalCafSd$). The maximum came from the fit whose rotation did not
converge, and the other fits reached at most \valCalCafConvMax. The observed CAF (\valCaf) was
lower than in all of the \valCalReps{} random fits. This suggests that the similarities the
five factors leave unexplained still contain shared structure that random
embeddings lack, although, as for RMSR, a baseline of near-zero
similarities makes this a weak contrast. The Technical Reference in the Supplementary Materials
describes how the random fits are built and why they are a weak baseline.

Finally, \texttt{as\_psych(fit)} converts the fit into the format of the
psych package, so that functions such as \texttt{psych::fa.sort()} and
\texttt{psych::factor.congruence()} can treat it like a response-based
solution.

\subsection{Choosing an Encoding}

There are several ways to turn item embeddings into the similarity
matrix that gets factored. The package implements four. The \texttt{atomic} encoding takes the cosine similarity between the raw embeddings.
The \texttt{atomic\_reversed} encoding first multiplies each reverse-keyed item's
embedding by $-1$ (here, \valNRev{} of the 50 items), then takes
cosines \citep{guenole-etal-2025}. The \texttt{squid} encoding (Survey and Questionnaire Item Embeddings
Differentials) subtracts the questionnaire embedding, the mean of all
the items' embeddings, from every item's embedding
\citep{pellert-etal-2026}. The package then takes cosines, where the
original method correlated the results.
The \texttt{mean\_centered\_pearson} encoding, the package's default,
subtracts each embedding's own mean value from all of its entries. The
cosine of two such centered vectors is exactly their Pearson
correlation, so the similarity matrix is a correlation matrix. The function \texttt{sfa\_similarity()} builds all four
from the same embedding matrix:

{\scriptsize
\begin{verbatim}
> sim_at  <- sfa_similarity(E8, "atomic",
+                           factors = big5$factors, codes = big5$codes)
> sim_ar  <- sfa_similarity(E8, "atomic_reversed", scoring = big5$scoring,
+                           factors = big5$factors, codes = big5$codes)
> sim_sq  <- sfa_similarity(E8, "squid",
+                           factors = big5$factors, codes = big5$codes)
> sim_mcp <- sfa_similarity(E8, "mean_centered_pearson",
+                           factors = big5$factors, codes = big5$codes)
> enc_range <- t(sapply(list(atomic = sim_at, atomic_reversed = sim_ar,
+                            squid = sim_sq, mean_centered_pearson = sim_mcp),
+                       function(s) round(range(s[lower.tri(s)]), 3)))
> colnames(enc_range) <- c("min", "max")
> enc_range
                         min   max
atomic                 0.394 0.938
atomic_reversed       -0.777 0.938
squid                 -0.267 0.827
mean_centered_pearson  0.394 0.938
\end{verbatim}
}

\begin{figure}[tbp]
\centering
\captionsetup{justification=raggedright,singlelinecheck=false}
\includegraphics[width=0.6\textwidth]{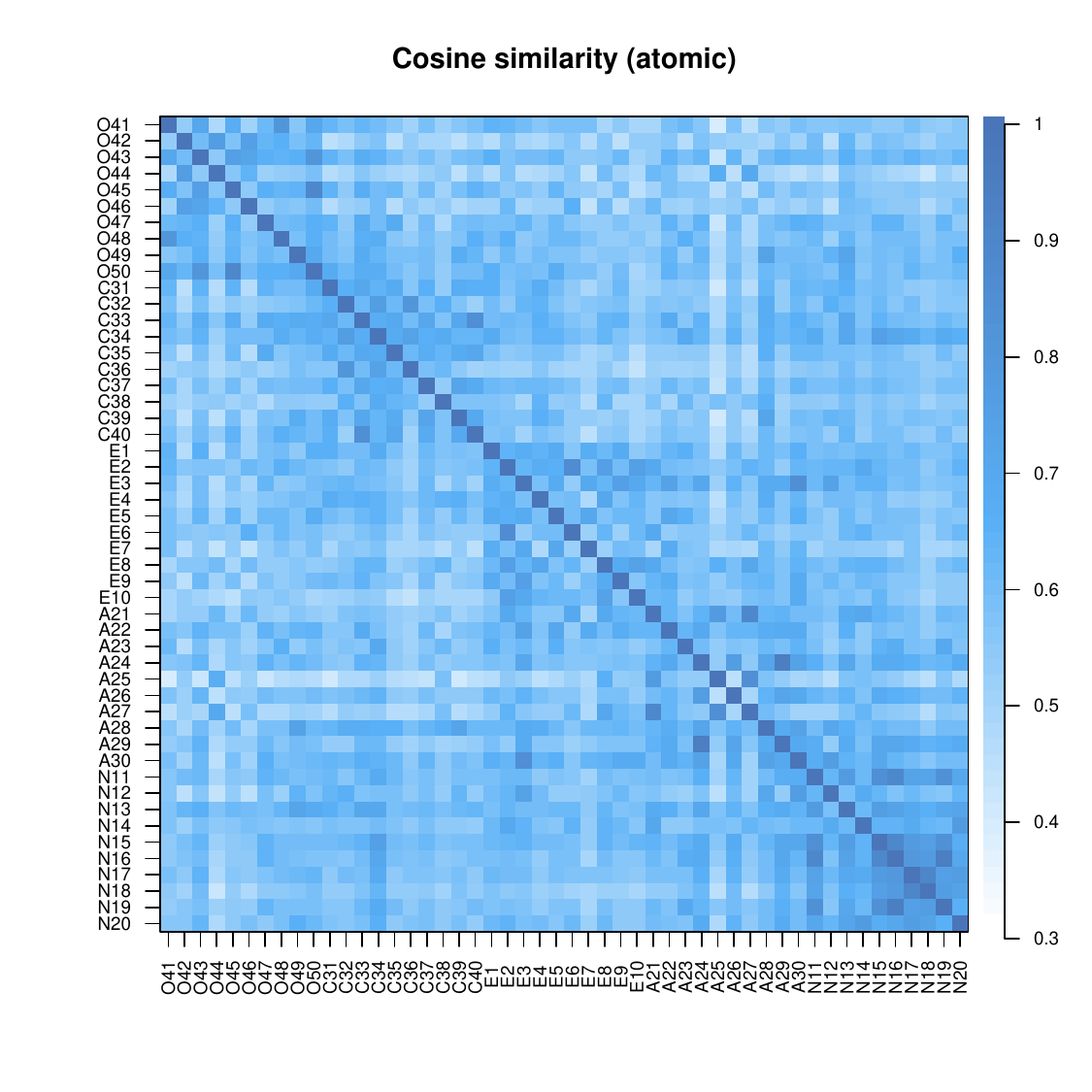}
\caption{Cosine similarity (\texttt{atomic} encoding) among the 50
International Personality Item Pool Big-Five Factor Marker items, grouped
by domain in the order Openness, Conscientiousness, Extraversion,
Agreeableness, and Neuroticism (\texttt{sfa\_corplot()}). All pairwise
similarities are positive (range = \valRangeCosMin--\valRangeCosMax),
so the color scale runs from 0.3 (white) to 1 (dark blue). The diagonal
is 1 by definition and is excluded from the stated range. Darker cells
mark higher similarity. The block structure along the diagonal is
clearest among the Neuroticism items N15 to N19 and faint in the other
domains. Item codes give the domain (O, C, E, A, N) and the item's
position in the inventory.}
\label{fig:corplot}
\end{figure}

All \valNPairs{} raw cosines are positive, from \valRangeCosMin{} to
\valRangeCosMax{} (Figure~\ref{fig:corplot}). They are positive even
between items keyed in opposite directions, such as E1 (``I am the life
of the party.'') and E2 (``I don't talk a lot.''), whose cosine is
\valAtOpp. This replicates the finding that raw cosines between
embeddings rarely mark opposite meanings with negative values
\citep{hommel-arslan-2025,pellert-etal-2026}. The negative values in the other rows are imposed by
the transformation: sign-flipping produces the $\valRangeArMin$ and subtracting the
questionnaire embedding the $\valRangeSqMin$. The default mean-centered Pearson encoding gave the same range as the
raw cosines, and entry by entry the two sets of similarities differed by
at most \valEncMaxDiff{} ($r = \valEncR$), so centering each embedding
made no practical difference for these embeddings.

The NLI matrix (see Method) differs from all four embedding encodings, because its
negative values come from the cross-encoder's judgment that two items
contradict each other rather than from a transformation of embeddings.
It scores E1 (``I am the life of the party.'') against the reverse-keyed
E2 (``I don't talk a lot.'') at $\valNliOpp$, which means that the cross-encoder judged the pair to be almost
certainly contradictory. Sign-flipping gives the same pair a cosine of
$\valArOpp$, the raw cosine with its sign reversed. Multiplying E2's embedding by $-1$ makes the cosine negative, but the
pair is not the most dissimilar one. The lowest value in the
sign-flipped matrix is $\valRangeArMin$, \valArOppPct\% of its
similarities are at or below the E1--E2 value, and the pairs of an item
and a reverse-keyed item of the same domain average $\valArOppMean$. The
transcript excerpt, the full NLI similarity heatmap, and the full
comparison with the four embedding encodings are reported in the
Supplementary Materials (Figure~\ref{fig:corplot-nli}). Which encoding
best matches human responses is tested in the last step of the Results.

\subsection{Choosing the Number of Factors}

The five factors of the main fit were chosen by the retention step inside
\texttt{sfa()}, and \texttt{sfa\_parallel()} runs that step on its own.
With response data, parallel analysis compares each eigenvalue with
those of random data sets of the same size \citep{horn-1965}. SFA has no
respondents, so \texttt{sfa\_parallel()} builds the comparison matrices
from random embeddings instead. It compares each eigenvalue with the
95th percentile of the corresponding eigenvalues from 100 such matrices.
It then counts the leading eigenvalues that exceed their thresholds and
stops at the first that does not (Horn's sequential rule):

{\scriptsize
\begin{verbatim}
> pa <- sfa_parallel(sim_mcp, E8)
> pa
Embedding-adapted parallel analysis
  Suggested factors: 5
  Embedding dimension: 4096
  Iterations: 100
  Percentile: 95
\end{verbatim}
}

Because parallel analysis is only one of several rules for this decision,
and the rules can disagree, \texttt{sfa\_nfactors()} adds four more and
reports their consensus. The consensus is the number the criteria choose
most often, and a tie goes to the smaller number:

\begin{enumerate}
\item The Kaiser rule (\texttt{kaiser} in the printout) keeps every factor
whose eigenvalue is greater than one \citep{kaiser-1960}.
\item The empirical Kaiser criterion (\texttt{EKC}) compares each
eigenvalue with a reference value that depends on the eigenvalues before
it and on the ratio of the number of items to the sample size, and it
never lets that reference fall below one \citep{braeken-vanassen-2017}.
The criterion was derived for correlation matrices of sample data, so
the package puts the embedding dimension in place of the sample size.
\item The TEFI criterion fits solutions with one to 15 factors and keeps
the number of factors whose solution has the lowest TEFI. The package
uses the Von Neumann entropy version of the index
\citep{golino-etal-2021}. Its authors developed the index to compare
candidate structures for response data, so its use on a similarity
matrix goes beyond the settings they tested.
\item EGA \citep{golino-epskamp-2017} keeps the number of communities of
closely connected items that it finds in the item network. The benchmark of \citet{garrido-etal-2025}
used personality items written by a language model (GPT-4o) and a single
embedding model. In that benchmark, EGA recovered the dimensionality of
the embedding matrices more accurately than parallel analysis and the
Kaiser rule, both applied with principal component extraction.
\end{enumerate}

\noindent The Technical Reference describes how each criterion is adapted
to embeddings. The following call runs all five on the main similarity
matrix:

{\scriptsize
\begin{verbatim}
> nf <- sfa_nfactors(sim_mcp, embeddings = E8,
+                    methods = c("parallel", "kaiser", "TEFI", "EGA", "EKC"))
> nf
Factor retention analysis (embedding-adapted)

  Method       n_factors
  parallel     5
  kaiser       6
  TEFI         2
  EGA          5
  EKC          6
  ------------------------
  Consensus    5

Eigenvalues: 29.63   1.92   1.74   1.48   1.27   1.01   0.93   0.88   0.78   0.70  ...
\end{verbatim}
}

\begin{figure}[tbp]
\centering
\captionsetup{justification=raggedright,singlelinecheck=false}
\includegraphics[width=0.6\textwidth]{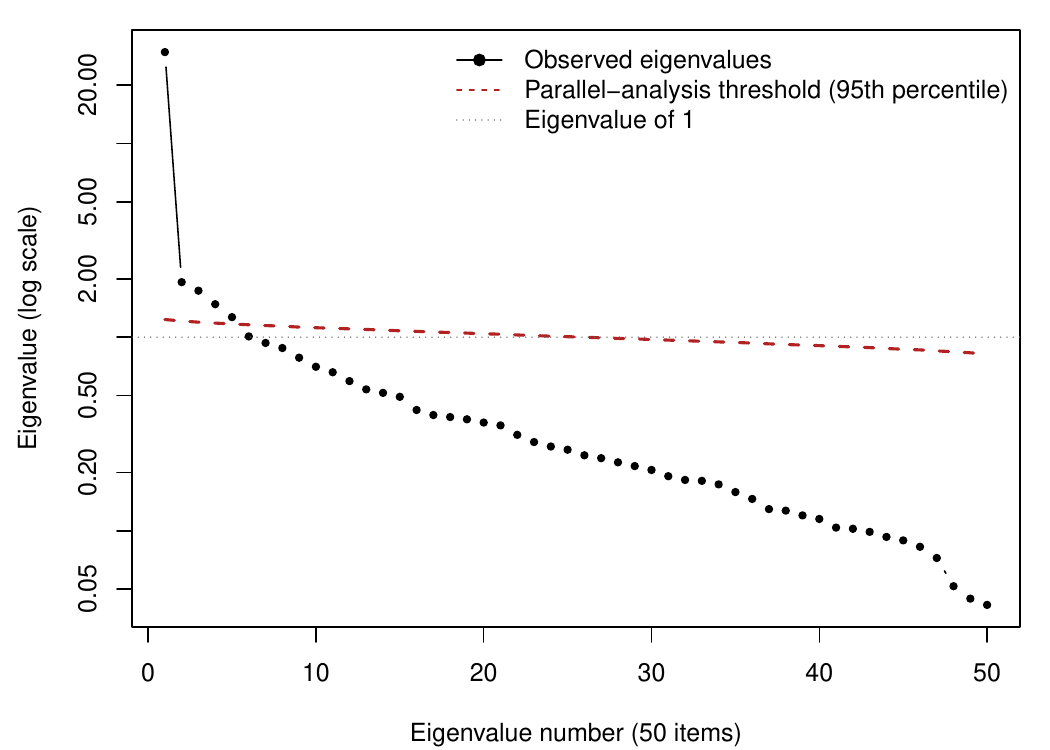}
\caption{Eigenvalues of the mean-centered Pearson similarity matrix of the
Qwen3-Embedding-8B embeddings of the 50 items in decreasing order (a scree plot, on a logarithmic scale). The
dashed red line is the parallel-analysis threshold, the 95th percentile
of the corresponding eigenvalues of 100 similarity matrices of random
embeddings, and the dotted line marks an eigenvalue of 1, the Kaiser
rule's cutoff. The first eigenvalue (\valEigFirst) is far larger than
the rest, and \valPaKWord{} eigenvalues exceed the parallel-analysis
threshold.}
\label{fig:scree}
\end{figure}

Parallel analysis and EGA indicated \valPaKWord{} factors, the Kaiser rule
\valKaiserKWord, the empirical Kaiser criterion \valEkcKWord, and TEFI
\valTefiKWord. Five and six were therefore each chosen by two criteria,
and the tie rule makes the consensus \valConsensusKWord, which matches the
number of domains the inventory was written to measure. The Kaiser rule
and the empirical Kaiser criterion agree here for a mechanical reason. Once the large first
eigenvalue is set aside, the empirical Kaiser criterion's reference
values fall below one and are raised to one, so for this matrix the two
criteria amount to a single vote. The first eigenvalue (\valEigFirst, or
\valEigFirstPct\% of the total of 50) is far larger than all the others
(Figure~\ref{fig:scree}). This means that one general component, the
similarity that every pair of items shares, accounts for most of the
total variation in the matrix. The eigenvalues after it decline
gradually. A scree plot is usually read by looking for an elbow, the
point after which the eigenvalues level off into a straight line
\citep{cattell-1966}, and
because this plot has no sharp elbow it does not settle the number of
factors by itself. The decision therefore rests on the formal criteria
above.

Embeddings raise one further question that response data do not: how
many of the 4{,}096 coordinates of each embedding are needed to recover
the structure. The number of leading coordinates kept is called the depth, and the
package's printouts call the full size of the embedding the embedding
dimension. The function \texttt{sfa\_dimselect()}, described in the
Technical Reference, rebuilds the similarity matrix from the first $d$
coordinates for a grid of depths $d$ and chooses the depth whose EGA
solution best balances agreement with the theoretical domains against
TEFI. The grid had \valDepthGridN{} depths spaced \valDepthStep{} apart,
from \valDepthMin{} to \valDepthGridLast{} leading coordinates, plus the
full embedding of \valDepthFull{} coordinates. The function selected
roughly the first third of the coordinates (\valDepthOpt{} of
\valDepthFull), at which EGA found \valDepthNDimWord{} communities. Their
agreement with the theoretical domains, measured by the normalized
mutual information (NMI), which runs from 0 for unrelated groupings to 1
for identical ones and is defined under Interpreting the Factors, was
\valDepthNmi, the maximum over the depths evaluated. Across depths NMI
ranged from \valDepthNmiMin{} to \valDepthNmiMax.
Figure~\ref{fig:s-depth} in the Supplementary Materials shows NMI and
TEFI at every depth. NMI rose quickly over the first few hundred
coordinates, stayed high but jagged over a broad middle range, and
declined toward the full embedding, while TEFI changed sharply from one
depth to the next. The selected depth is therefore the highest point of
a noisy curve rather than a sharply defined optimum. At the full depth, EGA found \valDepthFullNDimWord{}
communities (NMI = \valDepthFullNmi). These NMI values compare EGA
communities with the domains and are distinct from the NMI of the
five-factor fit reported under Interpreting the Factors. Because the
choice of depth already favors agreement with theory, this result is
consistent with the main fit rather than an independent confirmation of
it.

\subsection{Choosing an Embedding Model}

\begin{figure}[tbp]
\centering
\includegraphics[width=0.32\textwidth]{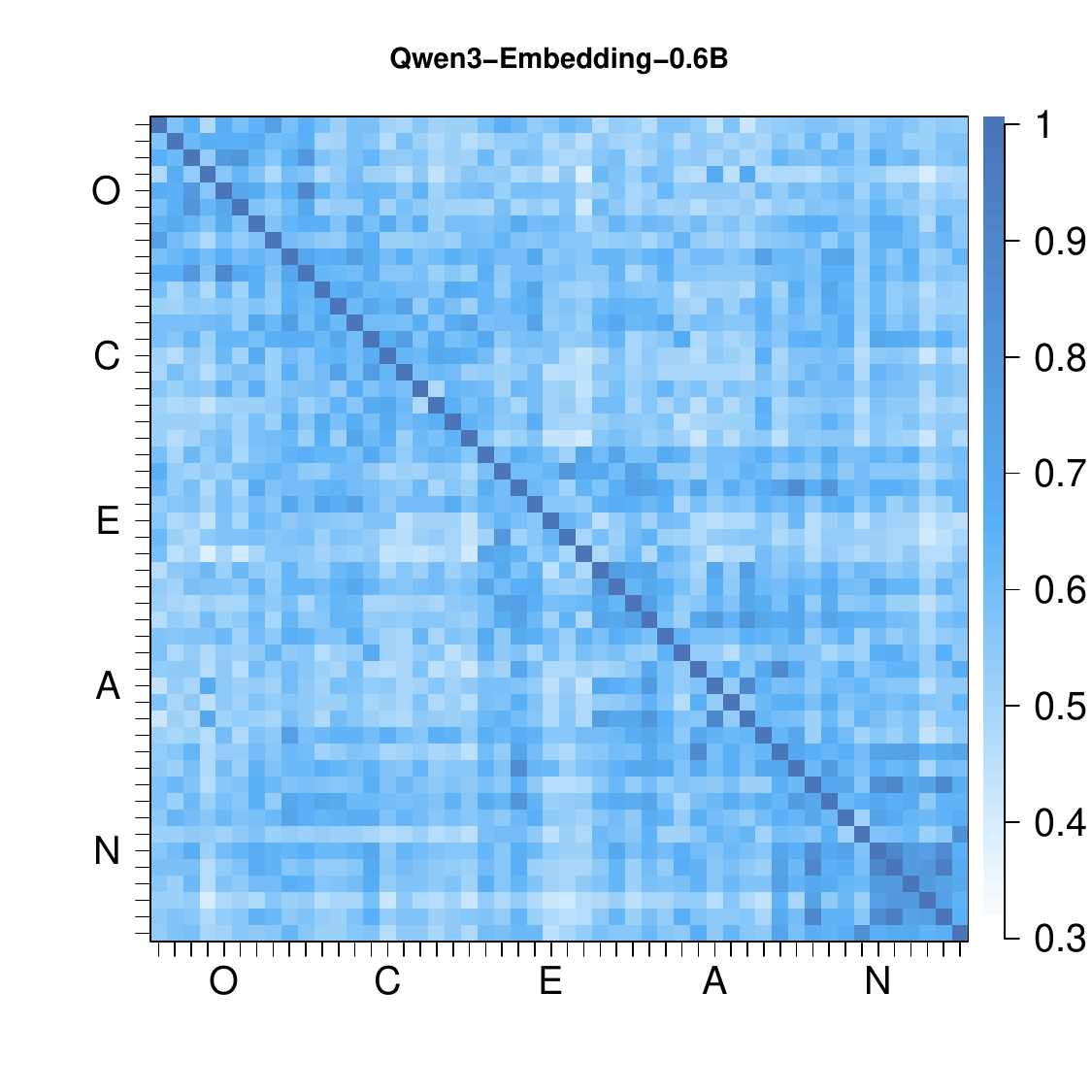}\hfill
\includegraphics[width=0.32\textwidth]{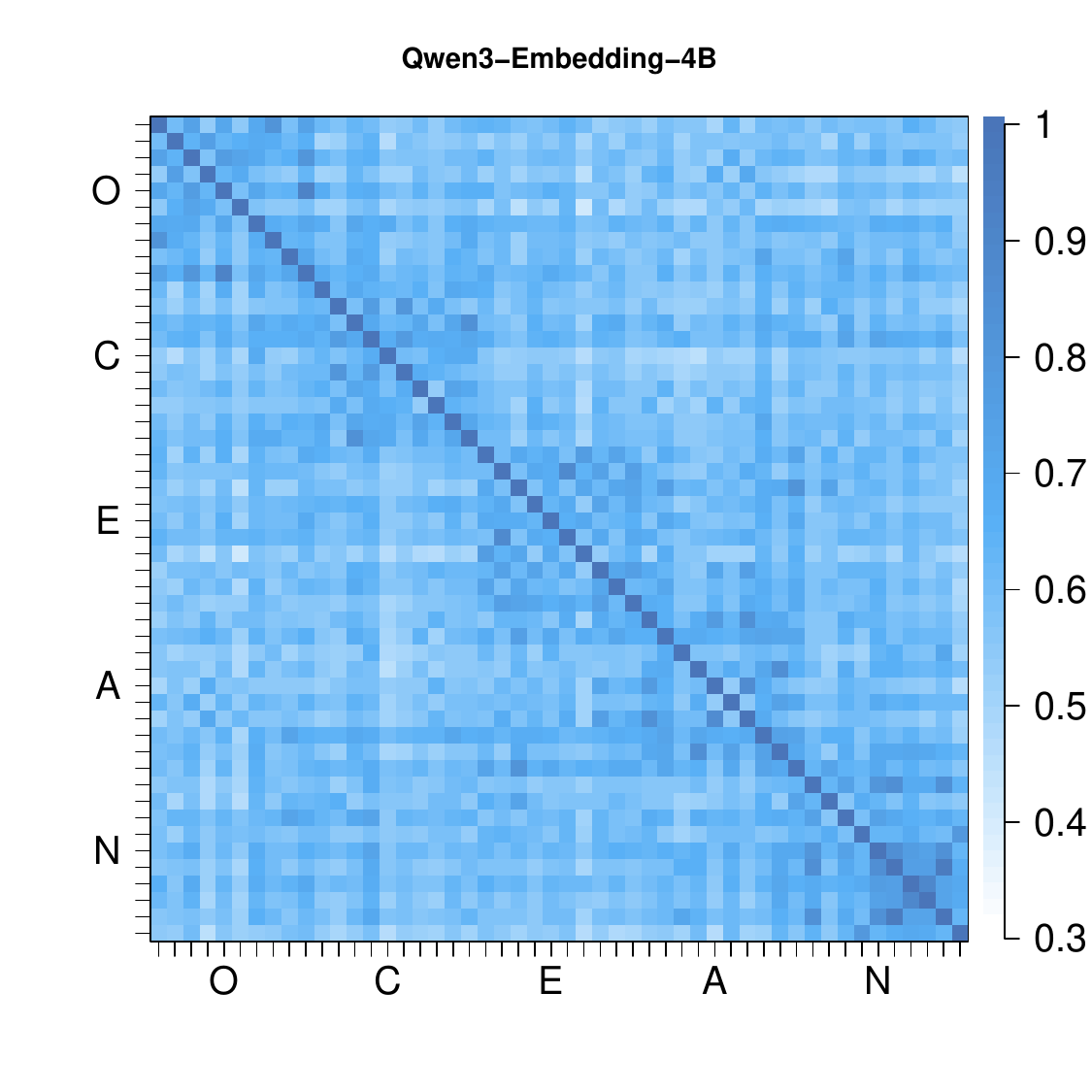}\hfill
\includegraphics[width=0.32\textwidth]{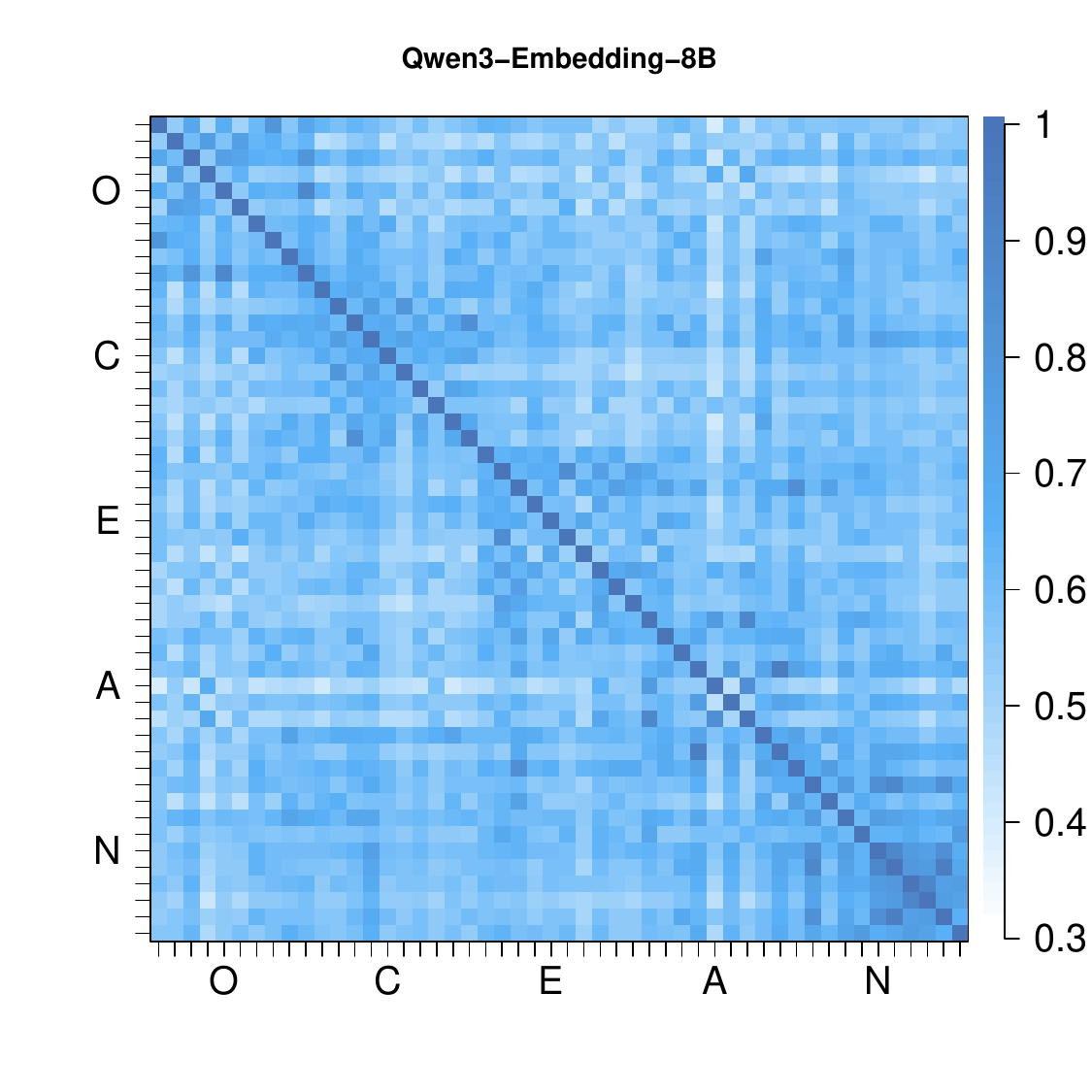}
\captionsetup{justification=raggedright,singlelinecheck=false}
\caption{Mean-centered Pearson similarity matrices of the 50
International Personality Item Pool items, grouped by domain in the order
Openness, Conscientiousness, Extraversion, Agreeableness, and
Neuroticism, for Qwen3-Embedding-0.6B (left), 4B (center), and 8B
(right). Each panel shows all 50 items against all 50, and the axes mark
each domain's block with the domain's initial letter (O, C, E, A, N) at
the middle of the block. The three panels share one color scale of
mean-centered Pearson similarity, from white at \valMsFloor, the lowest
similarity in any of the three matrices rounded down to the nearest
tenth, to dark blue at 1, and the color key is marked in steps of 0.1.
The diagonal is 1 by definition.
The Neuroticism block along the diagonal is clearly visible at every
size, and the blocks of the other domains are faint. The text reports
two measures of this separation. The point-biserial correlation between
sharing a domain and pair similarity is \valSepSmall{} (0.6B),
\valSepMedium{} (4B), and \valSepLarge{} (8B), and the mean similarity of
pairs within a domain exceeds that of pairs between domains by
\valSepGapSmall, \valSepGapMedium, and \valSepGapLarge, respectively.}
\label{fig:modelsize}
\end{figure}

The last choice the call depends on is the embedding model, and in
particular its size. To measure how much size matters, the same 50 items
were embedded with three sizes of one model family, Qwen3-Embedding-0.6B,
4B, and 8B (see Method), and the mean-centered Pearson similarity matrix
was built from each. Figure~\ref{fig:modelsize} shows the three matrices
side by side with the items grouped by domain. At every size, items are
on average more similar to items of their own domain than to items of
other domains. In the figure, the Neuroticism block along the diagonal is
clearly visible at every size, whereas the blocks of the other domains
are faint. Two statistics measure the separation more precisely. The
point-biserial correlation between sharing a domain and a pair's
similarity, computed over all \valNPairs{} item pairs, was \valSepSmall{}
for 0.6B, \valSepMedium{} for 4B, and \valSepLarge{} for 8B (item-level
bootstrap 95\% intervals \valSepCiSmallLo{} to \valSepCiSmallHi,
\valSepCiMediumLo{} to \valSepCiMediumHi, and \valSepCiLargeLo{} to
\valSepCiLargeHi), and the mean
similarity of pairs within a domain exceeded that of pairs between
domains by \valSepGapSmall, \valSepGapMedium, and \valSepGapLarge. By
both statistics, the separation between domains was similar at the three
sizes and did not rise steadily with model size. The intervals
resample items rather than pairs, because the pairs share items, and
they overlap widely, so the three sizes cannot be told apart on this
statistic. Computed for each domain separately, the mean similarity of
the domain's own pairs exceeded that of its pairs with other domains at
every size, and the excess was largest for Neuroticism
(\valDomGapNSmall, \valDomGapNMedium, and \valDomGapNLarge, against at
most \valDomGapOtherMax{} for any other domain), the domain whose block
stands out in the figure.

What changed with size was the number of factors that some retention
criteria indicated, each from a single seeded run with the package's
default settings.
Parallel analysis retained \valPaKSmallWord{} factors for 0.6B and
\valPaKMediumWord{} for 4B, while EGA retained \valEgaKSmallWord{} and
\valEgaKMediumWord, and the consensus across the five criteria was
\valConsSmallWord{} for 0.6B and \valConsMediumWord{} for 4B. For 0.6B,
the parallel-analysis verdict of \valPaKSmallWord{} rested on a narrow
margin, because the fourth eigenvalue, \valPaSmallFourObs, fell just
below its threshold of \valPaSmallFourThr. Repeated with each of ten further
seeds, parallel analysis gave the same verdict in every run for each
model size (\valPaKSmallWord{} for 0.6B, \valPaKMediumWord{} for 4B, and
\valPaKWord{} for 8B), so the verdicts did not depend on the random
embeddings drawn. Under the 8B embeddings, parallel analysis and EGA both retained the
theoretical \valPaKWord, and the consensus was \valConsensusKWord. The
other criteria did not move toward five. Under the 8B embeddings the
Kaiser rule indicated \valKaiserKWord{} factors and the empirical Kaiser
criterion \valEkcKLargeWord, as at the smaller sizes, and TEFI indicated
\valTefiKWord{} at every size, so the criteria still disagreed at 8B.
Within this family, then, the 8B model was the only one for which
parallel analysis and EGA both indicated the five theoretical factors,
even though the separation between domains changed little. Model size
and embedding dimension also vary together here (1,024, 2,560, and
4,096 dimensions), and the dimension enters both the random baseline of parallel analysis,
whose first threshold was \valPaThrSmall, \valPaThrMedium, and
\valPaThrLarge{} at the three sizes, and the reference values of the
empirical Kaiser criterion. The
comparison is therefore descriptive and cannot separate size from
dimension. Agreement with the theory the inventory was written to
express was also the reason the 8B model was used for the main analysis.
That agreement concerned the number of factors, because only under the
8B embeddings did the retention criteria lead to five factors. When five
factors were extracted at every size, the 8B solution did not match the
domains best. The NMI between each item's strongest factor and its
domain, a measure of agreement defined under Interpreting the Factors,
was highest for 4B (\valBigNmiMedium, against \valBigNmiSmall{} for 0.6B
and \valBigNmiLarge{} for 8B, Table~\ref{tab:s-modelsize}).

A second inventory, the Depression
Anxiety Stress Scales \citep{lovibond-lovibond-1995}, showed a clearer
effect of size (see Model Size With a Second Scale in the Supplementary
Materials). There, the gap between within-scale and between-scale
similarity rose from the 0.6B to the 8B model, although the paired 95\%
interval of the increase, from $\valDassGapDiffLo$ to
\valDassGapDiffHi, just included zero. The three-factor solution merged
two of the three scales under the 0.6B model but separated all three
under the 8B model. The NMI between each item's strongest factor and its
scale rose with each larger model, from \valDassNmiSmall{} to
\valDassNmiMedium{} and \valDassNmiLarge, and the rise held for a second
0.6B solution described in the Supplementary Materials, whose NMI was
\valDassAltNmi. For some scales, then, a larger model can give factors
that match the scales more closely, even though it did not for the Big
Five inventory. For users of
this model family, we therefore suggest the largest model their
computer's memory allows. Both results rest on three
models of one family and two English inventories, so whether the
suggestion holds for other model families and inventories remains to be
tested. The package's default for embedding new text
remains the 0.6B model, because it needs the least memory (see Method).
The Supplementary Materials apply one further retention method, the
comparison-data method, to the same three matrices and compare its
results with those for the human responses.

\subsection{Interpreting the Factors}

Once the factors are fitted, the next task is to understand what each
one means and whether each item sits with the construct it was written
for. The function \texttt{sfa\_anchor()} adapts the approach of
\citet{wulff-mata-2025}. It computes the cosine similarity between each
item's raw embedding and every construct, represented by the construct's
centroid or by the embedding of its name, and it leaves each item out of
its own construct's centroid. The printout lists the eight items whose
similarity to their own construct is lowest and marks those that are
more similar to another construct:

{\scriptsize
\begin{verbatim}
> anc <- sfa_anchor(fit, anchor = "centroid")
> anc
Semantic anchoring
  Method: Wulff & Mata (2025)
  Anchor: centroid
  Constructs: 5 | Items: 50

[50 x 5 item-by-construct cosine matrix omitted]

  Weakest own-construct similarity (review/removal candidates):
    N12          Neuroticism        own=0.69   <-- higher on Extraversion (0.72)
        "I am relaxed most of the time."
    O44          Openness           own=0.69
        "I am not interested in abstract ideas."
    A23          Agreeableness      own=0.71   <-- higher on Extraversion (0.71)
        "I insult people."
    A25          Agreeableness      own=0.71
        "I am not interested in other people's problems."
    [list abridged]
\end{verbatim}
}

In the worked example the constructs are the inventory's five domains.
All four items at the top of the centroid list are reverse-keyed. Two of
them are more similar to another domain than to their own, N12 clearly
and A23 only by \valAnchorATwThreeGap. N12 (``I am
relaxed most of the time.''), for example, has a cosine of
\valAnchorNTwRival{} with the Extraversion centroid and \valAnchorNTwOwn{}
with the Neuroticism centroid. This means that its wording resembles the Extraversion items slightly
more than the Neuroticism items it was written to join. Because the
printout lists only the eight lowest items, the script also lists every
item whose similarity to another domain's centroid (\texttt{rival})
exceeds its similarity to its own (\texttt{own}):

{\scriptsize
\begin{verbatim}
> closer_other
   item   own rival
3    E3 0.792 0.803
4    E4 0.733 0.751
12  N12 0.689 0.724
23  A23 0.708 0.714
30  A30 0.774 0.807
49  O49 0.732 0.736
\end{verbatim}
}

\noindent Across all 50 items, \valAnchorCloserN{} were more similar to
another domain's centroid than to their own. Every such margin was small compared with the
positive similarity that all items share, so these items are candidates
for review rather than misplaced items. Comparing items with the
construct names instead, embedded with the same 8B model, separated the
domains less clearly, because only \valAnchorOwnNameN{} of the 50 items
were most similar to the name of their own construct. A25
(``I am not interested in other people's problems.'') had the lowest
similarity to its own construct name, \valAnchorATwFiveOwn{} with
\emph{agreeableness} against \valAnchorATwFiveNeur{} with
\emph{neuroticism}, which means that its wording is no closer to the name of its own
construct than to the name of another. E6, A27, C36, N12, and C32
followed, so N12 is near the top of both lists.

The function \texttt{sfa\_project()} adapts the semantic projection of
\citet{grand-etal-2022}. For each axis, it subtracts the embedding of the
low-pole word from the embedding of the high-pole word and scores each
item by the cosine between its embedding and that difference, rescaled so
that the low-pole word itself scores 0 and the high-pole word scores 1.
Scores are therefore relative to the two pole words and can fall outside
the range from 0 to 1. Each pole here is a single word, so the scores
also depend on which words are chosen. The four pole words were
embedded with the same 8B model as the items, and the reproduction
archive stores their vectors together with those of the five construct
names. The script loads these auxiliary vectors as \texttt{aux}, uses
\texttt{aux\_vec()} to look up the vector of one word, and passes the
pole vectors to \texttt{sfa\_project()} as \texttt{poles}:

{\scriptsize
\begin{verbatim}
> AUX_NPZ   <- "data/Big5FM_aux_8B.npz"
> aux <- sfa_load_npz(AUX_NPZ, codes_key = "codes")
> aux_vec <- function(w) aux$embeddings[match(w, aux$codes), , drop = TRUE]
> poles <- list(
+   stability   = list(low = aux_vec("anxious"),  high = aux_vec("calm")),
+   sociability = list(low = aux_vec("solitary"), high = aux_vec("sociable")))
> prj <- sfa_project(fit,
+                    axes = list(stability   = c(low = "anxious",  high = "calm"),
+                                sociability = c(low = "solitary", high = "sociable")),
+                    pole_embeddings = poles)
> prj
Semantic projection onto 2 axis/axes
  Method: Grand et al. (2022)
  Scale: 0 = low pole, 1 = high pole

Axis 'stability'  [anxious <-> calm]
  lowest: N13=0.28  N11=0.33  N16=0.37  N19=0.38  N15=0.42
  highest: N12=0.82  A30=0.72  E3=0.65  E4=0.64  C37=0.62

Axis 'sociability'  [solitary <-> sociable]
  lowest: O46=0.37  A27=0.38  A21=0.38  N20=0.39  E4=0.41
  highest: E3=0.82  E7=0.81  A22=0.68  E5=0.68  A30=0.65
\end{verbatim}
}

The scores follow the item's wording and ignore the scoring key.
Reverse-keyed N12 (``I am relaxed most of the time.'') lands nearest the
calm pole (\valProjNTwCalm), opposite the anxious pole that the
forward-keyed Neuroticism items fill. This sensitivity to wording
suggests a use in scale construction, which the example illustrates but
does not test. Because the researcher
chooses the two words that define each axis, projection could show, before
anyone responds, whether an item pool covers both ends of its target
dimension, which items sit near the neutral middle, and how items from
different domains are ordered along a shared axis. The example did not
check whether an axis tracks its intended dimension or whether other
pole words would order the items in the same way. On the sociability
axis, the top five items include three Extraversion items about being
with and talking to people (E3, E7, and E5) and two Agreeableness items
about other people (A22 and A30). This mixing of the two domains at the sociable end is consistent with
the blending of Agreeableness and Extraversion in the factor solution,
although it rests on one axis and five items.

The figures so far show the similarity matrix (Figure~\ref{fig:corplot}),
the eigenvalues (Figure~\ref{fig:scree}), and the loadings
(Figure~\ref{fig:loadings}). The function \texttt{sfa\_itemplot()} adds a map of the
items themselves. It reduces each item's 4{,}096 coordinates to two, so
that items with similar embeddings are drawn near each other, using one
of four methods: $t$-distributed stochastic neighbor embedding ($t$-SNE)
\citep{vandermaaten-hinton-2008} computed with the Rtsne package
\citep{krijthe-2023}, uniform manifold approximation and projection
(UMAP) \citep{mcinnes-etal-2018} computed with the uwot package
\citep{melville-2025}, principal component analysis (PCA), or classical
multidimensional scaling (MDS). The following lines draw all four maps
and keep the coordinates of each, which \texttt{sfa\_itemplot()} returns
as \texttt{Y}, for the neighbor shares reported below:

{\scriptsize
\begin{verbatim}
> maps <- list()
> for (m in c("tsne", "umap", "pca", "mds")) {
+   maps[[m]] <- sfa_itemplot(fit, method = m, seed = 42)$Y
+ }
\end{verbatim}
}

\begin{figure}[tbp]
\centering
\captionsetup{justification=raggedright,singlelinecheck=false}
\includegraphics[width=0.7\textwidth]{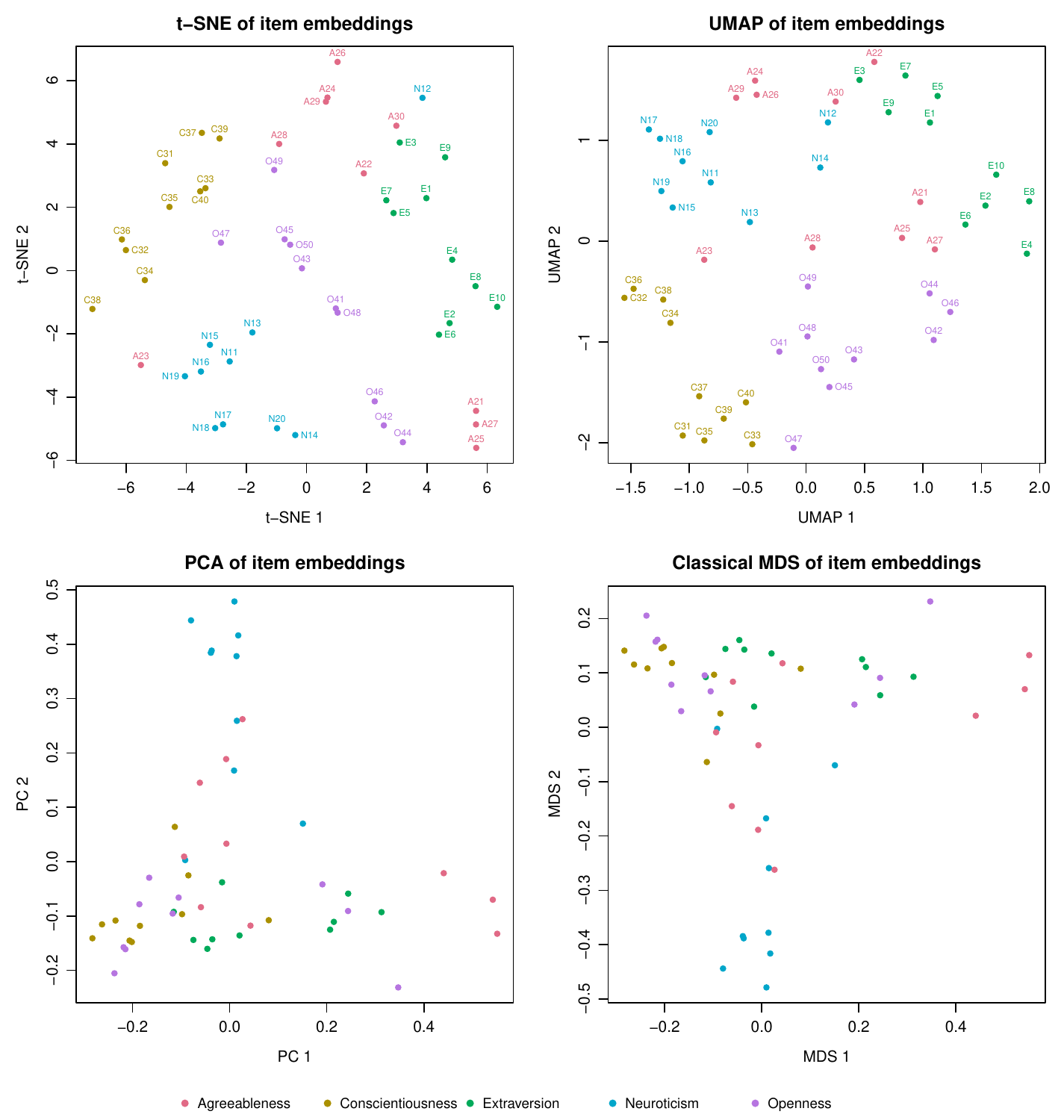}
\caption{Two-dimensional item maps of the 50 item embeddings produced by
\texttt{sfa\_itemplot()}: $t$-SNE (top left), UMAP (top right), PCA
(bottom left), and classical MDS (bottom right), with items colored by
theoretical domain in the same colors in all four panels (legend below
the panels). The maps are
computed from the Qwen3-Embedding-8B item embeddings (4{,}096
coordinates) after the default encoding, which centers each embedding
across its coordinates and scales it to unit length.
In the $t$-SNE and UMAP maps, item codes, added for this figure, give
the domain (O, C, E, A, N) and the item's position in the inventory and
are placed where they cover no other item or code. The PCA and MDS panels color
the items by domain without item codes, because the codes overprint in
those panels. $t$-SNE
= $t$-distributed stochastic neighbor embedding, UMAP = uniform manifold
approximation and projection, PCA = principal component analysis, MDS =
multidimensional scaling.}
\label{fig:itemmap}
\end{figure}

Most Neuroticism items are set apart from the other domains in the two
nonlinear maps (Figure~\ref{fig:itemmap}), although a few, such as N12
in the $t$-SNE map, lie among the items of other domains. In the two
linear maps they form a band along the second axis, with a few inside
the main cluster. The two nonlinear maps in the top row
emphasize local neighborhoods, so items with similar embeddings tend to
appear next to each other. In them, the Conscientiousness
items occupy their own region, as do most Openness items, and the
Extraversion items form a group. Several Agreeableness items lie next to
that group, and the rest are spread across the map rather than forming a
group of their own. The two linear maps in the bottom row preserve the
overall distances between items instead. The first two principal
components, the axes of the PCA map, account for \valPcaVarOne\% and
\valPcaVarTwo\% of the variance of the embeddings. Classical MDS on Euclidean
distances gives the same configuration as PCA up to reflection, so the
two panels are mirror images of one map rather than two separate views.
In that map the items of the four domains other than Neuroticism
overlap. The maps thus show, in two dimensions, a
blending of Agreeableness and Extraversion similar to the one in the
two blended factors of the semantic solution, \valMrA{} and \valMrE{}
(Table~\ref{tab:assignment}).
The axes of the $t$-SNE and UMAP maps have no interpretable scale, and
nonlinear maps can distort larger distances. \citet{mcinnes-etal-2018}
also advise care with UMAP for data sets of fewer than 500 points,
because its approximations can then give suboptimal maps, and they note
that it can find apparent structure in noise. The $t$-SNE and UMAP maps
are therefore for visual inspection only.
The share of each item's five nearest neighbors that came from its own
domain, averaged over the 50 items, puts these impressions in numbers.
It was \valPurityTsne{} in the $t$-SNE map, \valPurityUmap{} in the
UMAP map, and \valPurityLin{} in the linear map, against
\valPurityChance{} expected by chance, because nine of the other 49
items share an item's domain. The nonlinear maps are built to keep
each item's nearest neighbors near it, so their higher shares partly
reflect that design. In the linear map the Neuroticism items had the
highest share of any domain, \valPurityNLin, against \valPurityNTsne{}
and \valPurityNUmap{} in the $t$-SNE and UMAP maps.

The function \texttt{sfa\_congruence()} measures how closely the semantic factors
reproduce the theoretical assignment of items to the Big Five. It reports
two agreement metrics here, after assigning each item to the factor on
which its absolute loading is largest. The normalized mutual information
(NMI) is the information that two groupings share, divided by the
geometric mean of the entropies of the two groupings
\citep{strehl-ghosh-2002}. The entropy of a grouping is the information
that the grouping carries on its own. NMI therefore runs from 0 (nothing
shared) to 1 (identical groupings). The adjusted Rand index
(ARI) is based on the proportion of item pairs that both groupings treat
the same way, either together or apart, corrected so that 0 means chance
agreement and 1 means identical grouping \citep{hubert-arabie-1985}. The
call compares the semantic factors with the theoretical domains:

{\scriptsize
\begin{verbatim}
> cong_theory <- sfa_congruence(fit, target = big5$factors,
+                               metrics = c("nmi", "ari"))
> cong_theory
Factor structure congruence

  NMI:            0.527 (moderate - higher is better)
  ARI:            0.402 (weak - higher is better)
\end{verbatim}
}

The NMI of \valNmiTheory{} means that the semantic factors and the
theoretical domains share about half their information. To show what
chance would give, the script repeated the comparison 1,000 times with
the domain labels randomly permuted across the items. The median, 95th percentile, and maximum of the permuted values were

{\scriptsize
\begin{verbatim}
> round(apply(perm_na, 2, quantile, probs = c(.50, .95, 1)), 3)
       nmi    ari
50%  0.117 -0.004
95%  0.191  0.048
100% 0.292  0.149
\end{verbatim}
}

\noindent No permutation reached an NMI above \valNmiNullMax{} or an ARI
above \valAriNullMax, so the observed agreement is well above chance. It is also far from
complete, mainly because of how the Agreeableness and Extraversion items
are grouped (Table~\ref{tab:assignment}). The labels \emph{moderate}
and \emph{weak} in the printout come from a rule of thumb in the package
that has not been validated.

The function \texttt{sfa\_jinglejangle()} checks whether scales with similar names have
similar content \citep{wulff-mata-2025,wulff-mata-2026}. It represents each scale by two vectors, the average of its raw item
embeddings and the embedding of its name, and for every pair of scales it compares the
cosine between their item averages with the cosine between their names.
A pair is flagged when the two cosines differ by 0.20 or more. Here the
five 10-item subscales were treated as separate scales. In the call,
\texttt{sub\_items} holds the items of each subscale, \texttt{sub\_emb}
their 8B embeddings, and \texttt{jj\_lab\_emb} the 8B embeddings of the
five construct names:

{\scriptsize
\begin{verbatim}
> sub_items <- split(big5$items, big5$factors)
> sub_emb   <- lapply(split(seq_len(50), big5$factors),
+                     function(i) E8[i, , drop = FALSE])
> jj_lab_emb <- rbind(
+   Agreeableness     = aux_vec("agreeableness"),
+   Conscientiousness = aux_vec("conscientiousness"),
+   Extraversion      = aux_vec("extraversion"),
+   Neuroticism       = aux_vec("neuroticism"),
+   Openness          = aux_vec("openness"))[names(sub_items), ]
> jj <- sfa_jinglejangle(sub_items, item_embeddings = sub_emb,
+                        label_embeddings = jj_lab_emb)
> jj
Jingle-jangle detection across 5 scales
  Method: Wulff & Mata (2025, 2026)
  Flag threshold |content - label| >= 0.20

  No jingle/jangle pairs flagged.
> jj$flags
             scale_a           scale_b content_sim label_sim divergence type
2      Agreeableness      Extraversion       0.885     0.720      0.165    -
4      Agreeableness       Neuroticism       0.849     0.715      0.134    -
3  Conscientiousness      Extraversion       0.833     0.717      0.116    -
10       Neuroticism          Openness       0.807     0.703      0.104    -
6       Extraversion       Neuroticism       0.821     0.729      0.092    -
5  Conscientiousness       Neuroticism       0.822     0.733      0.088    -
9       Extraversion          Openness       0.831     0.750      0.080    -
8  Conscientiousness          Openness       0.841     0.762      0.079    -
7      Agreeableness          Openness       0.825     0.812      0.013    -
1      Agreeableness Conscientiousness       0.822     0.815      0.007    -
\end{verbatim}
}

No pair of subscales was flagged, which means that for every pair the
similarity of the construct names and the similarity of the items differed
by less than 0.20. The largest divergence, \valJjAEDiv, was for
Agreeableness and Extraversion, the two domains whose items the factor
solution blended. Their item averages had a cosine of \valJjAEContent,
but the divergence still fell short of the threshold. Because the raw cosines between these
embeddings are all strongly positive, both similarities lie near the top
of their range, which makes a divergence of 0.20 hard to reach. A null
result on five subscales of one inventory therefore says little about
how sensitive the screen is. We see the tool's main use in comparing
scales, including scales from different instruments, where mismatches
between labels and content can arise \citep{wulff-mata-2025}.

\subsection{Refining the Items}

Three functions refine a scale before any respondent is recruited.
The function \texttt{sfa\_redundancy()} detects redundant item pairs with Unique
Variable Analysis \citep{christensen-etal-2023}. It first builds a
network from the similarity matrix in which each pair of items is linked
by its partial correlation, the association between the two items that
remains after all other items are accounted for, and in which weak links
are set to zero. It then computes the weighted topological overlap (wTO)
of every pair:

{\scriptsize
\begin{verbatim}
> red_uva <- sfa_redundancy(fit)                     # UVA / wTO (EGAnet)
> red_uva
Redundant-item detection
  Method: Christensen et al. (2023)
  Measure: wto | threshold: 0.25
  Redundant pairs: 11 | redundant clusters: 10

  Top redundant pairs:
    A24        ~ A29         overlap=0.462
    N17        ~ N18         overlap=0.402
    O45        ~ O50         overlap=0.374
    N16        ~ N19         overlap=0.361
    E2         ~ E6          overlap=0.339
    A25        ~ A27         overlap=0.331
    A21        ~ A27         overlap=0.320
    C32        ~ C36         overlap=0.309
    C33        ~ C40         overlap=0.286
    O41        ~ O48         overlap=0.268

  Suggested removals (keep one per cluster): E6, N19, N18, A25, A27, A29,
    C36, C40, O41, O44, O45
> red_uva$pairs                                      # every flagged pair
   item_i item_j overlap
   [ten rows omitted: the pairs listed above]
9     O42    O44   0.253
\end{verbatim}
}

The printout lists the pairs from the largest overlap to the smallest
but keeps the row numbers they had before sorting. Each flagged pair
consists of two items whose relations to the rest of the scale overlap
strongly. The overlap column is the pair's wTO, the degree to which the two items'
connections, including their direct link and their links to the other 48
items in the network, coincide, running from 0 (no shared pattern) to 1
(identical patterns). Pairs at or above 0.25, the cutoff of Unique Variable
Analysis, are flagged. The top pair, A24 (``I sympathize with others'
feelings.'') with A29 (``I feel others' emotions.''), overlaps at
\valWtoTop. This means that the two items relate to the rest of the scale
in similar ways in their wording, so either one is a candidate to drop.
Other pairs near the top of the list are also near-paraphrases, such as
N17 (``I change my mood a lot.'') and N18 (``I have frequent mood
swings.''), and O45 (``I have excellent ideas.'') and O50 (``I am full of
ideas.''). Every one of the \valUvaPairsWord{} flagged pairs, including O42 with O44,
which the first printout leaves out, joins items from one domain, so the redundancy is repetition inside domains rather
than confusion between them. That pair is also the weakest flagged
pair and lies just above the cutoff, so the number of flagged pairs
depends on exactly where the cutoff falls. The number of flagged pairs
did not depend on the nominal sample size with which the network is estimated (see Method):
with 500 or 2,000 in place of 1,000, \valUvaPairsWord{} pairs again
reached the cutoff. The flagged pairs form
\valUvaClustersWord{} redundant clusters, each a set of items linked by
flagged pairs, and the suggested-removals line names every item to drop
so that one item of each cluster remains, the one with the highest mean
absolute similarity to all the items. A simpler criterion, which flags any pair whose similarity in the fitted
matrix is 0.85 or higher (here matching the raw cosine), found
\valCosPairsWord{} pairs in \valCosClustersWord{} clusters. Its top pairs
were N16 (``I get upset easily.'') with N19 (``I get irritated easily.'',
\valCosTopA) and A24 (``I sympathize with others' feelings.'') with A29
(``I feel others' emotions.'', \valCosTopB), and both of these pairs were also flagged by Unique Variable Analysis.
In all, \valCosInUvaWord{} of the \valCosPairsWord{} pairs that the
cosine criterion found were also flagged by Unique Variable Analysis.
Near-paraphrase pairs inflate internal consistency without
broadening content coverage \citep{christensen-etal-2023}.

The function \texttt{sfa\_simplify()} builds a candidate short form without any
responses \citep{jung-seo-2025,wang-etal-2026}. Within each domain, it
keeps the items most similar to the centroid of the domain's other items,
here five per domain:

{\scriptsize
\begin{verbatim}
> short <- sfa_simplify(fit, target_n = 5, method = "anchor")   # 50 -> 25 items
> short
Scale simplification (response-free)
  Method: centroid/medoid item selection (in the spirit of Wang et al. 2026;
    Jung & Seo 2025)
  Note: candidate short form -- validate psychometrically before use
  Selection: anchor | groups: theoretical | target_n = 5 per group
  Items: 50 -> 25
  Factors retained (parallel analysis): 5 -> 3
  Structure recovery vs theory (NMI / ARI):
    full:    NMI=0.527  ARI=0.402
    reduced: NMI=0.810  ARI=0.571
  Items kept per construct:
    [counts omitted: five per construct]

  Dropped 25 item(s); see $drop. Reduced fit in $reduced_fit.
> short$keep
 [1] "E2"  "E3"  "E5"  "E8"  "E10" "N11" "N15" "N16" "N17" "N19" "A21" "A22"
[13] "A24" "A28" "A29" "C33" "C34" "C35" "C37" "C40" "O42" "O43" "O45" "O48"
[25] "O50"
> L_sf <- unclass(short$reduced_fit$loadings)
> table(factor = colnames(L_sf)[apply(abs(L_sf), 1, which.max)],
+       domain = short$reduced_fit$item_data$factor)
      domain
factor Agreeableness Conscientiousness Extraversion Neuroticism Openness
   MR1             0                 5            0           0        5
   MR2             0                 0            0           5        0
   MR3             5                 0            5           0        0
\end{verbatim}
}

In the 25-item short form, the factors agreed more closely with the
theoretical domains, with NMI rising from \valSfNmiFull{} to
\valSfNmiRed{} and ARI from \valSfAriFull{} to \valSfAriRed. This change
is descriptive, because it compares the five factors of the full form
with the three factors retained for the short form. Two further
comparisons, in which every item set was fitted with five factors, show
where the gain comes from. First, \valSfRandN{} random sets of five items per domain reached
a median NMI of \valSfRandNmiMed{} (95th percentile \valSfRandNmiHi),
already above the full form's \valSfNmiFull, and a median ARI of
\valSfRandAriMed{} (95th percentile \valSfRandAriHi), slightly below the
full form's \valSfAriFull. Part of the gain in NMI therefore comes from
having fewer items per domain. Second, the short form itself, fitted
with five factors, reached NMI = \valSfFiveNmi{} and ARI =
\valSfFiveAri. Both values were higher than in any of the random sets (maximum NMI
\valSfRandNmiMax, maximum ARI \valSfRandAriMax). This comparison with the random sets measures the
effect of selection, and it shows that selection by centrality improved
agreement with the domains beyond what a random choice of five items per
domain gives. Parallel analysis, which had given five factors for the full inventory,
kept only \valSfKRedWord{} factors for the short form, which means that
with five items per domain the similarities no longer cleared its
threshold for one factor per domain. Parallel analysis is a weak
criterion for embedding matrices (see the Technical Reference), so this
is one run of one criterion rather than a general verdict on the short
form. The table at the end of the excerpt assigns each item to the
factor on which it loads most strongly. Neuroticism kept a factor of its
own, while the Conscientiousness and Openness items shared one factor and
the Agreeableness and Extraversion items shared another. Because each of
these three factors holds whole domains, with every item on the same
factor as the rest of its domain, the three-factor NMI of
\valSfNmiRed{} is the value that any merge of two pairs of domains into
single factors gives. The items flagged as redundant were not removed
before the short form was built, and the short form kept both items of
\valSfRedKeptWord{} of the flagged pairs, A24 and A29 among them,
because its rule selects
central items rather than distinct ones. As the printed note warns, a semantic short form
is therefore a candidate for psychometric validation, not a finished
instrument.

The function \texttt{sfa\_item\_fit()} extends the anchoring approach of
\citet{wulff-mata-2025} to newly written items. It embeds each candidate and computes its cosine with the centroid of
every construct's raw item embeddings. The report gives
the construct with the highest cosine, the margin over the next
construct, how the candidate's cosine compares with the average cosine of
that construct's existing items, and the most similar existing item. So
that the candidates are compared with the items in the same 8B space, the
reproduction archive stores their Qwen3-Embedding-8B vectors, generated
with the same settings as the item vectors. Before the call below, the
reproduction script places these vectors, and the archived 8B vectors of
the five construct names, in the package's embedding cache, so the call
reads them from the cache instead of running the model. Without these
cache entries, the same call embeds the candidates and the construct
names with the model. The reference fit, \texttt{fit\_ns}, is the main five-factor fit
rebuilt without the scoring column. Because the keying-free encoding
ignores that column, \texttt{fit\_ns} is equivalent to the main fit, and
\texttt{sfa\_item\_fit()} in any case uses only the item embeddings,
their theoretical domains, and the construct-name vectors:

{\scriptsize
\begin{verbatim}
> fit_ns <- sfa(big5_df[, c("code", "item", "factor")],
+               embeddings = E8,
+               encoding = "mean_centered_pearson", nfactors = 5)
> cand <- c("I make friends easily.",
+           "I am the life of every party.",
+           "I finish what I start.")
> ifit <- sfa_item_fit(fit_ns, cand, model = "Qwen/Qwen3-Embedding-8B",
+                      redundancy_cutoff = 0.85)
> ifit
Candidate item fit
  Method: Wulff & Mata (2025), extended

Candidate 1: "I make friends easily."
  [similarity-by-construct table omitted]
  Best construct : Extraversion
  Cross-loading  : gap to 2nd (Agreeableness) = 0.04  ->  ambiguous -
    cross-loads with Agreeableness
  Strength       : about typical for Extraversion (items 0.73 vs avg 0.77)
  Redundancy     : closest item E3 "I feel comfortable around people."
    (0.81) -> distinct
  Verdict        : cross-loads (Extraversion vs Agreeableness) - poor
    discrimination

Candidate 2: "I am the life of every party."
  [similarity-by-construct table omitted]
  Best construct : Extraversion
  [cross-loading and strength lines omitted]
  Redundancy     : closest item E1 "I am the life of the party." (0.95)
    -> near-duplicate
  Verdict        : redundant (near-duplicate of an existing item)

Candidate 3: "I finish what I start."
  [similarity-by-construct table omitted]
  Best construct : Conscientiousness
  Cross-loading  : gap to 2nd (Openness) = 0.09  ->  some cross-loading
    risk
  Strength       : about typical for Conscientiousness (items 0.82 vs
    avg 0.78)
  Redundancy     : closest item C35 "I get chores done right away."
    (0.75) -> distinct
  Verdict        : good fit for Conscientiousness - clean and
    non-redundant
\end{verbatim}
}

The three candidates, written to illustrate the three verdicts, received
the verdicts they were written for, which shows that the tool flags what
it was designed to flag rather than how often it is right. The first,
``I make friends easily.'', was more similar to the Extraversion centroid
than to the Agreeableness centroid by only \valIfGap, so it would not
distinguish the two domains. The second, ``I am the life of every
party.'', had a cosine of \valIfNearest{} with E1 (``I am the life of the
party.''), above the 0.85 cutoff, which marks it as a near-duplicate of
an existing item. The third, ``I finish what I start.'', fit Conscientiousness. Its
similarity to the Conscientiousness centroid was about as high as the
average for the domain's existing items, it exceeded its similarity to
the next construct by \valIfGoodGap, and no existing item was a
near-duplicate of it. A margin of this size falls in the band that the
printout labels \emph{some cross-loading risk}, but it is above the
0.07 cutoff for a cross-loader, so the overall verdict is a clean fit.

\subsection{Comparing With Human Responses}

To compare the semantic data to the human benchmark, we load the
responses from the Open-Source Psychometrics Project
\citep{openpsychometrics-2018}, drop every row with a missing response
or a response outside the 1--5 range (keeping \valNHuman{} of
\valNHumanRaw{} response records), apply reverse scoring to the \valNRev{}
reverse-keyed items ($6 - x$), and run a conventional factor analysis:

{\scriptsize
\begin{verbatim}
> HUMAN_CSV <- Sys.getenv("BIG5_HUMAN_CSV", "data/Big5FM_data.csv")
> human_raw <- read.csv(HUMAN_CSV)
> keep <- rowSums(human_raw < 1 | human_raw > 5 | is.na(human_raw)) == 0
> human <- human_raw[keep, ]
> n_human <- nrow(human)
> cat("Respondents after cleaning:", n_human, "of", nrow(human_raw), "\n")
Respondents after cleaning: 874434 of 1015341
> human_keyed <- human
> rev_idx <- which(big5$scoring < 0)
> human_keyed[, rev_idx] <- 6 - human_keyed[, rev_idx]
> R_human       <- cor(human_keyed)
> human_fa <- psych::fa(R_human, nfactors = 5, n.obs = n_human,
+                       rotate = "oblimin", fm = "minres")
There were 48 warnings (use warnings() to see them)
\end{verbatim}
}

The warnings report that the noncentral chi-square probabilities, which
psych computes for its significance tests of fit, did not converge at
this sample size. They do not affect the loadings. Five factors were fixed at the theoretical five-factor structure
\citep{goldberg-1990, goldberg-1992}. A conventional parallel analysis
of these correlations, computed with \texttt{psych::fa.parallel()},
suggested \valHumanPaKWord{} instead. This analysis compares each
common-factor eigenvalue with the mean of the corresponding eigenvalues
of random data of the same size, following \citet{horn-1965} but with
common-factor eigenvalues:

{\scriptsize
\begin{verbatim}
> human_pa <- psych::fa.parallel(R_human, n.obs = n_human, fa = "fa",
+                                plot = FALSE)
Parallel analysis suggests that the number of factors =  10  and the number
  of components =  NA
> round(rbind(observed = human_pa$fa.values[1:12],
+             random   = human_pa$fa.sim[1:12]), 3)
          [,1]  [,2]  [,3]  [,4]  [,5]  [,6]  [,7]  [,8]  [,9] [,10] [,11]
observed 6.858 4.142 3.112 2.737 1.992 0.617 0.386 0.175 0.082 0.051 0.003
random   0.015 0.013 0.012 0.012 0.011 0.010 0.010 0.009 0.008 0.008 0.007
          [,12]
observed -0.028
random    0.007
\end{verbatim}
}

\noindent At $N = \valNHuman$ the random-data eigenvalues are almost
exactly zero. They come from simulations that \texttt{psych} runs in
parallel processes and that the script does not seed separately, so their
last digits can differ slightly from one run
of the script to the next, although the suggested number of factors did
not change across our runs. Even very small factors exceed them, so the
\valHumanPaKWord{} factors may reflect the very large sample rather than
\valHumanPaKWord{} substantive dimensions, although this explanation was
not tested here. The five-factor solution
reproduced the Big Five with simple structure, which means that each item
loads strongly on one factor and weakly on the others. Here, every item's
largest loading was on the factor of its own domain
(Table~\ref{tab:s-human-loadings}).

\subsubsection{Factor-Level Agreement}

The function \texttt{sfa\_congruence()} takes a fitted \texttt{psych} model as the
target:

{\scriptsize
\begin{verbatim}
> cong_human <- sfa_congruence(fit, target = human_fa,
+                              metrics = c("tucker", "nmi", "ari", "frobenius"))
> cong_human
Factor structure congruence

Tucker phi (factor-by-factor):
  [5 x 5 matrix omitted: reported in the Tucker congruence table]

  NMI:            0.527 (moderate - higher is better)
  ARI:            0.402 (weak - higher is better)
  Frobenius:      0.664 (moderate - higher is better)
\end{verbatim}
}

Tucker's congruence coefficient ($\phi$), which \citet{burt-1948}
described for comparing two columns of coefficients and which later
became known by Tucker's name \citep{lorenzo-seva-tenberge-2006}, measures how
similar the loading patterns of two factors are. It runs from $-1$ to 1, and values near 1
mean that the same items load on both factors in nearly the same
proportions \citep{lorenzo-seva-tenberge-2006}. By the benchmarks of
\citet{lorenzo-seva-tenberge-2006}, values from .85 to .94 indicate fair
similarity, and values above .95 indicate good similarity, at which two
factors can be considered equal. The benchmarks rest on averaged ratings
by experienced factor analysts who compared loading patterns on 15 or 20
variables. The individual ratings varied widely and differed between the
two lengths. For pairs of low congruence they also differed between
research fields, with researchers in personality judging the factors to
be more different than researchers in intelligence did. The benchmarks
are therefore only approximate for the 50-item patterns compared here
\citep{lorenzo-seva-tenberge-2006}. Each semantic factor matched a different human factor
(Table~\ref{tab:tucker}). The factor matched to Openness reached $\phi =
\valPhiO$, within the range of fair similarity, and the factors matched
to Conscientiousness and Neuroticism reached \valPhiC{} and \valPhiN,
below it. The factors matched to Extraversion and Agreeableness
reached only \valPhiE{} and \valPhiA. The semantic Extraversion factor
also matched human Agreeableness (\valPhiEA) as closely as the semantic
Agreeableness factor did, which fits its mixture of sociability and
warmth items (Table~\ref{tab:assignment}). The mean matched congruence,
each semantic factor's largest absolute congruence with any human factor
averaged over the five factors, was $\bar{\phi} = \valPhiMean$, the
average of three closer matches and two weak ones. The NMI and ARI
between the semantic and human groupings equal the values reported above
for the theoretical assignment (\valNmiTheory{} and \valAriTheory),
because every item's largest human loading falls on its own domain's
factor, so the human grouping is the theoretical one. These two metrics
therefore add nothing to that earlier comparison. The Frobenius similarity (\valFrob) is the cosine between the two
solutions' factor-correlation matrices, each written out as a single
list of numbers. It does not first match the factors, so its
value depends on the arbitrary order in which each solution lists them,
and it is not interpreted here.

\begin{table}[tbp]
\centering
\captionsetup{justification=raggedright,singlelinecheck=false}
\caption{Tucker congruence coefficients between the semantic factors
(rows) and the human-response factors (columns)}
\label{tab:tucker}
\tighttable
\begin{tabular}{lrrrrr}
\toprule
& \multicolumn{5}{c}{Human factor} \\
\cmidrule(lr){2-6}
Semantic factor & O (MR5) & C (MR2) & E (MR4) & A (MR3) & N (MR1) \\
\midrule
O (MR4) & \textbf{.92} & .14 & .18 & .17 & .02 \\
C (MR1) & .28 & \textbf{.84} & .27 & .10 & .11 \\
E (MR5) & .16 & .11 & \textbf{.64} & .52 & .07 \\
A (MR2) & .27 & .05 & .41 & \textbf{.52} & $-$.01 \\
N (MR3) & .12 & .13 & .03 & .30 & \textbf{.80} \\
\bottomrule
\end{tabular}

\caption*{\textit{Note}. O = Openness, C = Conscientiousness, E =
Extraversion, A = Agreeableness, and N = Neuroticism. Each cell is
Tucker's congruence coefficient $\phi$ between a semantic factor's and a
human factor's loadings on the 50 items, which runs from $-1$ to 1. MR
labels are each solution's automatic factor names (MR = minimum
residual), and the domain beside each label is the one on whose items the
factor's mean absolute loading is largest. Bold values mark each semantic
factor's largest absolute congruence with any human factor, and their
mean is the mean matched $\bar{\phi} = \valPhiMean$. Semantic solution:
minimum-residual extraction with oblimin rotation, applied to the
mean-centered Pearson similarity matrix. Human solution: the same method
applied to the correlations of the reverse-scored responses, $N =
\valNHuman$.}
\end{table}

\subsubsection{Encoding-Level Agreement}

Fitting a five-factor model under each of the four embedding encodings
and to the NLI matrix gives the comparison in Table~\ref{tab:s-encodings}
of the Supplementary Materials. Each is compared with the human
correlations that match it. The keying-free encodings and the NLI matrix
are compared with the correlations among raw responses. The
sign-flipping encoding applies the scoring key, so it is also compared
with the correlations of the reverse-scored responses, which apply the
same key. Sign-flipping left the grouping of the items unchanged (its NMI
and ARI equal those of \texttt{atomic}). Against the raw responses, it
gave no gain ($r = \valArRRaw$ for the flip, against \valMcpRRaw{} for
the default encoding). Against the reverse-scored responses, its
agreement was lower than that of \texttt{atomic}, which applies no flip
($r = \valArRKeyed$, against $r = \valAtRKeyed$), and so was its mean
matched congruence with the human factors. A likely source of the drop in
both is the sign pattern that the flip imposes. Every unflipped
cosine is positive (the smallest is \valRangeCosMin), so the flip makes
every similarity between a forward-keyed and a reverse-keyed item
negative, whereas reverse-scoring is meant to make the items of a domain
correlate positively whatever their keying.

Against the raw responses, the NLI matrix agreed best ($r =
\valNliRRaw$), followed by the \texttt{squid} encoding ($r = \valSqRRaw$)
and by \texttt{atomic} and \texttt{mean\_centered\_pearson} ($r =
\valMcpRRaw$). The \texttt{squid} matrix is a poor basis for factor
analysis, however. Its KMO was only \valSqKmo, which means that its
correlations are small compared with its partial correlations. The KMO
describes each matrix's own factorability, and the three fit statistics
in the table (RMSR, CAF, and TEFI) describe how well each solution fits
its own matrix, so none of the four is comparable across the table's
rows. The \texttt{squid} CAF near 1, for example, follows from its KMO
near 0. The NLI matrix matched the human loading patterns worst, with the
lowest mean matched congruence ($\bar{\phi} = \valNliPhi$). This
matrix had negative eigenvalues, however, which were raised to just above
zero before factoring, as the Supplementary Materials describe. Its
factor-based columns in the table (NMI, ARI, $\bar{\phi}$, KMO, TEFI,
RMSR, and CAF) therefore come from the repaired matrix, and its rankings
on them reflect the repair. Its $r$ with the raw responses uses the
original matrix and is the cleaner NLI result. These rankings are
descriptive, from a single analysis of 50 items, and carry no estimate
of their uncertainty. Gaps of a few hundredths between encodings, such as
those among the $r$ values above, cannot be distinguished without such
estimates, and even the larger lead of the NLI matrix on the raw
responses was not tested. The default mean-centered Pearson encoding
had the highest mean matched congruence ($\bar{\phi} = \valPhiMean$, tied
with \texttt{atomic}) and produced a matrix well suited to factor analysis
(KMO = \valKmo). The table therefore does not single out one best
encoding, and it is consistent with the reasons the Method gives for the
default. On these embeddings the default and \texttt{atomic} gave the same
values to three decimals on every agreement metric, so the comparison
cannot separate them.

\subsubsection{Item-Pair Agreement}

\begin{figure}[tbp]
\centering
\captionsetup{justification=raggedright,singlelinecheck=false}
\includegraphics[width=0.6\textwidth]{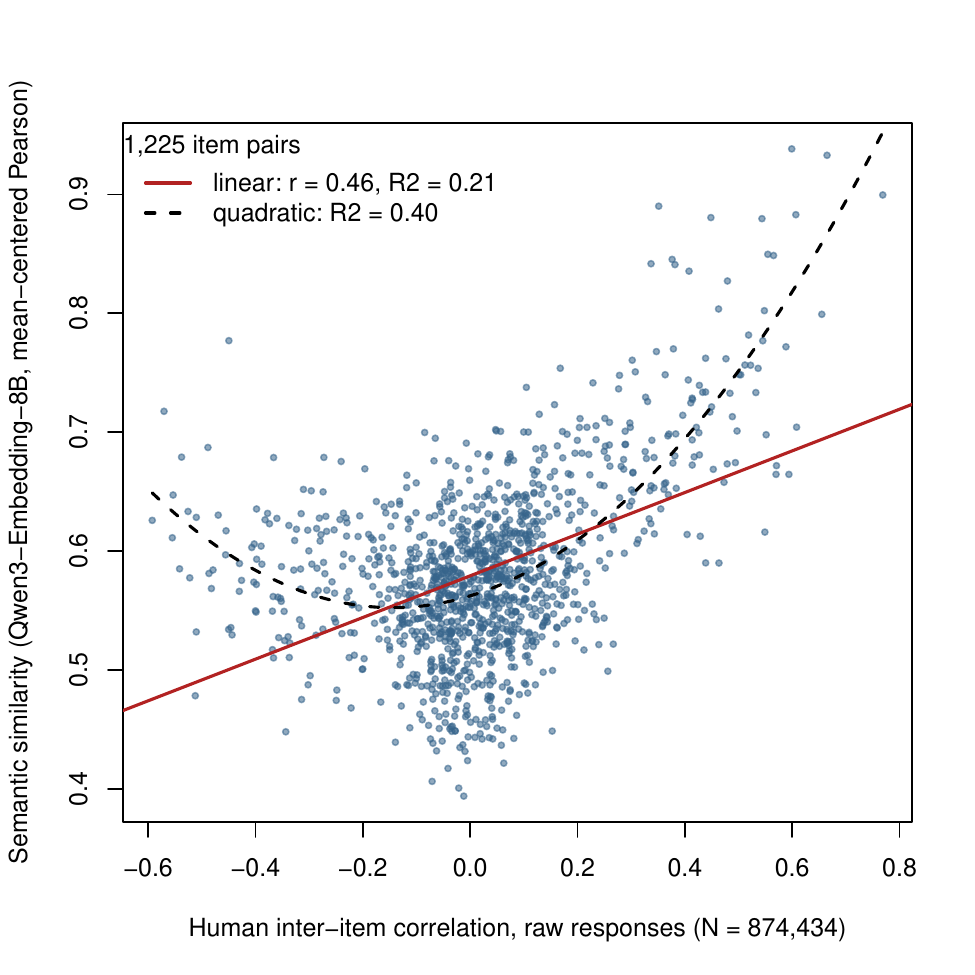}
\caption{Semantic similarity (mean-centered Pearson encoding of the
Qwen3-Embedding-8B item vectors) against human inter-item correlations
from raw responses ($N = \valNHuman$), for all \valNPairs{} item pairs.
Each point is one pair of items. The red line is the linear fit ($r =
\valPairR$). The dashed black curve is the quadratic fit, which accounts
for \valPairRsqQuad{} of the variance in the semantic similarities,
against \valPairRsqLin{} for the line. Similarity rises steeply toward
the positive end of the horizontal axis and only slightly toward the
negative end.}
\label{fig:semantic-human}
\end{figure}

Across all \valNPairs{} item pairs, the mean-centered Pearson similarities
correlated $r = \valPairR$ with the correlations among the raw responses
of \valNHuman{} respondents (Figure~\ref{fig:semantic-human}). Raw
responses are the right comparison here, because this encoding ignores
the scoring key (against the reverse-scored correlations, $r =
\valMcpRKeyed$). The correlation means that pairs of items that are more
similar in meaning tended to be answered more alike, although meaning
leaves most of the variation in the human correlations unexplained. The
\valNPairs{} pairs share 50 items, so they are not independent
observations. An item-level bootstrap, which resampled the 50 items,
gave a 95\% percentile interval for the correlation from \valPairRLo{}
to \valPairRHi.

The relation is also curved rather than straight
(Figure~\ref{fig:semantic-human}). A quadratic curve accounts for
\valPairRsqQuad{} of the variance in the semantic similarities, against
\valPairRsqLin{} for the straight line, a gain of \valPairGain{} (item-level
bootstrap 95\% interval \valPairGainLo{} to \valPairGainHi). Most of the
curvature lies at the positive end of the human-correlation axis. Mean
similarity was \valPairSimPos{} for pairs whose human correlation was
above .20, \valPairSimZero{} for pairs within .10 of zero, and
\valPairSimNeg{} for pairs below $-.20$. Similarity therefore rises
steeply as the human correlation becomes more positive and only slightly
as it becomes more negative. These cutoffs are descriptive and were
chosen after the scatterplot was seen.

Grouping the pairs by domain and keying describes the negative end
further. Pairs of items from the same domain that are keyed alike
(\valPairSameKeyN{} pairs) had a mean similarity of \valPairSameKeySim{}
and a mean raw-response correlation of \valPairSameKeyR. Pairs that join
an item and a reverse-keyed item of the same domain (\valPairOppKeyN{}
pairs) correlated negatively in the responses (mean $r =
\valPairOppKeyR$, negative for \valPairOppKeyNegPct\% of them), and their
mean similarity, \valPairOppKeySim, was about the same as that of pairs
from different domains (\valPairCrossN{} pairs, mean similarity
\valPairCrossSim, mean $r = \valPairCrossR$). The difference of \valPairOppCrossGap{} had
an item-level bootstrap 95\% interval from $\valPairOppCrossLo$ to
$\valPairOppCrossHi$, which includes zero. The embeddings thus
separated the opposite-keyed items of a domain from the items keyed
alike, by \valPairGap{} in mean similarity (item-level bootstrap 95\%
interval \valPairGapLo{} to \valPairGapHi). They placed them at the level
of unrelated items, however, not below it, which means that they
registered that the two items differ but not that they are opposed. Of
the \valPairNegN{} pairs whose human correlation was below $-.20$,
\valPairNegOppN{} joined a forward-keyed and a reverse-keyed item of the
same domain, and the other \valPairNegOtherN{} were similarly close to the
level of unrelated pairs (mean similarity \valPairSimNegOther). This
pattern is consistent with the reverse-keying problem described in the
Introduction: the embeddings did not represent the opposition that
people's answers show. The similarities also correlated $r =
\valPairRAbs$ with the absolute values of the human correlations, against
$r = \valPairR$ with the signed values, a descriptive difference without
an interval.

The worked example therefore gives the semantic--behavioral coherence
of this inventory, for this embedding model, a numerical value, a correlation of $r = \valPairR$
between semantic and human similarities across item pairs and a mean
matched congruence of $\bar{\phi} = \valPhiMean$ across factors. The
coherence is uneven across domains. The Openness, Conscientiousness, and
Neuroticism factors resemble their human counterparts most closely,
whereas the Agreeableness and Extraversion factors resemble theirs only
partly. In the semantic solution, most of the Extraversion items and
\valAsgnAOnBlend{} of the Agreeableness items were divided between two
blended factors, which also took \valAsgnBlendOther{} items from other
domains, and \valAsgnAOnN{} Agreeableness items joined the Neuroticism
factor (Table~\ref{tab:assignment}). The composition of the two blended
factors is consistent with a grouping by topic and phrasing. One gathers forward-keyed items about
sociability and warmth from both domains, and the other gathers
reverse-keyed items that describe the absence of a disposition, from
these and other domains. The responses, once reverse-scored, keep
the two domains separate. This divergence identifies the specific items
on which the wording of the scale and the responses it elicits organize
the constructs differently.

\newpage
\section{Discussion}

In this example, with the 8B embeddings, the consensus of five
retention criteria, with ties going to the smaller number, matched the number of factors the
inventory's authors intended, with two criteria choosing \valPaKWord{}
factors, two choosing \valKaiserKWord, and one choosing \valTefiKWord. Three of the five domains corresponded mainly to single
factors. Where the semantic structure and the structure of the human responses
disagreed, the disagreement suggested where to look for a source. This
Discussion first summarizes what the example found and then explains
what the agreement between the semantic and response structures means,
because the recommendations, applications, and limits that follow depend
on what that agreement can and cannot show. It then draws
recommendations for users, describes where the package is most useful,
states the limits of the method, and ends with three directions for
further research.

\subsection{Summary of Findings}

This article introduced \texttt{semanticfa}, an R package that runs the
whole factor-analytic workflow on the similarities between the
embeddings of a scale's items, and can also factor a similarity matrix
from an NLI cross-encoder, without any response data. The worked
example applied every stage and each refinement tool to the 50-item
IPIP Big-Five Factor Markers and compared the results with a factor
analysis of \valNHuman{} people's responses. It produced four main
findings.

First, the semantic analysis recovered part of the structure that the
inventory's authors intended. Two retention criteria chose five factors,
two chose six, and one chose two, and the package's rule for ties, which keeps the
smaller number, gave a consensus of \valConsensusKWord, the number of
domains. Three of the five factors corresponded mainly to one domain
each (Openness, Conscientiousness, and Neuroticism). Most of the Extraversion items and \valAsgnAOnBlend{} of the
Agreeableness items were divided between the other two, which are
blended factors. One gathered forward-keyed items about sociability and
warmth, and the other gathered reverse-keyed items that describe the
absence of a disposition. The blended factors also took
\valAsgnBlendOther{} items from other domains, and \valAsgnAOnN{}
Agreeableness items joined the Neuroticism factor
(Table~\ref{tab:assignment}). Second,
the agreement with human responses differed by domain. The semantic factors matched the human factors more closely for
Openness ($\phi = \valPhiO$, in the range of fair similarity),
Conscientiousness (\valPhiC), and Neuroticism (\valPhiN) than for
Extraversion (\valPhiE) and Agreeableness (\valPhiA). Across all
\valNPairs{} item pairs, pairs that were more similar in meaning tended
to be answered more alike, as measured by the correlations among the raw
responses ($r = \valPairR$). The relation was curved. Similarity
rose steeply among pairs that people answered alike and only slightly among
pairs answered in opposite directions. An item and a reverse-keyed item
of its domain, which people answered oppositely, were on average about
as similar as two items from different domains (\valPairOppKeySim{}
against \valPairCrossSim, a difference whose 95\% interval included zero),
so the embeddings registered that the two items
differ but not that they are opposed.

Third, two analysis choices changed the conclusions. The first was the
encoding. Sign-flipping the reverse-keyed items' embeddings, the
apparent analogue of reverse-scoring, gave no gain in agreement with the
correlations among raw responses ($r = \valArRRaw$, against
\valMcpRRaw{} for the default encoding). Because the flip applies the
scoring key, its matched comparison is with the correlations among
reverse-scored responses, and there it gave the worst agreement of the
four embedding encodings ($r = \valArRKeyed$, against \valAtRKeyed{}
without the flip). A likely source of that drop is the sign pattern that
the flip imposes on every pair of a forward-keyed and a reverse-keyed
item. The second was the reference
used to judge fit. By the cutoff conventionally applied to response
data \citep{hu-bentler-1999}, the RMSR of \valRmsr{} indicated good fit, but every fit to random
embeddings also passed that cutoff, so passing it says little about
embeddings. Fourth, the refinement tools proposed specific revisions to
the item pool. They flagged \valUvaPairsWord{} redundant item pairs and
built a 25-item short form. Its agreement with the theoretical domains
rose from NMI = \valSfNmiFull{} for the full inventory to \valSfNmiRed,
but parallel analysis on the short form kept only \valSfKRedWord{}
factors, so that change compares a five-factor with a three-factor
partition and is descriptive. Random sets of five items per domain,
fitted with five factors, already reached a median NMI of
\valSfRandNmiMed, so part of the gain comes from having fewer items per
domain. Fitted the same way, the short form reached NMI = \valSfFiveNmi{} and
ARI = \valSfFiveAri, both above every random set, which is the effect of choosing the items closest to
the center of their own domain. The tools also sorted three candidate items,
written to illustrate the three verdicts, into one that cross-loaded,
one that nearly duplicated an existing item, and one that fit
Conscientiousness without duplicating any item.

\subsection{Theoretical Implications}

The semantic and human analyses start from different data. The semantic
analysis factors the similarities between item embeddings and uses no
respondents, whereas the human analysis factors the correlations among
the answers of \valNHuman{} people. The two agreed more closely for Openness, Conscientiousness, and
Neuroticism ($\phi = \valPhiO$, \valPhiC, and \valPhiN) than for the
other two domains. Because the two analyses use different information about the same
items, that agreement is evidence, for this inventory and embedding
model, that the items' wording and people's responses group the items of
these three domains in broadly similar ways, most closely for Openness,
the only domain in the approximate range of fair similarity. The embedding model may,
however, have learned the published domain assignment from listings of
the items in its training text, which would inflate the agreement and
cannot be excluded here (see Limitations). Such agreement can be counted
among the observable relations that make up a construct's
nomological network \citep{cronbach-meehl-1955}. It is not proof that
the traits exist or cause the responses \citep{borsboom-etal-2004}.

The disagreements are also informative, because they were concentrated
in two domains rather than spread across the inventory. In the semantic
solution, most of the Extraversion items and \valAsgnAOnBlend{} of the
Agreeableness items were divided between two blended factors, which also
took \valAsgnBlendOther{} items from other domains, and \valAsgnAOnN{}
Agreeableness items joined the Neuroticism factor
(Table~\ref{tab:assignment}), although a few of these counts rest on
near ties. The composition of the blended factors is consistent with a grouping by topic and phrasing rather than by domain. One gathered forward-keyed
items about sociability and warmth from both domains, and the other
gathered reverse-keyed items that describe the absence of a disposition,
from both domains and from Openness and Neuroticism. The semantic
Extraversion factor matched human Agreeableness (\valPhiEA) as closely as
the semantic Agreeableness factor did (Table~\ref{tab:tucker}), which
fits this mixture. Two reasons are
plausible, although neither was tested here. English describes
sociability and warmth in overlapping words, and statements phrased as
the absence of something share the wording of negation across topics, whereas people's reverse-scored answers keep the domains apart.
Other studies have found divergences of this kind and offered
plausible sources for them. Two studies found the same overlap of Agreeableness and Extraversion
items in embeddings and noted that the two domains are also associated
in human responses \citep{casella-etal-2024,milano-etal-2025a}. The wording may
thus draw two related dimensions closer together than people's answers
do, which is consistent with the mixture of warmth and sociability here,
where the human solution kept the two domains apart. The connotations of a word may also shift its
position in text away from the position that other measures give it.
\citet{kmetty-etal-2021}, for example, found that health-care occupations
ranked lower in text than on a socioeconomic index of occupations,
plausibly because of their connotations of illness. People's answers may
also share causes that the wording does not carry, as when insomnia and
physical symptoms formed one factor only in responses, plausibly because
they occur together in people, perhaps through bodily or physiological
causes, without being described alike \citep{kojima-etal-2026}.
A disagreement therefore points to the items and domains where these
possible sources can be investigated. How large the divergence is may also vary with the
construct. In one unpublished study of a single sample and nine
dimensions from four scales, wording appeared to predict correlations
least well in the dimensions whose items already correlated strongly,
by the authors' reading of a random-effect correlation estimated across
those nine dimensions, in a model whose average effect of wording was
significant only narrowly ($p = .044$) \citep{zhang-qiu-2026}.

Because coherence can be estimated, for a given embedding model,
theories of a construct can be asked to predict it, and those
predictions can be tested. If some clinical
constructs are networks of symptoms that cause one another
\citep{borsboom-2008}, coherence should be lower for them than for
constructs whose symptoms share a common cause, because symptoms that
cause one another can correlate strongly in people's answers without
resembling one another in meaning. The function \texttt{sfa\_congruence()} makes that
prediction testable across instruments of both kinds. The same
reasoning yields predictions for two other cases. Formative indicators
jointly determine a construct and need not correlate
\citep{bollen-lennox-1991}, so coherence cannot be assumed for them.
Method variance, the part of the correlations among responses that comes
from the method of measurement rather than from the content measured
\citep{campbell-fiske-1959}, is a form of construct-irrelevant variance
\citep{messick-1995}. We expect, although this was not tested,
that it also lowers coherence, because embeddings register the wording
of an item but not other features of the method, such as the response
format. Embeddings do register some features of wording beyond topic,
however, such as the negated phrasing that appears to have grouped
\valAsgnARev{} of the \valNRev{} reverse-keyed items here, so a method effect carried by the wording could instead
raise coherence. Studies of the
semantic theory of survey response suggest that the wording of a scale
can shape factor analyses of its ratings
\citep{arnulf-etal-2014,arnulf-etal-2018}, but a factor analysis of
responses alone does not estimate how much of a scale's structure comes
from its wording. Coherence indexes how closely the response structure
follows the wording. It is presumably related to that contribution of
the wording but is not the same thing, and the present data do not test
the relation.
Theories of a construct can be asked to predict this index.

\subsection{Recommendations for Users}

The worked example suggests six practices for anyone running SFA. The
first five each follow from a specific finding, and the sixth follows
from the uneven agreement with human responses across domains. Where a
practice rests on the worked example, that evidence comes from one
inventory, one family of embedding models plus one NLI cross-encoder,
and English items, so each practice is a starting point that further
applications can revise.

\begin{enumerate}
\item \emph{Use a keying-free encoding, and do not sign-flip
reverse-keyed items.} Reverse-scoring a response keeps the item's
content and inverts only the scale on which it is answered. Multiplying
an embedding by $-1$, by contrast, reverses the sign of its cosine with
every item that is not also flipped, so the item becomes dissimilar even
to items on its own topic. In this inventory, sign-flipping gave no gain
in agreement with the correlations among raw responses ($r =
\valArRRaw$, against \valMcpRRaw{} for the default encoding), and it
left the grouping of the items unchanged. Because the flip applies the
scoring key, its matched comparison is with the correlations among
reverse-scored responses, which apply the same key. On that comparison
it gave the worst agreement of the four embedding encodings, both over
item pairs and in loading patterns (Table~\ref{tab:s-encodings} in the
Supplementary Materials). A likely source of the drop from the unflipped cosines is
the sign pattern that the flip imposes. After reverse-scoring, a forward-keyed
and a reverse-keyed item of the same domain correlate positively,
whereas the flip gives such pairs an average similarity of
$\valArOppMean$. In earlier work, which flipped item embeddings before
averaging them within each facet, sign-flipping raised factor recovery
for one inventory. For the HEXACO, however, it recovered the factors
less often than the other encodings tested and consistently failed to
recover the Emotionality factor \citep{guenole-etal-2025}. Embeddings from
general-purpose models do not represent reverse-keyed items as the
opposites of the forward-keyed items they contradict
\citep{hommel-arslan-2025,suarez-alvarez-etal-2026}, raw embeddings rarely
give negative similarities between opposing items
\citep{pellert-etal-2026}, and text methods older than embeddings could
not detect that an item is reversed either, so studies using them
corrected the signs of reverse-scored items by hand
\citep{arnulf-etal-2018} and, for scales expected to relate negatively,
assigned negative signs from theory
\citep{arnulf-etal-2014,arnulf-etal-2018}. In the worked
example, an item and a reverse-keyed item of its domain were on average
about as similar as two items from different domains (\valPairOppKeySim{}
against \valPairCrossSim, a difference whose 95\% interval included
zero), although people
answered them in opposite directions.
Three of the four embedding encodings use no scoring key: the unflipped cosines, SQuID,
and the default mean-centered Pearson encoding. SQuID agreed best with
the correlations among reverse-scored responses, but on the like-with-like
comparison with raw responses it fell below the NLI matrix (\valSqRRaw{}
against \valNliRRaw). It also left a matrix poorly suited to factor analysis
(KMO = \valSqKmo), so the default \texttt{mean\_centered\_pearson}
encoding is the better choice for factor analysis.
\item \emph{Decide the number of factors from several criteria, and check
it against theory.} The criteria disagreed here, with \valPaKWord{} and
\valKaiserKWord{} each chosen by two criteria and \valTefiKWord{} chosen
by TEFI. In simulations with personality items written by a language
model (GPT-4o) and embedded with a single embedding model, the Kaiser
rule and parallel analysis chose too many factors more often than EGA
\citep{garrido-etal-2025}, although here the adapted parallel analysis
agreed with EGA. The human analysis had a different problem. With
\valNHuman{} respondents, conventional parallel analysis endorsed
\valHumanPaKWord{} factors, possibly because with that many respondents
even very small factors exceed the mean random-data eigenvalues, although
this explanation was not tested. Reporting the consensus, and fixing the final number at the theoretical
five \citep{goldberg-1990} rather than letting a single criterion decide,
avoids relying on any one criterion. The consensus alone would not have
been enough here, because it reached five only with the 8B embeddings,
and only through its tie rule.
\item \emph{Compare fit statistics with fits to random embeddings.} By
the cutoff conventionally applied to response data
\citep{hu-bentler-1999}, the RMSR of \valRmsr{} indicated good fit, but every fit to random embeddings also
passed that cutoff. Without a sample size, many of the usual tests of fit cannot be
computed \citep{suarez-alvarez-etal-2026}. Random-embedding baselines are a
cruder relative of the random-word baseline that \citet{pokropek-2026}
used for confirmatory models, because random vectors lack the general
similarity that real items share. They are therefore a weak reference
(see Limitations), but they show which conventional readings do not
carry over to embeddings.
\item \emph{Prefer the largest model that memory allows, and check how
the number of factors depends on model size.} All three model sizes placed
items from the same domain closer together than items from different
domains, and for this inventory the separation did not grow with size.
For a second inventory in the Supplementary Materials, the separation
rose with size, although its paired interval just included zero, and
that inventory's factors matched its scales more closely with each
larger model. Only with the 8B
model, however, did parallel analysis and EGA both indicate the
theoretical five factors, while the Kaiser rule, the empirical Kaiser
criterion, and TEFI did not change with size. Because the comparison
covered three sizes of one model family on two English inventories, and
size was confounded with embedding dimension, this recommendation may
not hold for other families or inventories.
\item \emph{Report agreement with human data on several metrics, and
compare each encoding with the matching human correlations.} Encodings
that ignore the scoring key should be compared with the correlations
among raw responses, and encodings that use the key with the
correlations among reverse-scored responses, so that both sides of each
comparison either use the scoring key or ignore it. Because an
encoding that uses the key imposes a sign pattern of its own, its
comparison with raw responses should be reported as well, as the first
practice showed for sign-flipping. Different metrics (loading shape, item grouping, and item-pair
correlation) can rank the encodings differently. The NLI matrix, for
example, was best on item grouping and on item-pair correlation with
raw responses but worst on loading shape (Table~\ref{tab:s-encodings}
in the Supplementary Materials). Its item grouping and loading shape,
however, come from a repaired version of the matrix, in which the
negative eigenvalues were raised to just above zero, so those two
rankings partly reflect the repair. Its item-pair correlation with raw
responses uses the original matrix, so it does not depend on the repair,
although its lead there was not tested either. Gaps of a
few hundredths between encodings, such as the NLI matrix's lead in item
grouping, cannot be distinguished without intervals.
\item \emph{Treat every semantic verdict as a hypothesis for response
data to test.} Agreement with the structure of the human responses was closer for three domains ($\phi$
from \valPhiN{} to \valPhiO) and weak for two (\valPhiA{} and
\valPhiE). In the semantic solution, Agreeableness and Extraversion
items shared two blended factors (Table~\ref{tab:assignment}), but
nothing in that solution could show whether people's responses would
separate the two domains. Extending the same
caution to a redundancy flag, a short form, or a verdict on a new item
is a precaution rather than a finding of the example, because each of
these describes wording. Whether people's responses show the same
redundancy, structure, or fit cannot be assumed from wording alone.
\end{enumerate}

\subsection{Applications}

The package is intended for the early stage of scale development that
\citet{clark-watson-2019} call crucial, the writing of the item pool,
which comes before any data are collected. At that stage a researcher can screen a large pool of
candidate items, remove redundant ones early, and draft a short form,
all without collecting responses. The worked example flagged redundant
pairs, drafted a short form, and screened new candidate items. It
made no removal on the basis of the redundancy flags, however, and its
screen covered three candidate items rather than a large pool. Redundancy screening flagged \valUvaPairsWord{}
redundant pairs, the strongest of them near-paraphrases, led by A24 (``I
sympathize with others' feelings.'') and A29 (``I feel others'
emotions.''), whose wTO of \valWtoTop{} was the highest of any pair,
against a flag threshold of 0.25. Such pairs would be expected to raise
internal consistency without adding content \citep{christensen-etal-2023},
although the example did not check this in the responses. The short form
halved the inventory. Its agreement with the theoretical domains cannot
be compared directly with the full form's, because parallel analysis
kept \valSfKRedWord{} factors for the short form. The rise in NMI from
\valSfNmiFull{} to \valSfNmiRed{} therefore compares a partition of the
items into five factors with one into \valSfKRedWord{} and is
descriptive only. The informative comparison is with random item sets.
\valSfRandN{} random sets of five items per domain, fitted with five
factors, already reached a median NMI of \valSfRandNmiMed, above the
full form's \valSfNmiFull, so part of the gain comes from having fewer
items per domain. Choosing the items closest to the center of their own
domain improved agreement further. Fitted the same way, the short form
reached an NMI of \valSfFiveNmi{} and an ARI of \valSfFiveAri, above
every random set on both (maxima \valSfRandNmiMax{} and
\valSfRandAriMax). The short form also kept both items of
\valSfRedKeptWord{} of the flagged redundant pairs, so building a short
form and screening for redundancy are separate steps. A short form
chosen for semantic centrality can also lose breadth that the construct
needs, in the same way that maximizing internal consistency can
\citep{clark-watson-2019}. It is therefore a candidate for validation
rather than a finished instrument \citep{jung-seo-2025,wang-etal-2026},
and the package prints this caveat with every short form it builds.

A second use arises when response data are scarce, unavailable, or
untrustworthy. \citet{muller-2026} recovered semantic structure, from
adjective vectors weighted by the mean responses of 55 people, in a
study where a confirmatory factor analysis of the same responses fit
poorly and merged two of its five factors, a failure that the author
attributed in large part to the small sample. The semantic analysis in
that study still used the
responses as weights, unlike SFA, so it is a precedent for semantic
structure helping where a small-sample analysis failed rather than for
response-free use. \citet{kambeitz-etal-2025} also note that collecting
the samples needed to develop a clinical questionnaire is costly and can
miss hard-to-reach groups, and that in some countries the cost can make
a sufficient validation sample infeasible. Synthetic
respondents that can pass attention checks \citep{westwood-2025} are a newer
threat to response data, and an analysis that uses no respondents cannot
be contaminated by them. The risk of mistaking machine output for human
data also applies to SFA itself, because structure derived from a
language model must never be reported as evidence about how people
respond. On a College Algebra test, none of six language models on its
own reproduced the distribution of human abilities, although an ensemble
of them approximated it better \citep{liu-etal-2025}. One framework that
simulates respondents also does not claim to replicate the psychological
processes by which people answer survey items \citep{lim-etal-2026}. For this
reason the package measures coherence with human data rather than
assuming it.

A third use is comparing instruments. The jingle--jangle screen found nothing to flag in the Big-Five Factor
Markers. That null result from one inventory says little about the
screen's sensitivity, because the two cosines it compares, between subscale item
averages and between construct names, were all strongly positive, so a
difference of 0.20 is hard to reach. We intend the screen mainly for
comparing scales, including those of different instruments, where
mismatches between labels and content can arise \citep{wulff-mata-2025,wulff-mata-2026}. The package
also complements item-generation tools such as AI-GENIE
\citep{russell-lasalandra-etal-2026}. Such a tool writes new items and already removes redundant ones, and
\texttt{semanticfa} could then apply a further screen for redundancy and
fit against the existing items before anyone answers them. This
combination was not tested here, and the worked example screened only
three candidate items written to illustrate the verdicts.

Finally, design choices affect what an analysis recovers, how private
and reproducible it is, and where it can run. Within this package, the
encoding and the size of the embedding model changed what the worked
example recovered (see the Results). Outside it, in a small preliminary
study of a different pipeline, clustering item embeddings recovered
structure that prompted chatbots did not recover consistently
\citep{huang-etal-2025}. Open models that run on the researcher's own
computer can also keep sensitive data private.
\citet{hussain-etal-2024} make this point in their tutorial on
open-source language models for behavioral science. A model that is
saved and run locally, they note, keeps the data on the researcher's own
hardware, where the model's creators cannot access them. For SFA, recording the version of a
locally saved model has a further benefit, because every reanalysis then
uses the same measurement instrument. Once the item embeddings and the embeddings of any auxiliary words
exist, every step of the workflow described here that works from stored
embeddings runs without a network connection, provided Python with
NumPy \citep{harris-etal-2020} is installed for reading them. The semantic analysis is therefore designed
to give the same numerical results in a clinic that cannot send data off
site and in a teaching laboratory, when both use the same embeddings,
seed, and software versions, including the linear-algebra library. Even
then, the layouts of the two nonlinear item maps in
Figure~\ref{fig:itemmap} can differ slightly between platforms (see the
Open Practices Statement). Across the package, only three steps call Python, namely
embedding new text, reading embeddings stored in NumPy files, and
building the NLI matrix, and embedding new text and building the NLI
matrix also download their models on first use.

\subsection{Limitations}

The central limitation is built into the method, because SFA
characterizes the wording of items rather than the people who answer
them. Without respondents, it cannot show that a trait causes any
response, and it supplies none of the theory of how people arrive at their answers,
which \citet{borsboom-etal-2004} make the focus of validation research. Coherence must therefore never be reported as a validity
coefficient. Semantic evidence bears on two of the aspects of validity
that \citet{messick-1995} distinguishes. It addresses the content aspect,
which concerns which content a scale covers and how well, and it
addresses the structural aspect in part, because it
shows whether the meanings of a scale's items group in the way its
scoring assumes. The remaining aspects require response and outcome
data. Close agreement between the two structures does not remove these
limits. \citet{kojima-etal-2026} found that factor scores from
responses predicted lifetime diagnoses better than scores that weighted
the responses by loadings derived from text, at least for the general
factor. SFA is best described, in the words that \citet{hommel-arslan-2025} use
for their own web application, as ``a free pilot study of survey items''
(p.~15), and it should be followed by human data wherever they can be collected.
\citet{schoenegger-etal-2025} also found that language models predicted
item correlations more accurately than most individual experts but did
not outperform the experts' pooled judgment. Finally, in one study that
tested it, a finding that a model
fits the semantic matrix was more reliable than a finding that a model
does not fit. \citet{feraco-toffalini-2025} fitted a one-factor model to
each four-item scale of a character-strengths inventory. Every model
that fit the semantic matrix also fit the responses, whereas
about half of the models that the semantic matrix rejected fit the
responses as well.

Coherence that is high would be consistent with part of the critique of
\citet{uher-2025}, who argues that rating-scale data are people's
judgments of verbal statements rather than measurements of what the
statements describe. Coherence estimates how closely the structure found
in rating-scale responses follows the meaning of the items, which is the part of that
critique that item text can test. Uher also argues that language-model
methods cannot by themselves show how responses relate to the phenomena
they describe, and SFA shares this limit. Studies of
the semantic theory of survey response estimated, with older text
methods, how closely item wording predicts the correlations among
responses, and those methods predicted almost nothing for a five-factor
personality inventory \citep{arnulf-etal-2014,arnulf-etal-2018}. Their
main estimates were in-sample proportions of variance explained,
computed on other instruments and samples, so they
are not directly comparable with the values found here. Here coherence
was $r = \valPairR$ at the level of item pairs, against the correlations
among raw responses, with a mean matched congruence of $\bar{\phi} =
\valPhiMean$ across factors, values for which no benchmark of coherence across instruments exists
yet. The theory of
meaning behind embeddings sets a further limit. Embeddings summarize how words are used in the language
\citep[\S43]{wittgenstein-1953}, not how an individual respondent reads
an item, so the part of each response that reflects one person's
reading lies outside what SFA can measure. Coherence below 1 should therefore be expected even from a flawless
instrument, although how far below 1 is not known.

A further limitation is specific to the inventory used here. The IPIP
Big-Five Factor Markers are reproduced on public websites together with
their domain labels, including in the documentation of the very data set
used as the benchmark here \citep{openpsychometrics-2018}. The embedding
model was initialized from a Qwen3 generative language model
\citep{zhang-etal-2025}. That model was pretrained on about 36 trillion
tokens of text in 119 languages and dialects, which its technical report
describes only by broad domain, such as code, books, and multilingual
text \citep{yang-etal-2025}, so whether listings of this inventory were
part of that text cannot be checked. Generative language models can
memorize and reproduce text from their training
data \citep{carlini-etal-2021}, so the model may have learned which items
belong together from those listings rather than from what the sentences
mean, which would inflate the agreement between the semantic and response
structures. The semantic solution divided most of the Extraversion items
and \valAsgnAOnBlend{} of the Agreeableness items between two blended
factors, which also took \valAsgnBlendOther{} items from other domains,
and \valAsgnAOnN{} Agreeableness items joined the Neuroticism factor
(Table~\ref{tab:assignment}). This is not what one would expect if the
model had simply reproduced the published domain assignment, because
such a model would have kept the items of each domain together. Partial memorization of the
three domains that were recovered cannot be excluded without reworded
items, however, so coherence measured on well-known inventories may be an
upper bound. An important use of the package, checking new and
unpublished items, is the case that memorization is least likely to
help with. Re-embedding systematically reworded
versions of the items, with their meaning preserved, would bear on this
without new respondents and is left to future work.

The random-embedding baselines have a limit of their own. Parallel
analysis and the fit calibration compare the observed matrix with
matrices built from random embeddings. Random vectors in thousands of
dimensions are nearly unrelated to one another, whereas real item embeddings are positively similar to one another, as
in the worked example and as representations from contextual language
models commonly are, a property called anisotropy
\citep{ethayarajh-2019}. Every raw cosine in the worked example lay
between \valRangeCosMin{} and \valRangeCosMax. Exceeding the random baseline is therefore a weak test for parallel
analysis. For TEFI, a value better than every random fit shows only that
the items are not random. The RMSR comparison is weak for the same
reason, and the CAF comparison should be read with the same caution,
because the baseline matrix is nearly an identity matrix. The Technical Reference in the
Supplementary Materials explains this problem and what a better baseline
would require. Until one exists, the number of factors is better treated
as a judgment informed by the consensus of several criteria and by
theory, as recommended above.

The human benchmark has limits of its own. It is an open, self-selected online sample, cleaned only by removing
records with a missing or out-of-range response
\citep{openpsychometrics-2018}, so its factor structure need not match
that of a representative sample. \citet{guenole-etal-2025} also note
that the people who wrote the text on which an embedding model was
trained may differ from the people who answer a scale, and that the more
the two populations differ, the less well the embeddings may stand in
for responses.

The results also depend on the embedding model. The texts a model was trained on, how much context it reads, and how it
was built can all change which human behaviors its similarities predict,
and no model is best for every task, at least among word-level models
\citep{mandera-etal-2017}. The embedding model is thus part of
the measurement instrument, so the stability of results across models
should be reported. The present results cannot show that stability,
because they rest on one model family (Qwen3-Embedding) and on English
items. Even a model's full embedding need not be optimal, because preliminary
simulation evidence suggests that the signal can concentrate in a subset
of its coordinates \citep{golino-2026}. The
selected depth of \valDepthOpt{} of \valDepthFull{} coordinates here, the
highest point of a noisy curve, is consistent with this but gives it
little support, because the depth criterion favors agreement with the
theoretical domains and so is not an independent test. In one study, embedding methods were also less accurate than expert
judges at assigning items phrased as behaviors, rather than as traits,
to their intended constructs, although on an inventory of concise trait
descriptions every model tested was more accurate than the judges
\citep{milano-etal-2025b}. In that study, the expert judges were
graduate students in psychology who, unlike the models, were given
definitions of the constructs, item length and instrument were
confounded with phrasing, and the authors note that
translating the behavioral items from Italian into English for some
models may have distorted their content. If the finding holds, the
structure of scales whose items describe behaviors may be recovered even
less accurately than the partial recovery obtained here for the Big-Five
Factor Markers, whose short first-person statements mix descriptions of
traits and of behaviors and were not classified by phrasing here. Clinical
experiences that are hard to put into words, and likely rare in training
texts, may be where semantic structure is least trustworthy
\citep{kambeitz-etal-2025}.

Finally, the refinement and screening tools carry their own caveats. Removing items
assumes a reflective model, in which the construct causes the responses
to every item. In a formative model, by contrast, each item is part of
the construct's definition, so removing one usually changes the construct
\citep{bollen-lennox-1991}. The cutoff that Unique Variable Analysis uses
was calibrated on simulated response data rather than on similarity
matrices \citep{christensen-etal-2023}, so how often it errs on
similarity matrices is unknown. Two items with nearly the same meaning
can still differ in how people respond to them, for example in how
readily people endorse them or how extreme their ratings are, so
similarity of meaning does not show that two items measure the construct
in the same way \citep{wang-etal-2026}. Every purely semantic verdict
about redundancy, including this package's, should inform expert
judgment rather than replace it, and so should the screen for
jingle--jangle fallacies \citep{wulff-mata-2026}.

\subsection{Future Directions}

Three directions follow from these results and their limits. The first
is comparative,
because no one can say what counts as high or low coherence until it
has been estimated across instruments, domains, and populations. The
worked example, which analyzed one carefully constructed inventory,
provides a single data point, and the largest differences may lie
between kinds of constructs. One prediction and two findings bear on this question, and each awaits
a systematic test with a common toolkit. If the symptoms of some clinical
constructs cause one another \citep{borsboom-2008}, those constructs
should show lower coherence than common-cause constructs, as argued
above. In one unpublished study, the strength of the relation between
meaning and correlation appeared to vary across nine dimensions from four
scales that measure different constructs \citep{zhang-qiu-2026}.
\citet{milano-etal-2025b} found that embedding methods were generally
less accurate than expert judges at assigning items phrased as behaviors to
their constructs. The judges were graduate students in psychology who, unlike
the models, were given definitions of the constructs, item length and
instrument were confounded with phrasing, and translating the behavioral
items into English for some models may have distorted their content. That finding
concerns how well the intended structure is recovered rather than
coherence, and the phrasing of items rather than the kind of construct.
If it holds, however, whether items are phrased as behaviors or as
traits, which varies across instruments, may also produce differences
in coherence between them.

The second direction is methodological. It concerns four settings that
the present analysis fixed or examined only narrowly, and one test that
it left open. The analysis used a single model
family and compared only three sizes within it. Because results can depend
on the embedding model, as \citet{mandera-etal-2017} found among
word-level models, their stability across
model families should be reported the way agreement across raters is. The number of an embedding's leading coordinates to use
\citep{golino-2026} was examined here only with a criterion that favors
agreement with the theoretical domains, and the main fit used the full
embedding, which is the package's default. The depth could be examined
in every analysis, with a criterion that does not depend on the
theoretical domains. The cutoffs for redundancy
screening should be recalibrated on similarity matrices instead of being
inherited from simulations of response data \citep{christensen-etal-2023}.
The random-embedding baselines could be computed once for common numbers
of items, embedding sizes, and encodings, and matched to the positive
similarity that real item embeddings share, so that a Monte Carlo reference of the kind \citet{pokropek-2026} used
would require looking up a table rather than running a new simulation.
A precomputed baseline would still leave open what it should hold fixed
from one draw to the next (see the Technical Reference). A particularly hard open problem among these is a
better baseline for deciding the number of factors, which the Technical
Reference discusses in detail. The test left open is the one for
memorization described in the Limitations. Re-embedding systematically
reworded versions of a published inventory, with the meaning of each
item preserved, and comparing their semantic structure with that of the
original wording and with the existing human responses would help show,
without new respondents, whether the agreement found here depends on the
model having seen the published items.

The third direction extends screening before data collection to
translated scales. In one study, a scale adapted from English into Polish with a generative
model performed at or near a human adaptation on most psychometric
criteria, with slightly lower reliability \citep{grobelny-etal-2025}, but without responses their structure cannot be
checked until the adapted scale has been administered. SFA could check
the adaptation's semantic structure earlier by embedding the original and the adapted items in a shared
multilingual space. If the model represents the two languages
comparably, which would itself need to be tested, the agreement metrics
could indicate whether the wording of the adaptation has drifted from
the original semantic structure. The response structure of the adapted
scale would still have to be confirmed once it has been administered.

The first and third directions keep the division of labor that these
results suggest, in which the analysis of wording can be automated but its
verdicts remain hypotheses for response data to test. More
than thirty years ago, \citet{lara-etal-1992} proposed testing a
candidate meaning for each factor against the variables that load on it,
and anticipated that automatic semantic interpretation of factor
solutions might become necessary for very large problems. Sentence embeddings now make construct-level versions of such tests
computable, such as comparing items with construct names and centroids, and \texttt{semanticfa}
implements semantic factor analysis as a standard factor-analytic
workflow in R. Three tasks still require response data
and expert judgment: explaining how people arrive at their answers,
relating scores to outcomes, and deciding when a well-characterized set
of items has become a validated measure.

\newpage
\section{Conclusion}


A rating scale has two structures that factor analysis can examine: the
correlations among people's responses and the organization of the
language that drew those responses. Methods for the first have been
developed for over a century \citep{spearman-1904}. Methods for the
second, which work from the meanings of the items, are recent. Most use
sentence-embedding models, such as Sentence-BERT
\citep{reimers-gurevych-2019}, and their precursors represented words
rather than whole sentences \citep{arnulf-etal-2014}. As the Introduction
described, the published ones have been scattered across separate
papers, scripts, and software. The
\texttt{semanticfa} package brings the stages of the factor-analytic
workflow together in one R package. It runs that workflow on the
similarities between item embeddings and can also factor the NLI matrix,
whose similarities come from a natural language inference cross-encoder
rather than from embeddings, although the steps that need embeddings,
such as the random-embedding baselines and the anchoring of items to
construct centroids, do not apply to it. The workflow covers a choice among
encodings, retention by several criteria, diagnostics with an optional
comparison of three fit statistics with fits to random embeddings, tools
for interpretation, and tools for refinement. The semantic analysis
itself needs no responses. Because it works from fixed, archived
embeddings, it is designed so that its numerical results can be
reproduced from those embeddings with the seeds given in the Method and
the software versions and linear-algebra library given in the Open
Practices Statement, apart from small platform
differences in the $t$-SNE and UMAP layouts. The comparison with human
responses also needs the downloaded response file, and psych's parallel
analysis of those responses is not seeded and runs in parallel
processes, so the last digits of its random-data eigenvalues can differ
between runs. The NLI matrix is a
further exception. It does not come from the archived embeddings, and the
cross-encoder that computes it is downloaded on first use. The Open
Practices Statement records the revision of the cross-encoder that was
used. On a well-understood
inventory embedded with Qwen3-Embedding-8B, the consensus of five
retention criteria, with ties going to
the smaller number, was \valConsensusKWord{} factors, the number of
domains the inventory was written to measure, after two criteria chose
five, two chose six, and one chose two. With the two smaller models of
the same family, the consensus was \valConsSmallWord{} for 0.6B and
\valConsMediumWord{} for 4B. Three factors
corresponded mainly to single domains (Openness, Conscientiousness, and
Neuroticism). Most of the Extraversion items and \valAsgnAOnBlend{} of
the Agreeableness items were divided between two blended factors, which
also took \valAsgnBlendOther{} items from other domains, and
\valAsgnAOnN{} Agreeableness items joined the Neuroticism factor
(Table~\ref{tab:assignment}), although a few of these counts rest on near
ties. One blended factor gathered forward-keyed
items about sociability and warmth, and the other gathered reverse-keyed
items that describe the absence of a disposition. Agreement with the
factor structure of \valNHuman{} respondents was closer for Openness,
Conscientiousness, and Neuroticism ($\phi = \valPhiO$, \valPhiC, and
\valPhiN, with only Openness in the range of fair similarity) and weaker
for Extraversion and Agreeableness. Two reasons are plausible, although
neither was tested here. English describes warmth and sociability in
overlapping words, and statements that describe the absence of a
disposition share the wording of negation across topics. For this
inventory and embedding model, sign-flipping the embeddings of
reverse-keyed items agreed less with reverse-scored human responses than
leaving them unflipped did. The package's refinement tools, which need
no responses, flagged redundant item pairs, built a candidate short
form, and screened three new items written to illustrate its verdicts,
so revisions of this kind can be drafted before any data are collected,
as candidates to be validated once responses exist.
We propose that the agreement between the two structures,
semantic--behavioral coherence, be treated as a measurable property of a
scale, for a given embedding model, and counted among the relations in
the nomological network of the construct the scale measures. In the
worked example, coherence was summarized by the correlation across item
pairs between semantic similarities and the correlations among raw
responses ($r = \valPairR$) and by the mean matched congruence of the
factors ($\bar{\phi} = \valPhiMean$). For a given embedding model,
high coherence is consistent with a scale's wording and its respondents'
answers grouping the items in the same way, although what counts as high
remains to be established.
Low coherence points to the domains and items where the two part and
gives a starting point for asking why. For a given embedding model, SFA
estimates the structure that a scale's wording implies before any
respondent is recruited, and once responses exist, coherence shows how
closely the structure of the responses agrees with that wording.

\newpage
\setcounter{secnumdepth}{0}
\section{Declarations}

\subsection{Funding}

This work received no external funding.

\subsection{Acknowledgements}

This research was enabled in part by computing resources provided by the
Digital Research Alliance of Canada.

\subsection{Competing Interests}

The authors have no competing interests to declare.

\subsection{Ethics Approval and Consent}

Not applicable. No new human data were collected. The human benchmark is a
secondary analysis of the publicly released, anonymized Open-Source
Psychometrics Project data set \citep{openpsychometrics-2018}.

\subsection{Availability of Data and Materials}

The embeddings, outputs, and other materials of the worked example are
available in the package's GitHub repository and in its archived Zenodo
release (see the Open Practices Statement for the links and contents).
The human response data are publicly available from the Open-Source
Psychometrics Project \citep{openpsychometrics-2018} and are not
redistributed.

\subsection{Code Availability}

All code is available in the same GitHub repository and Zenodo release. The \texttt{semanticfa} package is available on the Comprehensive R Archive Network (CRAN), and its development sources and issue tracker are at \url{https://github.com/devon7y/semanticfa}.

\subsection{Use of Large Language Models}

During the preparation of this work the authors used various Claude
(Anthropic) and GPT (OpenAI) models to search and summarize the
literature, write and review the code of the package and of the
analyses, draft and revise the text of the manuscript under the authors'
direction, and check its citations, statistics, and claims against the
cited sources and the data. The design of the package and of the worked
example, the analytic decisions, and the interpretations are the
authors'. After using these tools, the authors reviewed and edited the
content as needed and take full responsibility for the content of the
published article.

\section{Open Practices Statement}

A self-contained reproduction archive regenerates every number, table,
and figure in the worked example from raw inputs. It is kept in the
\texttt{reproduce} folder of the package's GitHub repository
(\url{https://github.com/devon7y/semanticfa}), and the release that
accompanies this paper is archived on Zenodo
(\url{https://doi.org/10.5281/zenodo.23696188}). It contains the master script (\texttt{reproduce.R}) and a full-run console transcript (\texttt{reproduce.Rout}) against which every command can be checked. It also contains the precomputed Qwen3-Embedding-8B embeddings of the
auxiliary words (the five construct names and four projection poles)
and of the candidate items, together with the scripts that generated
them, the item embeddings from the two smaller models used in the
model-size comparison, the embeddings of the 42 items of the Depression
Anxiety Stress Scales from all three models, used in the Supplementary
Materials, and a script that re-embeds the items with each
model size and compares the result with the stored vectors. The 8B item embeddings themselves are bundled with the package. The item embeddings came from the Hugging Face revisions \texttt{\valRevSmall} (Qwen3-Embedding-0.6B), \texttt{\valRevMedium} (4B), and \texttt{\valRevLarge} (8B) of the model repositories \citep{qwen-2025}. Re-embedding the 50 items with these revisions, in single precision, without an instruction prompt, and on an NVIDIA H100 GPU, reproduced the stored vectors: the cosine between each new vector and the stored one was at least \valEmbCheckMinCos{} for every item and model size, and \valDassEmbCheckCos{} to six decimal places for the Depression Anxiety Stress Scales items. The output of this check is part of the archive. Results are stored as an \texttt{.rds} object and as CSV tables,
including the encoding comparison, the congruence matrix, both loading
matrices, the NLI similarity matrix, and the fits to random embeddings, alongside the regenerated
figures. The script documents where to download the human response file \citep{openpsychometrics-2018} and reads its location from an environment variable, so the archive does not depend on redistributing it. Two steps of the worked example call Python: reading the archived auxiliary and candidate-item embeddings from NumPy files, which uses the NumPy library \citep{harris-etal-2020}, and building the NLI matrix, which uses sentence-transformers \citep{reimers-gurevych-2019} and downloads its cross-encoder from the Hugging Face Hub on first run. All factor analyses, retention methods,
diagnostics, congruence metrics, refinement analyses, and figures are pure R
and run offline. The worked example was not preregistered.

The reported run used R 4.5.0 \citep{rcoreteam-2025} with semanticfa 0.5.1 \citep{yanitski-westbury-2026}, psych 2.6.5 \citep{revelle-2026}, EGAnet 2.4.1 \citep{golino-christensen-2026}, GPArotation 2026.4-1 \citep{bernaards-jennrich-2005}, Rtsne 0.17 \citep{krijthe-2023}, uwot 0.2.4 \citep{melville-2025}, and reticulate 1.43.0 \citep{ushey-etal-2025}, with seed 42 and the
reference NETLIB linear-algebra library, on a Linux (x86-64, AlmaLinux 9.8) compute
node of the Nibi cluster of the Digital Research Alliance of Canada. The
Python steps used Python \valPyPython{} with NumPy \valPyNumpy{}
\citep{harris-etal-2020}, torch \valPyTorch{} \citep{paszke-etal-2019},
transformers \valPyTransformers{} \citep{wolf-etal-2020}, and
sentence-transformers \valPySentTrans{} \citep{reimers-gurevych-2019}, and the NLI cross-encoder was the Hugging Face revision
\texttt{\valNliRevision}. The NLI matrix was computed on the processors
of that compute node, without a GPU, with the library's default
settings, so its scores may differ slightly on other hardware. The transcript records the Python library versions and this revision. The layouts of the $t$-distributed stochastic neighbor embedding and uniform manifold approximation and projection maps (Figure~\ref{fig:itemmap}) depend on the floating-point arithmetic of the machine, so the positions of the items in those two maps can differ slightly on other platforms. Reproducing the worked example requires semanticfa 0.5.1 or later. The package's development sources are in the same repository, and a getting-started vignette, a tutorial installed with the package, runs the full analysis on the bundled inventory.

\newpage
\bibliography{references}

\newpage
\section{Supplementary Materials}

\setcounter{figure}{0}
\setcounter{table}{0}
\renewcommand{\thefigure}{S\arabic{figure}}
\renewcommand{\thetable}{S\arabic{table}}

These Supplementary Materials hold the detail that a reader needs to
check or extend the analysis but not to follow the main text. The
Technical Reference comes first, because the later sections use its
definitions. It describes the package's functions, how each criterion
for the number of factors is adapted to embeddings, how fit is
calibrated against random embeddings, how the four embedding encodings
and the NLI matrix compare on every metric, and where the random-embedding baselines fall short. The
full loading matrices of the semantic and human solutions follow. The next section shows the NLI matrix, the signed similarity matrix
built by natural language inference (NLI), which uses no embeddings. The
comparison-data misfit profiles follow, because they extend the
discussion of baselines that ends the Technical Reference. The last
section repeats the model-size comparison with a second inventory, the
Depression Anxiety Stress Scales.

\subsection{Technical Reference}

This section gives the technical detail behind the workflow described in
the main text. It has five parts. The Function Reference describes the
package's functions for building, fitting, interpreting, and refining a
semantic factor solution, with the statistical rationale and source of
each. Retention Criteria and Monte Carlo Calibration then explain how
the package chooses the number of factors and how it judges model fit.
These are the two places where methods built for response data had to
be adapted to embedding similarity matrices. The Encoding Comparison
compares all the encodings on the agreement and fit metrics of the main
analysis. The last part explains the limits of the random-embedding
baselines that retention and calibration rely on, which is the main
problem the package leaves open.

\subsubsection{Function Reference}

\textbf{Extraction and diagnostics.} The main function, \texttt{sfa()},
runs the whole
workflow from raw item text, from precomputed embeddings, or from a
similarity matrix. It extracts factors by minimum residual through the
psych package \citep{revelle-2026}. Minimum residual chooses the loadings
so that the similarities the factors reproduce differ as little as
possible from the observed similarities. The function then rotates the
factors by oblimin. Oblimin is an oblique rotation, which allows the
factors to correlate, and it belongs to the family of analytic rotations
that includes varimax \citep{kaiser-1958}. Every fit reports the five
diagnostics that Table~\ref{tab:diagnostics} in the main text defines:
the Kaiser--Meyer--Olkin (KMO) measure of sampling adequacy
\citep{kaiser-1970}, printed with the descriptive labels of
\citet{kaiser-rice-1974}, the root mean square residual (RMSR), the common
part accounted for (CAF) of \citet{lorenzo-seva-etal-2011}, the total
entropy fit index (TEFI) of \citet{golino-etal-2021}, and McDonald's omega total for each factor \citep{revelle-zinbarg-2009}.
Omega is computed from one factor's loadings $\lambda_i$ on a set of
items as $\omega = (\sum_i \lambda_i)^2 / [(\sum_i \lambda_i)^2 + \sum_i
(1 - \lambda_i^2)]$. It is computed twice, once over the items whose
largest absolute loading is on the factor and once over the items of
the theoretical domain that the factor matches best, which is the domain
on whose items the factor's mean absolute loading is largest. This is a
single-factor omega computed from the oblique pattern loadings, so it
ignores cross-loadings and differs from an omega for the full
five-factor model. Assignment by the largest absolute loading is also
arbitrary for items whose two largest loadings are nearly equal. When the
user supplies domain labels, \texttt{sfa()}
also reports a matrix of average absolute loadings. Each entry of this
matrix is the mean absolute loading of one factor across the items of one
theoretical domain, and each factor is matched to the domain with its
largest entry. The matrix adapts the dominant average absolute loading
procedure that \citet{guenole-etal-2025} used to label factors. They
computed the averages over facets and added rules for factors that match
no domain or several, whereas the package computes them over items and
applies no such rules. The
\texttt{calibrate = TRUE} option compares RMSR, CAF, and TEFI with their
values in fits to random embeddings (see Monte Carlo Calibration below).
The function \texttt{as\_psych()} converts a fit into a \texttt{psych::fa} object, so
that the psych package's own functions can be applied to it.
The function \texttt{sfa\_load\_npz()} reads embeddings that were
computed elsewhere and saved as NumPy \texttt{.npz} files, and it needs a
Python installation with the NumPy library. The function
\texttt{sfa\_embed()} embeds new text through Python with the
sentence-transformers library, and it downloads the chosen model on
first use.

\textbf{Similarity.} The function \texttt{sfa\_similarity()} turns a matrix of item
embeddings into an item-by-item similarity matrix. An encoding is the
rule it uses to do so, and the function offers four:

\begin{enumerate}
\item The \texttt{atomic} encoding takes the cosine similarity between the item
embeddings. \citet{guenole-etal-2025} applied this encoding and the next
to embeddings pooled over the items of each facet, and the package
applies them to single items.
\item The \texttt{atomic\_reversed} encoding multiplies each reverse-keyed item's
embedding by $-1$ before taking cosines \citep{guenole-etal-2025}. It is
the only encoding that needs the scoring key.
\item The \texttt{squid} encoding of \citet{pellert-etal-2026}, named for
Survey and Questionnaire Item Embeddings Differentials (SQuID), subtracts
the questionnaire embedding from every item's embedding before taking
cosines. The questionnaire embedding is the mean of all the item
embeddings, and subtracting it can produce negative similarities between
items, which raw cosines rarely do \citep{pellert-etal-2026}. The
original method correlated the centered embeddings, whereas the package
takes cosines.
\item The \texttt{mean\_centered\_pearson} encoding, the default, subtracts each
embedding's own mean value from all of its entries before taking
cosines. The cosine of two centered vectors equals their Pearson
correlation, so the result is a Pearson correlation matrix, which can
differ from the matrix of raw cosines \citep{pokropek-2026}.
\end{enumerate}

\noindent A fifth route uses no embeddings. The function \texttt{sfa\_nli\_matrix()}
scores every ordered pair of items with a cross-encoder trained for
natural language inference (NLI), the task of judging whether one
sentence entails, contradicts, or is neutral toward another
\citep{bowman-etal-2015}. A
model of this kind reads two sentences together and estimates the probabilities that the first
entails the second, that it contradicts the second, and that it does
neither. The score of an ordered pair is the probability of entailment
minus the probability of contradiction, a signed mapping in the spirit
of \citet{hommel-arslan-2025}, who gave contradicting pairs negative
similarities. Entailment is directional, so the two orders
of a pair can receive different scores, and the function averages them
to make the matrix symmetric. The resulting matrix need not be positive
semi-definite. A matrix that is not positive semi-definite has at least
one negative eigenvalue, so it cannot be the correlation matrix of any
set of variables. The function returns the matrix as it is. When a matrix with a negative
eigenvalue is passed to \texttt{sfa()}, \texttt{sfa()} prints a message and
replaces it with a positive-definite correlation matrix. It raises every eigenvalue below
$10^{-6}$ to $10^{-6}$, rebuilds the matrix from the adjusted
eigenvalues, and rescales it so that every diagonal entry is one.

\textbf{Interpretation.} The function \texttt{sfa\_anchor()} adapts the
approach of \citet{wulff-mata-2025}. It computes each item's belonging to
each construct as the cosine similarity between the item's raw embedding
and the construct's anchor, so it does not use the similarity matrix that
was factored. Belonging is the semantic counterpart of a loading. An anchor is
either a construct's centroid or its name embedding. The centroid is the
mean embedding of the construct's items, and the name embedding is the
embedding of the construct's name, such as \emph{extraversion}. When the
function computes an item's belonging to its own construct, it leaves
the item out of the centroid, so that the item is not compared with an
average that contains it. The printed output lists the eight items with
the lowest belonging to their own construct and marks each of them that
belongs more strongly to another construct.

The function \texttt{sfa\_project()} adapts the semantic projection of
\citet{grand-etal-2022}. It defines an axis as the difference between the
embeddings of two poles, such as \emph{calm} minus \emph{anxious}, where
a pole given as several words is the mean of their embeddings. Grand et
al.\ projected words onto lines defined by several pairs of opposite
words. The package instead scores each item by the cosine $c$ between its
embedding and the axis and then rescales the score as $(c - c_{\mathrm{low}})
/ (c_{\mathrm{high}} - c_{\mathrm{low}})$, where $c_{\mathrm{low}}$ and
$c_{\mathrm{high}}$ are the cosines of the two pole embeddings with the
axis. The low pole (\emph{anxious}) therefore scores 0 and the high pole
(\emph{calm}) scores 1, and items can score outside that range.

The function \texttt{sfa\_congruence()} compares a fit with a theoretical assignment
of items to constructs, with a factor solution from responses, or with a
correlation matrix. It reports up to five metrics:

\begin{enumerate}
\item Tucker's congruence coefficient $\phi$ for each pair of factors,
computed by \texttt{psych::factor.congruence} \citep{revelle-2026}, with
the benchmarks of \citet{lorenzo-seva-tenberge-2006}. It is
the cosine between the two factors' columns of loadings, so it equals 1
when the two factors have the same pattern of loadings.
\item The normalized mutual information (NMI) between two assignments of
items to groups, which earlier embedding work also used to score
recovered structures \citep{garrido-etal-2025,golino-2026}. A factor
solution becomes an assignment by placing each item with the factor on
which its absolute loading is largest. NMI is the mutual information of
the two assignments divided by the geometric mean of their entropies
\citep{strehl-ghosh-2002}. \citet{garrido-etal-2025} divided by the
arithmetic mean of the entropies instead, which gives similar but not
identical values. It is 0 when knowing
one assignment says nothing about the other and 1 when the two
assignments are identical.
\item The adjusted Rand index (ARI) of \citet{hubert-arabie-1985} between
the same two assignments. It counts the item pairs that both assignments
place together or both place apart, as a proportion of all pairs, and
corrects that proportion so that chance agreement scores 0 and identical
assignments score 1.
\item The Frobenius similarity between the factor-correlation matrices of
the two solutions. It is the cosine between the two matrices after each
is written out as a single list of numbers. The matrices are compared in
the order in which each solution reports its factors, without first
matching the factors, so the value depends on that order. It is
therefore not interpreted in this article.
\item The disattenuated item-pair correlation. It is the correlation
between the item-pair entries of the semantic matrix and those of the
comparison matrix, divided by the square root of the product of the two
matrices' reliabilities. Each matrix's reliability is estimated by splitting its item-pair entries
into odd- and even-numbered halves in the order in which the entries are
stored, correlating the halves, and stepping the result up with the
Spearman--Brown formula \citep{brown-1910,spearman-1910}. The division by
the reliabilities is the correction for attenuation of
\citet{spearman-1904-proof}. This split depends on the order of the items
and is not a standard reliability estimate. The result is capped at 1,
so a value of 1 can hide a larger uncorrected value. When either
reliability is not positive, the correction is undefined, and the metric
is reported as missing with a warning.
\end{enumerate}

\noindent The function \texttt{sfa\_jinglejangle()} screens a set of scales for
jingle--jangle fallacies \citep{wulff-mata-2025,wulff-mata-2026}. A
jingle fallacy is giving the same name to scales that measure different things. A jangle fallacy is giving different names to scales that
measure the same thing. For every pair of scales, the function compares the cosine between the
embeddings of the two scales' names with the cosine between the means of
their raw item embeddings. By default it flags the pair when the two similarities
differ by 0.20 or more. This single threshold is the package's own.
\citet{wulff-mata-2025} instead set separate cutoffs for the two
similarities from quantiles of their distributions.

\textbf{Refinement.} The function \texttt{sfa\_redundancy()} flags pairs of items that
are locally dependent, which means that they share more than the factor
they both measure, typically because they say nearly the same thing. By
default it uses Unique Variable Analysis (UVA) as described by
\citet{christensen-etal-2023}. UVA first estimates a network in which the
items are nodes and each link is the partial correlation between two
items, the correlation that remains when all other items are held
constant. The network is estimated with EBICglasso
\citep{golino-christensen-2026}. EBICglasso is the graphical lasso \citep{friedman-etal-2008}, which sets
weak links to zero, with the strength of that penalty chosen by the
extended Bayesian information criterion \citep{foygel-drton-2010}. EBICglasso needs a sample size,
and a similarity matrix has none, so the package supplies a nominal
sample size of 1{,}000 or 20 times the number of items, whichever is
larger. This value is a fixed convention rather than an estimate. In the
worked example the number of flagged pairs was the same with 500 and
2,000, but for other scales the flagged pairs could change with it.
UVA then computes the weighted topological overlap (wTO) of every pair of
items. The wTO of two items is high when they are linked with similar
strengths to the same other items. Pairs whose absolute wTO is 0.25 or
more, the cutoff that \citet{christensen-etal-2023} recommend, are
flagged. From each redundant cluster the function keeps the item with
the highest mean absolute similarity to all the items and suggests
removing the others. A simpler alternative flags pairs whose absolute similarity in the
fitted matrix is 0.80 or more.

The function \texttt{sfa\_simplify()} builds a short form without responses
\citep{jung-seo-2025,wang-etal-2026}. It offers two ways of choosing
items. The default, \texttt{"anchor"}, keeps in each construct the items
most similar to the construct's centroid, with each item left out of the
centroid it is compared with. This is in the spirit of
\citet{wang-etal-2026}, who selected the items with the highest
cluster-membership probability. The alternative, \texttt{"medoid"}, chooses
items within each construct one at a time, each time taking an item that
is close to the centroid and far from the items already chosen. It gives
up some closeness to the centroid in exchange for broader coverage of the
construct. The function then refits the shortened scale with the same encoding,
chooses its number of factors by parallel analysis, and compares it
with the full scale on the number of factors retained and on agreement
with the theoretical domains (NMI and ARI). It labels the result a
candidate that still needs validation with response data.

The function \texttt{sfa\_item\_fit()} extends the anchoring approach of
\citet{wulff-mata-2025} to newly written items. It embeds each candidate
item with the model recorded in the fit, or with another model or
embedding function that the user supplies. It then reports the
candidate's cosine with each construct's name embedding and with the
centroid of each construct's raw item embeddings, the gap between the
construct it is most similar to and the next one, and its similarity to
the nearest existing item. A gap below 0.07 marks a cross-loader, and a
gap from 0.07 to 0.15 is reported as some cross-loading risk. The
candidate's similarity to its best construct is called typical when it
lies within 0.05 of the average for that construct's items, each item
compared with a centroid that leaves it out. A candidate whose
similarity to an existing item is 0.90 or more, the default cutoff, is
flagged as a near-duplicate. These cutoffs are heuristics of the
package, not published criteria.

\textbf{Display.} The function \texttt{sfa\_corplot()} draws the similarity matrix as
a heatmap with the items grouped by factor. The function \texttt{sfa\_itemplot()} maps
the items to two dimensions by $t$-distributed stochastic neighbor
embedding \citep{vandermaaten-hinton-2008}, computed with the Rtsne
package \citep{krijthe-2023}, uniform manifold
approximation and projection \citep{mcinnes-etal-2018}, computed with the
uwot package \citep{melville-2025}, principal component analysis, or classical multidimensional scaling. On these maps, items
with similar meanings appear close together. The \texttt{plot()} method
for a fit draws either a heatmap of the loadings or a scree plot of the
eigenvalues in decreasing order. The scree plot in the main text was
drawn directly from the eigenvalues of the similarity matrix, with the
parallel-analysis thresholds added.

\subsubsection{Retention Criteria}

Retention is the decision of how many factors to keep. Some retention
criteria compare the observed eigenvalues with the eigenvalues that
sampling error would produce in a sample of a given size. These criteria
cannot be applied to a similarity matrix unchanged, because a similarity
matrix comes from no sample. The package adapts them by putting random
embeddings of the same size as the real ones, or the embedding
dimension, where the sample would go. The function \texttt{sfa\_nfactors()} reports the consensus of up to
five criteria, which is the number of factors chosen most often, with a
tie going to the smaller number. The criteria are not independent. With
its reference values floored at one, for example, the empirical Kaiser
criterion gave the same answer as the Kaiser rule on all three embedding
matrices analyzed here. The five criteria are the following:

\begin{enumerate}
\item Parallel analysis \citep{horn-1965}, computed by
\texttt{sfa\_parallel()}. Its comparison eigenvalues come from the
similarity matrices of random embeddings. Each random embedding is a set
of random vectors of unit length with the same number of items and the
same embedding dimension as the real embeddings. By default the function
builds 100 such matrices and compares each eigenvalue with the 95th
percentile of the corresponding random eigenvalues, rather than with the
mean that \citet{horn-1965} used. Following Horn, it counts the leading
eigenvalues that exceed their thresholds and stops at the first that does
not, so an eigenvalue that exceeds its threshold after that point is not
counted.
\item The Kaiser rule \citep{kaiser-1960}, which keeps every factor whose
eigenvalue is greater than one. The package applies it unchanged.
\item The empirical Kaiser criterion of \citet{braeken-vanassen-2017},
computed by \texttt{sfa\_ekc()} with the embedding dimension $n$ in place
of the sample size. For $J$ items, each eigenvalue is compared with a
reference value. The reference value is $(1 + \sqrt{J/n})^2$, which is
approximately the largest eigenvalue expected from $n$ observations of
$J$ unrelated variables, multiplied by the average variance that remains
per eigenvalue once the earlier eigenvalues are subtracted from the
total. No reference value is allowed to fall below one. Factors are kept
up to the first eigenvalue that does not exceed its reference value.
\citet{braeken-vanassen-2017} derived and evaluated the criterion for
correlation matrices of sample data, so its use with the embedding
dimension in place of the sample size goes beyond the conditions they
studied.
\item The TEFI criterion, which keeps the number of factors whose solution
has the lowest TEFI, the Von Neumann entropy version of the index
\citep{golino-etal-2021}. \citet{golino-etal-2021} developed the index to
compare candidate structures of at least two factors with simple
structure. The package applies it unchanged to the similarity matrix,
fitting solutions with 1 to 15 factors by the extraction and rotation of
the main fit and keeping the number with the lowest TEFI. This scan over
the number of factors goes beyond the comparisons of candidate structures
that \citet{golino-etal-2021} tested. They also found that the index can
indicate too few factors when loadings and correlations between factors
are both high, although the problem seemed to vanish with more items per
factor. Their simulations used binary items and two or three factors with four
or eight items each, so the ten items per domain here lie outside the
conditions they studied. They also left the extension of the index to
one-factor structures to future work, so the one-factor candidate in the
package's scan lies outside the structures the index was developed
for. On these matrices it chose \valTefiKWord{} factors at
every model size, fewer than the theory contains.
\item Exploratory Graph Analysis (EGA), run through the EGAnet package
\citep{golino-christensen-2026}. EGA \citep{golino-epskamp-2017} estimates
a network of the items and keeps one factor for each community it
finds. A community is a group of
items more strongly linked to one another than to the other items. In
the benchmark of \citet{garrido-etal-2025}, which used personality items
written by a language model (GPT-4o) and a single embedding model, EGA
recovered the number of dimensions of embedding matrices more often than
parallel analysis and the Kaiser rule, both applied with principal
component extraction. That benchmark estimated its networks with
EBICglasso, which requires a sample size. The package instead uses the triangulated maximally filtered graph
\citep{massara-etal-2016}, which builds the network from the similarity
matrix alone, and it finds communities with the Walktrap algorithm
\citep{pons-latapy-2006}.
\end{enumerate}

Two further tools were not part of the consensus. Velicer's minimum
average partial test \citep{velicer-1976}, computed by
\texttt{sfa\_map()}, removes the first $m$ principal components from the matrix for
increasing values of $m$, up to two fewer than the number of items, and
keeps the $m$ at which the average squared partial correlation among the
items is smallest. On the present matrix it indicated removing \valMapK{}
components, well beyond the five domains the
inventory was written to measure. Including it would have added a vote
far above five, although with the plurality rule it would not by itself
have changed the consensus. The function \texttt{sfa\_cd()} adapts the comparison-data method of
\citet{ruscio-roche-2012}. Instead of a single number, it reports how
well each candidate number of factors reproduces the observed
eigenvalues. The resulting profile shows whether the matrix has a sharp
boundary between major and minor factors (see Comparison-Data Misfit
Profiles below).

The function \texttt{sfa\_dimselect()} addresses a question that only embeddings
raise, which is how many of an embedding's coordinates are needed to
recover the structure. It adapts the embedding-depth optimization of \citet{golino-2026}, using
ordinary EGA on the leading coordinates where Golino used dynamic EGA.
For a grid of \valDepthNEval{} depths $d$, it rebuilds the
similarity matrix from the first $d$ coordinates of each embedding, in
their stored order, and runs EGA with the triangulated maximally
filtered graph and the Walktrap algorithm. The grid has \valDepthGridN{}
depths spaced \valDepthStep{} apart, from \valDepthMin{} to
\valDepthGridLast{} leading coordinates, plus the full embedding of
\valDepthFull{} coordinates. The function then rescales two quantities across the depths so that each
runs from 0 at its lowest value to 1 at its highest. These are the NMI
between the EGA communities and the theoretical domains, and the TEFI of
the EGA solution. The chosen depth maximizes $0.70 \times \mathrm{NMI} -
0.30 \times \mathrm{TEFI}$ computed on the rescaled values, so the
weights favor agreement with the theoretical domains over entropy fit:

{\scriptsize
\begin{verbatim}
> ds <- sfa_dimselect(E8, factors = big5$factors,
+                     encoding = "mean_centered_pearson")
> ds
EGA-based embedding-dimension selection
  Method: EGA depth optimization (adapts Golino 2026)
  Full embedding dim: 4096
  Depths evaluated:   148 (range 3-4096)
  Criterion:          0.70*NMI - 0.30*TEFI (normalized)
  Optimal depth:      1431 of 4096 coordinates
    at optimum: n_dim=5  NMI=0.704  TEFI=-23.291
\end{verbatim}
}

\begin{figure}[tbp]
\centering
\captionsetup{justification=raggedright,singlelinecheck=false}
\includegraphics[width=\textwidth]{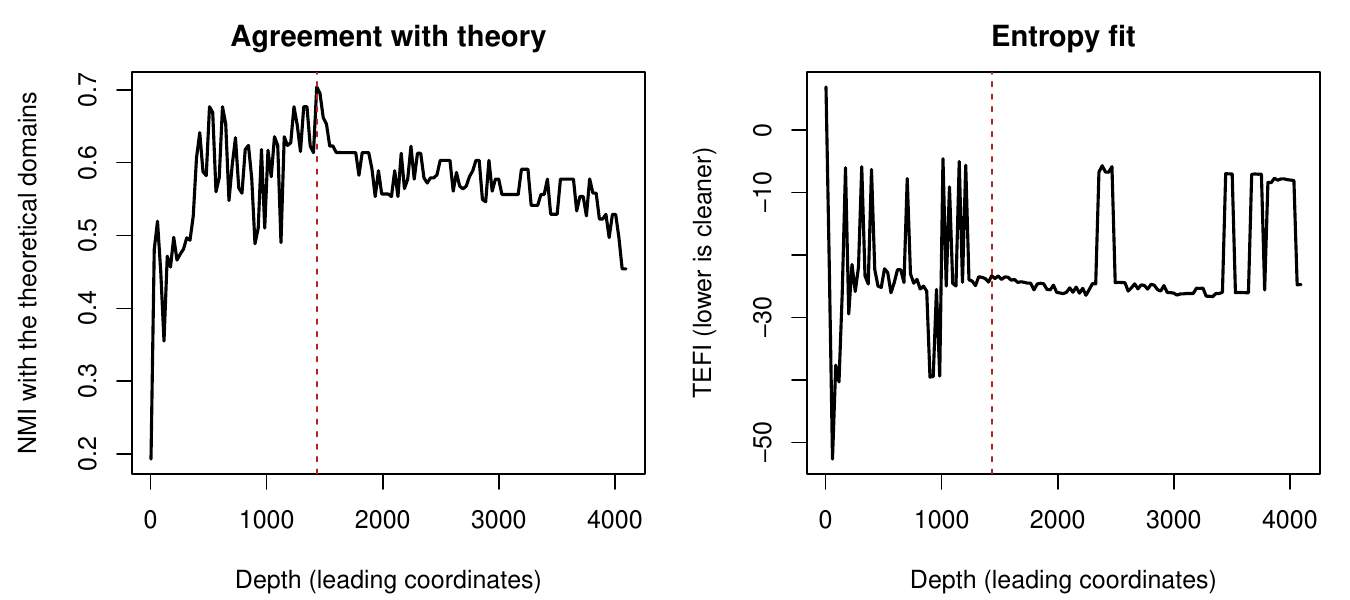}
\caption{Depth selection for the Qwen3-Embedding-8B item embeddings. Each
panel shows one quantity at every depth evaluated, the number of leading
coordinates kept (\valDepthGridN{} depths spaced \valDepthStep{} apart
from \valDepthMin{} to \valDepthGridLast, plus the full \valDepthFull).
Left: normalized mutual information (NMI) between the exploratory graph
analysis communities and the theoretical domains. Right: the total
entropy fit index (TEFI), for which lower values indicate a cleaner
division. The curves show the unrescaled values. The selected depth,
marked by the dashed line, is \valDepthOpt, the depth at which
$0.70 \times \mathrm{NMI} - 0.30 \times \mathrm{TEFI}$ is largest after
each quantity is rescaled across the depths to run from 0 to 1. NMI
there is \valDepthNmi, the maximum of a jagged curve.}
\label{fig:s-depth}
\end{figure}

In this printout, \texttt{n\_dim} is the number of EGA communities, and
the embedding-dimension selection it reports is the selection of the
depth. The chosen depth was \valDepthOpt{} of the \valDepthFull{} coordinates,
roughly the first third. At that depth EGA found \valDepthNDimWord{}
communities, and their agreement with the theoretical domains was NMI =
\valDepthNmi, where 1 would mean identical groupings. Because the criterion rewards agreement with the theoretical domains,
this choice of depth depends on the theory and is not an independent
test of it. The TEFI in this printout belongs to the EGA solution at the
chosen depth, and the NMI compares the EGA communities with the domains.
Neither is the TEFI or NMI of the five-factor \texttt{sfa()} fit.
NMI was the maximum over the depths evaluated, which ranged from
\valDepthNmiMin{} to \valDepthNmiMax. Figure~\ref{fig:s-depth} shows NMI
and TEFI at every depth.

The criteria disagreed on these data. Parallel analysis and EGA kept
\valPaKWord{} factors, the Kaiser rule kept \valKaiserKWord, the
empirical Kaiser criterion kept \valEkcKWord, and TEFI kept
\valTefiKWord. In simulations with personality items written by a language model and
embedded with a single embedding model, the Kaiser rule and parallel
analysis chose too many factors more often than EGA
\citep{garrido-etal-2025}, although here the adapted parallel analysis
agreed with EGA.
Criteria applied to response data can also indicate too many factors.
Conventional parallel analysis of the responses of the \valNHuman{}
respondents indicated \valHumanPaKWord{} factors, possibly because with
that many respondents even very small factors exceed the mean
random-data eigenvalues, although this explanation was not tested. On both kinds of data, then, some criteria indicated more factors
than the theory contains. We therefore recommend reporting the consensus
of several criteria and choosing the final number of factors with the theory in view, which for
this inventory is the five-factor structure \citep{goldberg-1990,goldberg-1992}.

\subsubsection{Monte Carlo Calibration}

Conventional cutoffs for fit statistics, such as a standardized root
mean square residual of about .08 or below \citep{hu-bentler-1999}, were
derived from fits to simulated samples of respondents, so they were set
with the sampling error of correlations in mind. An embedding similarity
matrix has no respondents. The values its fit statistics take by chance
depend instead on the items and the embedding space, so cutoffs such as
the one for the standardized root mean square residual can be too lenient
for it, as \citet{pokropek-2026} found for confirmatory models with few
indicators. In his simulations the share of random models that passed
the standardized root mean square residual cutoff fell from 95\% with
four indicators to 66\% with eight, and he judged the cutoffs for the
comparative fit index, the Tucker--Lewis index, and the root mean square
error of approximation appropriate with six or more indicators. Many model-based tests and indexes of fit
require a sample size, which a similarity matrix lacks
\citep{suarez-alvarez-etal-2026}. The \texttt{calibrate = TRUE} option
supplies a reference. With a fixed seed (42 by default), it draws sets of
vectors from a standard normal distribution, with the same number of
items and the same embedding dimension as the scale, and rescales each
vector to unit length. Such vectors share no semantic structure by
construction. The function takes the raw cosines among each set,
whatever encoding the real fit used, so the random vectors are not
centered, whereas the default encoding of the observed fit centers each
embedding. It fits the same number of factors
with the same extraction and rotation, and records the values of RMSR,
CAF, and TEFI. A replicate whose fit fails is dropped, and in the worked example
\valCalReps{} of 100 replicates were retained. Their values are written
to \texttt{output/calibration\_null.csv} in the reproduction archive. The transcript reports \valCalRotWarnN{} warnings that the oblimin
rotation stopped at its limit of 1,000 iterations without converging,
and a single fit can issue this warning many times. Rerunning the
replicates with the same random draws, which reproduced their RMSR
values exactly, and counting the warnings fit by fit showed that all
\valCalWarnTotal{} came from \valCalWarnReps{} fit, so the rotation
failed to converge in \valCalNonConvWord{} of the 100 fits. RMSR and CAF do not depend on the rotation, but TEFI,
which uses each item's assignment to a factor, does. The remaining
warning repeats that the keying-free encoding ignores the scoring
column. This procedure
adapts the reference distributions of \citet{pokropek-2026}, which were
built from words drawn at random from the same text corpus as the real
words. Words drawn this way keep the general thematic similarity that
real words share, whereas random unit vectors do not.

{\scriptsize
\begin{verbatim}
> fit_cal <- sfa(big5_df, embeddings = E8, encoding = "mean_centered_pearson",
+                nfactors = 5, calibrate = TRUE, calibrate_iter = 100)
There were 21 warnings (use warnings() to see them)
> fit_cal$calibration
[lists of random-embedding RMSR, CAF, and TEFI values omitted]
> cat("Calibration replicates retained:", length(fit_cal$calibration$rmsr),
+     "of 100\n")
Calibration replicates retained: 100 of 100
> summary(warnings())
Summary of (a total of 21) warning messages:
 1x : 'scoring' has reverse-keyed items but is ignored for encoding =
  "mean_centered_pearson": this method is keying-free by design. Use
  "atomic_reversed" for keyed sign-flipping.
20x : In engine(A, Tmat = current_Tmat, normalize = normalize,  ... :
  Convergence not obtained in GPFoblq. 1000 iterations used.
\end{verbatim}
}

Across the random replicates, RMSR ranged from \valCalRmsrMin{} to
\valCalRmsrMax{} (mean \valCalRmsrMean), CAF from \valCalCafMin{} to
\valCalCafMax{} (mean \valCalCafMean, $SD = \valCalCafSd$), and TEFI from
$\valCalTefiMin$ to $\valCalTefiMax$ (mean $\valCalTefiMean$, $SD =
\valCalTefiSd$). Over the \valCalConvN{} replicates whose rotation
converged, TEFI ranged from $\valCalTefiConvMin$ to
$\valCalTefiConvMax$. Each statistic needs its own
reading. The random fits passed the conventional RMSR cutoff easily,
because the cosines among random unit vectors are close to zero and
leave the factors almost nothing to reproduce. RMSR grows with the size
of the similarities being reproduced, so the observed RMSR (\valRmsr),
although above the whole random range, cannot be read as misfit, and
passing the cutoff says little about fit. The observed TEFI ($\valTefiMain$) was lower than every random value, so
none of the \valCalReps{} random fits was as low, which shows only that
the real items are not random. CAF is about .5 when the residuals share
no structure, as in the random fits, and the observed CAF (\valCaf) was
lower than every random value. This suggests that the similarities the
five factors leave unexplained still share structure that random
embeddings lack, although, as for RMSR, a baseline of near-zero
similarities makes this a weak contrast. These results echo the finding of \citet{pokropek-2026} that, for
confirmatory models with few indicators and for the standardized root
mean square residual, values produced by chance can pass conventional
thresholds. We therefore recommend reporting calibrated values alongside
the conventional ones, with the caveat that the random-embedding baseline
is weak. Which baseline to calibrate against is still the
analyst's choice, as discussed under Limits of the Random-Embedding
Baselines below.

\subsubsection{Encoding Comparison}

Table~\ref{tab:s-encodings} reports a five-factor solution under each of
the four embedding encodings and under the NLI matrix. Each solution is scored
against the theoretical domains, against the human factor solution, and
against the human item-pair correlations, and its diagnostics are
listed alongside. The NLI matrix was not positive semi-definite. It had
\valNliNegEig{} negative eigenvalues, the smallest $\valNliMinEig$, so it
was replaced by a positive-definite matrix before factoring, as
described in the Function Reference above. Across the item pairs, the
repaired values correlated \valNliRepairR{} with the original ones. The
factor solution in the \texttt{nli} row uses the repaired matrix, and the
item-pair correlations $r_{\text{keyed}}$ and $r_{\text{raw}}$ use the
original one.

\begin{table}[tbp]
\centering
\captionsetup{justification=raggedright,singlelinecheck=false}
\caption{Agreement of each five-factor solution with theory and with
human data, and its diagnostics, for the four embedding encodings and
the NLI matrix}
\label{tab:s-encodings}
\tighttable
{\footnotesize
\setlength{\tabcolsep}{3.6pt}
\begin{tabular}{lrrrrrrrrrr}
\toprule
& \multicolumn{2}{c}{Theory} & \multicolumn{4}{c}{Human data} &
\multicolumn{4}{c}{Diagnostics} \\
\cmidrule(lr){2-3} \cmidrule(lr){4-7} \cmidrule(lr){8-11}
Encoding or matrix & NMI & ARI & $\bar{\phi}$ & $r_{\text{keyed}}$ &
$r_{\text{raw}}$ & $r_{\text{dis}}$ & KMO & TEFI & RMSR & CAF \\
\midrule
atomic & .527 & .402 & .744 & .434 & .456 & .707 & .971 & $-27.72$ & .032 & .412 \\
atomic\_reversed & .527 & .402 & .450 & .137 & .444 & -- & .971 & $-27.72$ & .032 & .412 \\
squid & .391 & .272 & .622 & .567 & .490 & 1.000 & .022 & $-44.39$ & .072 & .996 \\
mean\_centered\_pearson & .527 & .402 & .744 & .434 & .456 & .707 & .971 & $-27.71$ & .032 & .412 \\
nli & .548 & .448 & .286 & .063 & .561 & -- & .551 & $-54.15$ & .094 & .688 \\
\bottomrule
\end{tabular}

}
\caption*{\textit{Note}. Each row is a five-factor solution of the 50 International Personality
Item Pool Big-Five Factor Marker items under one of the four embedding
encodings or, in the \texttt{nli} row, under the signed natural language
inference (NLI) matrix, the probability of entailment minus the
probability of contradiction. The encodings are \texttt{atomic}, the
cosine of the item embeddings, \texttt{atomic\_reversed}, the cosine after
the embeddings of reverse-keyed items are multiplied by $-1$,
\texttt{squid}, the cosine after the mean of all the item embeddings is
subtracted from each, and \texttt{mean\_centered\_pearson}, the default,
the cosine after each embedding is centered across its coordinates. For that row, the factor solution and its
diagnostics use the repaired NLI matrix, and $r_{\text{keyed}}$ and
$r_{\text{raw}}$ use the original one. Human comparisons use $N =
\valNHuman$ respondents. NMI = normalized mutual information and ARI = adjusted
Rand index, both measuring agreement between the solution's grouping of
the items and the theoretical Big Five domains (1 = identical groupings).
$\bar{\phi}$ = mean matched congruence, the mean Tucker congruence
between the semantic factors and the human factors, after each semantic factor is matched to the human
factor it resembles most (1 = identical loading patterns). Because the
sign of an oblimin factor is arbitrary, factors were matched on the
absolute value of their congruence, and $\bar{\phi}$ averages these
absolute values. $r_{\text{keyed}}$ and $r_{\text{raw}}$ = correlation
between the encoding's \valNPairs{} item-pair similarities and the human
inter-item correlations computed from reverse-scored and raw responses,
respectively. $r_{\text{dis}}$ = $r_{\text{keyed}}$ corrected for the reliability of
both matrices (disattenuated), each reliability estimated from an
odd-even split of the item-pair entries and capped so that
$r_{\text{dis}} \le 1$. It is computed against the reverse-scored human
matrix for every row. Dashes mark encodings for which $r_{\text{dis}}$ could not be computed
because the similarity matrix's split-half reliability was not positive,
probably because of the mixture of positive and negative entries that
the sign flip of \texttt{atomic\_reversed} and the contradiction scores of
the NLI matrix create. The warnings that report this are printed in the
transcript. KMO = Kaiser--Meyer--Olkin measure of sampling
adequacy, TEFI = total entropy fit index (Von Neumann version), RMSR = root mean square
residual, and CAF = common part accounted for. These four are the
diagnostics of each solution other than omega (see
Table~\ref{tab:diagnostics} in the main text). Within a row, lower TEFI
indicates a cleaner division of the items. KMO describes each matrix's
own factorability, and TEFI, RMSR, and CAF describe how well each
solution fits its own matrix, so none of the four is comparable across
rows. The
CAF of \texttt{squid} near 1, for example, follows from its KMO near 0. All rankings are
descriptive, from a single analysis of 50 items, and gaps of a few
hundredths cannot be distinguished without intervals.}
\end{table}

Three patterns appear in Table~\ref{tab:s-encodings}. First,
\texttt{atomic\_reversed} agreed least of the four embedding encodings
with the correlations among reverse-scored responses ($r_{\text{keyed}} = \valArRKeyed$, against
\valAtRKeyed{} for \texttt{atomic}, the same encoding without the sign
flip), and its agreement with the raw correlations ($r_{\text{raw}} =
\valArRRaw$) was close to that of \texttt{atomic} and
\texttt{mean\_centered\_pearson}.
Reverse-scoring a response keeps its content and inverts only the rating
scale. Multiplying an embedding by $-1$, by contrast, reverses the whole vector,
which plausibly makes it point away from the item's topic as well as
from its position on the trait. In the similarity matrix the flip only
changes the signs of some entries, so it cannot change the grouping of
the items, and the NMI, ARI, KMO, RMSR, and CAF of
\texttt{atomic\_reversed} equal those of \texttt{atomic} by construction.
Its lower agreement with the reverse-scored correlations likely reflects
the signs it imposes. Every unflipped cosine is positive, so the flip makes
the similarity of every forward-keyed item with every reverse-keyed item
negative, whereas reverse-scoring is meant to make the items of a domain
correlate positively whatever their keying. The flip is compared with the
reverse-scored correlations because it applies the same scoring key, and
against the raw responses it gave no gain. The sign flip also makes the split-half reliability of the
similarity matrix negative, so the disattenuated correlation is
undefined. The package reports that value as missing, with a warning that
names the cause, rather than dropping it from the output.

Second, SQuID agreed best with the correlations among reverse-scored
responses ($r_{\text{keyed}} = \valSqRKeyed$, with $r_{\text{dis}}$ at
its ceiling of \valSqDis). On the like-with-like comparison with raw
responses, which suits a keying-free encoding, it came second to the NLI
matrix ($r_{\text{raw}} = \valSqRRaw$ against \valNliRRaw). Its matrix was
also nearly unfactorable (KMO = \valSqKmo,
where values near 0 mean that the matrix is unsuited to factor analysis).
A plausible reason is that subtracting the questionnaire embedding keeps
the relations between pairs of items but removes the variance that all
the items share, which is the variance that factor extraction depends
on. Only
\texttt{mean\_centered\_pearson} and \texttt{atomic} combined agreement
with the theoretical domains within \valNmiGapNli{} NMI of the best
encoding, the closest match to the human loading patterns ($\bar{\phi} =
\valPhiMean$), and a matrix well suited to factor analysis (KMO =
\valKmo). On these embeddings the two encodings gave the same values to three
decimals on every metric in the table except TEFI, which differed in the
second decimal. The conclusions that this table supports for
\texttt{mean\_centered\_pearson}, the encoding of the main analysis,
therefore hold for \texttt{atomic} as well. The table cannot choose
between the two, and the default rests on its design, because it needs no
scoring key and yields a Pearson correlation matrix.

Third, the NLI matrix ranked differently from the four embedding
encodings on two kinds of metric. It matched the theoretical domains best
(NMI = \valNliNmi) and agreed best with the raw correlations
($r_{\text{raw}} = \valNliRRaw$), but it matched the human loading
patterns worst ($\bar{\phi} = \valNliPhi$). Its factor-based columns
(NMI, ARI, $\bar{\phi}$, KMO, TEFI, RMSR, and CAF) come from the repaired
matrix, whose values correlated only \valNliRepairR{} with the original
ones, so its rankings on them partly reflect the repair. Its
$r_{\text{raw}}$, which uses the original matrix, is the cleaner NLI
result.

No single encoding or agreement metric was best on every comparison, and
the table supports two reporting practices. The first is to compare like
with like, which means comparing keying-free encodings with correlations of raw
responses and key-using encodings with correlations of reverse-scored
responses. Mixing the two misstates coherence. The NLI matrix, for
example, agreed with the raw correlations at $r_{\text{raw}} =
\valNliRRaw$ but with the correlations among reverse-scored responses at
only $r_{\text{keyed}} =
\valNliRKeyed$. The second practice is to report agreement on several
metrics, namely $\bar{\phi}$ for loading patterns, NMI or ARI for
groupings of items, and $r$ for item pairs, because they can rank the encodings differently.

\subsubsection{Limits of the Random-Embedding Baselines}

Both the adapted parallel analysis and the fit calibration compare the
real similarity matrix with matrices built from random unit vectors. In
thousands of dimensions such vectors are nearly orthogonal, which means
that the cosine between any two of them is close to 0. Their similarity matrix is therefore close to an identity matrix, and
the parallel-analysis thresholds derived from it lie near one. For the
8B embeddings the first threshold was \valPaThrLarge, and the thresholds
declined across eigenvalue positions to \valPaThrEighthLarge{} at the
eighth. These random vectors are not centered, whereas the observed
matrix of the default encoding is. Real item embeddings, by contrast,
are positively similar to one another in the worked example, as
hidden-state representations from contextual language models commonly
are \citep{ethayarajh-2019}, so
their similarity
matrix has one very large eigenvalue that reflects this shared
similarity. Exceeding the random-embedding baseline, built from nearly
unrelated vectors, is a weak test for such a matrix. In practice the adapted parallel analysis behaves like a Kaiser rule
with a cutoff a little above one that declines slowly across eigenvalue
positions, which makes it somewhat stricter than the Kaiser rule here
(first thresholds \valPaThrSmall, \valPaThrMedium, and \valPaThrLarge{}
for the three model sizes). Its verdict therefore shows which
eigenvalues exceed that cutoff rather than where interpretable structure
ends.

A better choice of random vectors would not remove the problem.
Conventional retention baselines model what sampling respondents does to
a correlation matrix when no factors are present. An embedding
similarity matrix, by contrast, is fully determined by the items and the
embedding model, and nothing was sampled to produce it. Any baseline for
such a matrix must therefore stipulate what varies from one draw to the
next, a decision that a sampling design would otherwise make. The
baseline of \citet{pokropek-2026}, built from words drawn at random from
a text corpus, keeps the general thematic similarity that real items
share, which random unit vectors lack. Two other alternatives both indicate too many factors, for different
reasons. Parallel analysis with simulated-data comparison matrices
indicated too many factors on embedding matrices in the benchmark of
\citet{garrido-etal-2025}. With comparison
data generated from factor models, the reproduction of the observed
eigenvalues improves with every added factor. Presumably, a factor model with one unique variance per item cannot
reproduce the many small eigenvalues of an embedding similarity matrix,
which decline gradually, so each added comparison factor captures more
of them (see Comparison-Data Misfit
Profiles below). The
same procedure applied to the human responses shows a sharp elbow at
exactly five factors. The eigenvalues of embedding similarity matrices
therefore appear to decline with no sharp boundary between major and
minor factors. If so, a single number of factors for such a matrix
records how much minor structure the analyst chose to keep, not a
boundary in the data.

A retention test suited to embedding similarity matrices would need
reference data generated from item embeddings themselves. These would be
sets of items whose factor structure is known in advance and whose
embeddings have the geometry of real ones. Such a benchmark could also
train and validate a machine-learning criterion in the spirit of the
comparison data forest \citep{goretzko-ruscio-2024}. Until such a
benchmark exists, we recommend reporting retention on embedding
similarity matrices as the consensus of several criteria together with
the comparison-data misfit profile, and checking agreement with theory
across several candidate numbers of factors.

\subsection{Full Loading Matrices}

Tables~\ref{tab:s-semantic-loadings} and~\ref{tab:s-human-loadings}
report the complete oblimin-rotated loading matrices of the two
five-factor solutions compared in the main text. The semantic solution
factors the mean-centered Pearson similarity matrix of the
Qwen3-Embedding-8B item embeddings. The human solution factors the
correlations among the reverse-scored responses of the \valNHuman{}
respondents who remained after exclusions,
with the same minimum residual extraction and oblimin rotation. Each
factor column carries the label that the psych package assigns
automatically, which is MR, for minimum residual, followed by the
factor's number in order of extraction. The main text matches these
labels to the Big Five domains. The script
\texttt{reproduce/make\_paper\_assets.R} generates both tables directly
from the files \texttt{output/semantic\_loadings.csv} and
\texttt{output/human\_loadings.csv} in the reproduction archive.

{\renewcommand{\baselinestretch}{1}\footnotesize
\begin{longtable}{llrrrrr}
\caption{Factor loadings of the 50 International Personality Item Pool Big-Five Factor Marker items in the five-factor semantic solution (mean-centered Pearson encoding of the Qwen3-Embedding-8B embeddings, minimum residual extraction, oblimin rotation)}
\label{tab:s-semantic-loadings}\\
\toprule
Item & Domain & MR3 & MR1 & MR2 & MR5 & MR4 \\
\midrule
\endfirsthead
\multicolumn{7}{l}{Table~\ref{tab:s-semantic-loadings} (continued)}\\
\toprule
Item & Domain & MR3 & MR1 & MR2 & MR5 & MR4 \\
\midrule
\endhead
\bottomrule
\endfoot
\bottomrule
\multicolumn{7}{p{0.9\linewidth}}{\textit{Note}. Entries are oblimin pattern loadings, and the factors are allowed to correlate. Bold marks each item's largest absolute loading. Item codes give the item's domain initial and its position in the 50-item inventory, and Domain is the domain the inventory assigns it to. MR = minimum residual, the prefix of the psych package's automatic factor labels. Each factor is matched to the domain on whose items its mean absolute loading is largest (columns in table order): MR3 = Neuroticism, MR1 = Conscientiousness, MR2 = Agreeableness, MR5 = Extraversion, MR4 = Openness.}\\
\endlastfoot
E1 & Extraversion & .093 & .308 & .076 & \textbf{.354} & .148 \\
E2 & Extraversion & .051 & .196 & \textbf{.449} & .203 & .122 \\
E3 & Extraversion & .084 & .061 & .127 & \textbf{.752} & .013 \\
E4 & Extraversion & $-$.001 & \textbf{.546} & .206 & .161 & .008 \\
E5 & Extraversion & .037 & .205 & .117 & \textbf{.385} & .257 \\
E6 & Extraversion & .093 & .175 & \textbf{.432} & .065 & .190 \\
E7 & Extraversion & .005 & .110 & .120 & \textbf{.509} & .182 \\
E8 & Extraversion & .062 & .220 & \textbf{.570} & .169 & $-$.044 \\
E9 & Extraversion & $-$.001 & .232 & .225 & \textbf{.450} & .066 \\
E10 & Extraversion & .094 & .034 & \textbf{.408} & .390 & .014 \\
N11 & Neuroticism & \textbf{.817} & .073 & $-$.001 & .013 & .023 \\
N12 & Neuroticism & .176 & \textbf{.395} & .079 & .347 & $-$.093 \\
N13 & Neuroticism & \textbf{.490} & .177 & .154 & .022 & .157 \\
N14 & Neuroticism & .297 & .170 & \textbf{.319} & .141 & .022 \\
N15 & Neuroticism & \textbf{.801} & .074 & .069 & .054 & $-$.042 \\
N16 & Neuroticism & \textbf{1.003} & $-$.020 & .010 & $-$.053 & $-$.029 \\
N17 & Neuroticism & \textbf{.790} & .034 & $-$.009 & .058 & .053 \\
N18 & Neuroticism & \textbf{.860} & .027 & $-$.087 & .019 & .029 \\
N19 & Neuroticism & \textbf{.871} & .057 & .055 & $-$.071 & $-$.012 \\
N20 & Neuroticism & \textbf{.574} & .086 & .104 & .096 & .082 \\
A21 & Agreeableness & .149 & .039 & \textbf{.718} & .128 & $-$.043 \\
A22 & Agreeableness & .041 & .030 & .230 & \textbf{.485} & .261 \\
A23 & Agreeableness & \textbf{.267} & .193 & .146 & .211 & .113 \\
A24 & Agreeableness & \textbf{.453} & $-$.082 & .069 & .425 & .141 \\
A25 & Agreeableness & .084 & .000 & \textbf{.786} & .025 & $-$.031 \\
A26 & Agreeableness & \textbf{.390} & .106 & .043 & .303 & .100 \\
A27 & Agreeableness & $-$.023 & .030 & \textbf{.895} & .072 & $-$.029 \\
A28 & Agreeableness & .077 & \textbf{.459} & .171 & .308 & $-$.026 \\
A29 & Agreeableness & \textbf{.553} & $-$.158 & .073 & .390 & .119 \\
A30 & Agreeableness & .179 & .203 & .046 & \textbf{.596} & $-$.001 \\
C31 & Conscientiousness & .050 & \textbf{.626} & $-$.056 & .159 & .139 \\
C32 & Conscientiousness & .091 & \textbf{.617} & .144 & .070 & $-$.029 \\
C33 & Conscientiousness & .064 & \textbf{.566} & .041 & .091 & .233 \\
C34 & Conscientiousness & .354 & \textbf{.503} & .081 & $-$.046 & .070 \\
C35 & Conscientiousness & .103 & \textbf{.716} & .004 & $-$.001 & .050 \\
C36 & Conscientiousness & .207 & \textbf{.566} & .125 & $-$.103 & .022 \\
C37 & Conscientiousness & .082 & \textbf{.567} & .048 & .121 & .103 \\
C38 & Conscientiousness & .085 & \textbf{.567} & .336 & $-$.094 & $-$.048 \\
C39 & Conscientiousness & .082 & \textbf{.677} & $-$.034 & .093 & .066 \\
C40 & Conscientiousness & .074 & \textbf{.605} & .046 & $-$.032 & .207 \\
O41 & Openness & .063 & .263 & $-$.034 & .222 & \textbf{.450} \\
O42 & Openness & .244 & $-$.122 & .350 & $-$.095 & \textbf{.563} \\
O43 & Openness & .181 & .155 & $-$.027 & .169 & \textbf{.554} \\
O44 & Openness & $-$.045 & .071 & \textbf{.615} & $-$.128 & .403 \\
O45 & Openness & .012 & .300 & $-$.036 & .126 & \textbf{.582} \\
O46 & Openness & .155 & $-$.003 & \textbf{.438} & $-$.093 & .437 \\
O47 & Openness & .220 & .205 & $-$.009 & .217 & \textbf{.332} \\
O48 & Openness & .162 & .235 & .070 & .086 & \textbf{.425} \\
O49 & Openness & .164 & \textbf{.351} & .100 & .098 & .250 \\
O50 & Openness & .069 & .249 & $-$.024 & .150 & \textbf{.606} \\
\end{longtable}}

{\renewcommand{\baselinestretch}{1}\footnotesize
\begin{longtable}{llrrrrr}
\caption{Factor loadings of the 50 International Personality Item Pool Big-Five Factor Marker items in the five-factor solution of the human responses (correlations of reverse-scored responses, minimum residual extraction, oblimin rotation, $N = 874{,}434$ respondents after exclusions)}
\label{tab:s-human-loadings}\\
\toprule
Item & Domain & MR4 & MR1 & MR3 & MR2 & MR5 \\
\midrule
\endfirsthead
\multicolumn{7}{l}{Table~\ref{tab:s-human-loadings} (continued)}\\
\toprule
Item & Domain & MR4 & MR1 & MR3 & MR2 & MR5 \\
\midrule
\endhead
\bottomrule
\endfoot
\bottomrule
\multicolumn{7}{p{0.9\linewidth}}{\textit{Note}. Entries are oblimin pattern loadings, and the factors are allowed to correlate. Bold marks each item's largest absolute loading. Item codes give the item's domain initial and its position in the 50-item inventory, and Domain is the domain the inventory assigns it to. MR = minimum residual, the prefix of the psych package's automatic factor labels. Each factor is matched to the domain on whose items its mean absolute loading is largest (columns in table order): MR4 = Extraversion, MR1 = Neuroticism, MR3 = Agreeableness, MR2 = Conscientiousness, MR5 = Openness.}\\
\endlastfoot
E1 & Extraversion & \textbf{.712} & .042 & $-$.010 & $-$.015 & $-$.012 \\
E2 & Extraversion & \textbf{.713} & .086 & .053 & $-$.026 & .008 \\
E3 & Extraversion & \textbf{.623} & $-$.189 & .179 & .061 & $-$.068 \\
E4 & Extraversion & \textbf{.755} & $-$.023 & $-$.039 & .018 & $-$.027 \\
E5 & Extraversion & \textbf{.714} & .012 & .123 & .067 & .034 \\
E6 & Extraversion & \textbf{.522} & .005 & .073 & .010 & .235 \\
E7 & Extraversion & \textbf{.721} & $-$.004 & .069 & .012 & $-$.019 \\
E8 & Extraversion & \textbf{.618} & .039 & $-$.109 & $-$.070 & .026 \\
E9 & Extraversion & \textbf{.653} & .027 & $-$.105 & $-$.050 & .103 \\
E10 & Extraversion & \textbf{.684} & $-$.079 & $-$.018 & .014 & $-$.019 \\
N11 & Neuroticism & $-$.074 & \textbf{.708} & .125 & .061 & $-$.055 \\
N12 & Neuroticism & $-$.064 & \textbf{.559} & .013 & .101 & $-$.004 \\
N13 & Neuroticism & $-$.125 & \textbf{.611} & .208 & .092 & .017 \\
N14 & Neuroticism & $-$.129 & \textbf{.349} & .059 & $-$.069 & .092 \\
N15 & Neuroticism & .005 & \textbf{.505} & .004 & $-$.029 & $-$.095 \\
N16 & Neuroticism & .027 & \textbf{.746} & .031 & .005 & $-$.068 \\
N17 & Neuroticism & .077 & \textbf{.719} & $-$.021 & $-$.091 & .008 \\
N18 & Neuroticism & .063 & \textbf{.741} & $-$.035 & $-$.096 & .005 \\
N19 & Neuroticism & .055 & \textbf{.719} & $-$.175 & .041 & $-$.017 \\
N20 & Neuroticism & $-$.219 & \textbf{.574} & .017 & $-$.135 & .114 \\
A21 & Agreeableness & $-$.049 & $-$.056 & \textbf{.499} & $-$.006 & .061 \\
A22 & Agreeableness & .282 & $-$.041 & \textbf{.522} & $-$.040 & .060 \\
A23 & Agreeableness & $-$.180 & $-$.240 & \textbf{.418} & .144 & $-$.071 \\
A24 & Agreeableness & $-$.055 & .033 & \textbf{.807} & $-$.007 & $-$.004 \\
A25 & Agreeableness & .060 & $-$.027 & \textbf{.661} & $-$.038 & $-$.002 \\
A26 & Agreeableness & $-$.063 & .131 & \textbf{.618} & .001 & $-$.074 \\
A27 & Agreeableness & .225 & $-$.084 & \textbf{.613} & $-$.035 & .015 \\
A28 & Agreeableness & .077 & $-$.014 & \textbf{.559} & .061 & .001 \\
A29 & Agreeableness & .019 & .100 & \textbf{.700} & .019 & .033 \\
A30 & Agreeableness & .260 & $-$.100 & \textbf{.362} & .086 & .062 \\
C31 & Conscientiousness & .015 & .000 & $-$.032 & \textbf{.648} & .091 \\
C32 & Conscientiousness & $-$.051 & $-$.035 & $-$.069 & \textbf{.573} & $-$.105 \\
C33 & Conscientiousness & $-$.068 & .075 & .062 & \textbf{.405} & .251 \\
C34 & Conscientiousness & .012 & $-$.279 & $-$.010 & \textbf{.556} & $-$.003 \\
C35 & Conscientiousness & .073 & .009 & $-$.002 & \textbf{.632} & $-$.079 \\
C36 & Conscientiousness & $-$.024 & $-$.088 & $-$.037 & \textbf{.614} & $-$.043 \\
C37 & Conscientiousness & $-$.049 & .160 & $-$.004 & \textbf{.581} & .040 \\
C38 & Conscientiousness & .006 & $-$.152 & .097 & \textbf{.465} & .049 \\
C39 & Conscientiousness & .053 & .111 & .058 & \textbf{.634} & $-$.054 \\
C40 & Conscientiousness & .002 & .052 & .009 & \textbf{.456} & .249 \\
O41 & Openness & $-$.015 & $-$.011 & $-$.058 & .035 & \textbf{.596} \\
O42 & Openness & $-$.068 & $-$.189 & .014 & $-$.030 & \textbf{.579} \\
O43 & Openness & $-$.014 & .107 & .091 & $-$.099 & \textbf{.545} \\
O44 & Openness & $-$.095 & $-$.109 & .114 & $-$.100 & \textbf{.518} \\
O45 & Openness & .166 & $-$.027 & $-$.069 & .122 & \textbf{.575} \\
O46 & Openness & .011 & $-$.032 & .091 & $-$.052 & \textbf{.499} \\
O47 & Openness & .019 & $-$.115 & $-$.044 & .154 & \textbf{.474} \\
O48 & Openness & $-$.006 & .076 & $-$.126 & $-$.024 & \textbf{.564} \\
O49 & Openness & $-$.185 & .115 & .187 & .037 & \textbf{.396} \\
O50 & Openness & .122 & .017 & .008 & .000 & \textbf{.662} \\
\end{longtable}}

\subsection{Signed NLI Similarity Matrix}

The function \texttt{sfa\_nli\_matrix()} scores ordered pairs of items with the
cross-encoder \texttt{cross-encoder/nli-deberta-v3-base}, a DeBERTa-v3
model \citep{he-etal-2023} trained on natural language inference data
such as SNLI \citep{bowman-etal-2015} and MultiNLI
\citep{williams-etal-2018}. Each pair's score is the
probability of entailment minus the probability of contradiction, a
signed mapping in the spirit of \citet{hommel-arslan-2025}, averaged over the two orders of the pair (the function's default).
Before averaging, the scores of the two orders of each pair correlated
\valNliOrdersR. The
three class probabilities come from the model's softmax output, and the
probability of the neutral class is not used. The excerpt below compares three Extraversion items
under this matrix and under \texttt{atomic\_reversed}, whose matrix the
main analysis stored as \texttt{sim\_ar}. E1 (``I am the life of the
party.'') and E5 (``I start conversations.'') are forward-keyed, and E2
(``I don't talk a lot.'') is reverse-keyed:

{\scriptsize
\begin{verbatim}
> sim_nli <- sfa_nli_matrix(big5$items)   # cross-encoder/nli-deberta-v3-base
> round(sim_nli[c("E1", "E2", "E5"), c("E1", "E2", "E5")], 2)
      E1    E2    E5
E1  1.00 -1.00  0.03
E2 -1.00  1.00 -0.98
E5  0.03 -0.98  1.00
> round(sim_ar[c("E1", "E2", "E5"), c("E1", "E2", "E5")], 2)
      E1    E2    E5
E1  1.00 -0.63  0.67
E2 -0.63  1.00 -0.65
E5  0.67 -0.65  1.00
\end{verbatim}
}

\begin{figure}[tbp]
\centering
\captionsetup{justification=raggedright,singlelinecheck=false}
\includegraphics[width=0.6\textwidth]{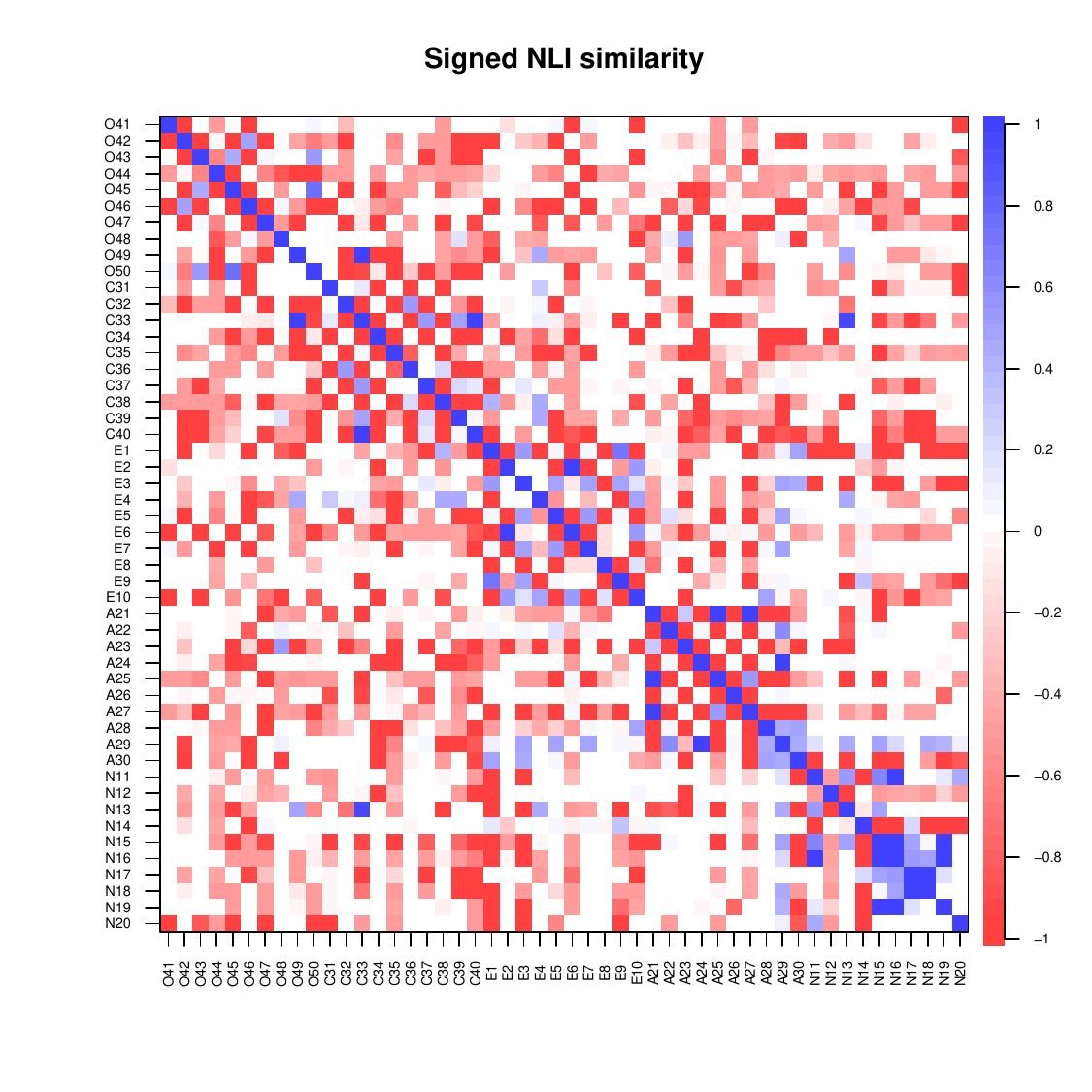}
\caption{Signed natural language inference (NLI) similarity, the
probability of entailment minus the probability of contradiction, among
the 50 International Personality Item Pool Big-Five Factor Marker items,
grouped by domain. The color scale runs from $-1$ (red) through 0
(white) to 1 (blue), and the diagonal is 1. Red cells are negative and
mark pairs that the cross-encoder judged to contradict each other. The
matrix is shown before the repair that made it positive definite for
factoring. Every similarity in the cosine matrix of
Figure~\ref{fig:corplot} in the main text is positive. Here, by contrast, pairs of items of the same domain averaged
\valNliSameKey{} when keyed the same way and \valNliOppKey{} when keyed
oppositely, which produces negative bands within the domain blocks.
Many pairs from different domains are negative as well
(\valNliCrossNegPct\% of them, and \valNliCrossStrongPct\% below $-.5$). Item codes give the domain (O, C, E, A, N) and the item's position
in the inventory.}
\label{fig:corplot-nli}
\end{figure}

The NLI matrix scores E1 and its reverse-keyed counterpart E2 at
$\valNliOpp$, which means that the cross-encoder judged the two
statements to contradict each other almost certainly. Under
\texttt{atomic\_reversed} the same pair scores $\valArOpp$. Flipping the sign of E2's embedding makes the pair's similarity
negative, but the pair is not the most dissimilar one in the
sign-flipped matrix, whose lowest value is $\valRangeArMin$, although
the two statements say opposite things. Figure~\ref{fig:corplot-nli}
shows the full NLI matrix.

The Encoding Comparison in the Technical Reference compares this matrix
with the four embedding encodings (Table~\ref{tab:s-encodings}).

\subsection{Comparison-Data Misfit Profiles}

The comparison-data method of \citet{ruscio-roche-2012} treats retention
as a comparison between models rather than as a test against a baseline
with no structure. For each candidate number of factors $k$, it asks how
well data generated to have exactly $k$ factors reproduce the observed
eigenvalues. The method has four steps:

\begin{enumerate}
\item For each $k$, generate a large population of comparison data whose
correlations follow the $k$-factor model fitted to the observed
correlations and whose variables have the observed distributions.
\item Draw random samples from that population, each the same size as
the observed data.
\item For each sample, compute the root mean square difference between
its eigenvalues and the observed eigenvalues. This difference is the
misfit of $k$ factors, and it is small when $k$ factors account for the
observed eigenvalues well.
\item Compare the misfits of successive values of $k$.
\end{enumerate}

\noindent Because each candidate number of factors is judged against the
one before it, the method does not need a baseline with no structure,
which is the weak point of the adapted parallel analysis (see Limits of
the Random-Embedding Baselines above). The comparison populations are generated with an adaptation of the GenData algorithm of
\citet{ruscio-kaczetow-2008}. GenData draws each variable from its target
distribution and iteratively adjusts an intermediate correlation matrix
so that the generated data reproduce the target correlations as closely
as possible. It stops when several adjustments in a row bring no
improvement. The adaptation in \texttt{sfa\_cd()} instead makes a fixed
sequence of six adjustments and keeps the population that reproduced the
target correlations best.

The function \texttt{sfa\_cd()} applies this method to embedding similarity matrices
by treating the embedding dimensions as the cases, the role that
respondents play in response data. This is a heuristic, because the
dimensions are not sampled like respondents, and the resulting numbers of cases (1,024 to 4,096) exceed the range of
sample sizes, 200 to 1,000, for which the method was validated
\citep{ruscio-roche-2012}. The Pearson correlations between
items computed across the embedding dimensions equal the
\texttt{mean\_centered\_pearson} similarities, so the profile describes
exactly the matrix that the main analyses factor. The profiles reported here use candidate numbers of factors from 1 to
10, 500 random samples for each number, comparison populations of
10{,}000 cases generated with six refinement iterations, minimum residual
factoring, the seed 42 for the random steps inside \texttt{sfa\_cd()},
and \texttt{semanticfa} 0.5.1. They were
computed for the three Qwen3-Embedding model sizes (0.6B, 4B, and 8B)
and for a random subsample of $n = \valCdHumanN$ respondents'
reverse-scored responses from the human benchmark, drawn after
\texttt{set.seed(42)}.

\begin{figure}[tbp]
\centering
\includegraphics[width=0.95\textwidth]{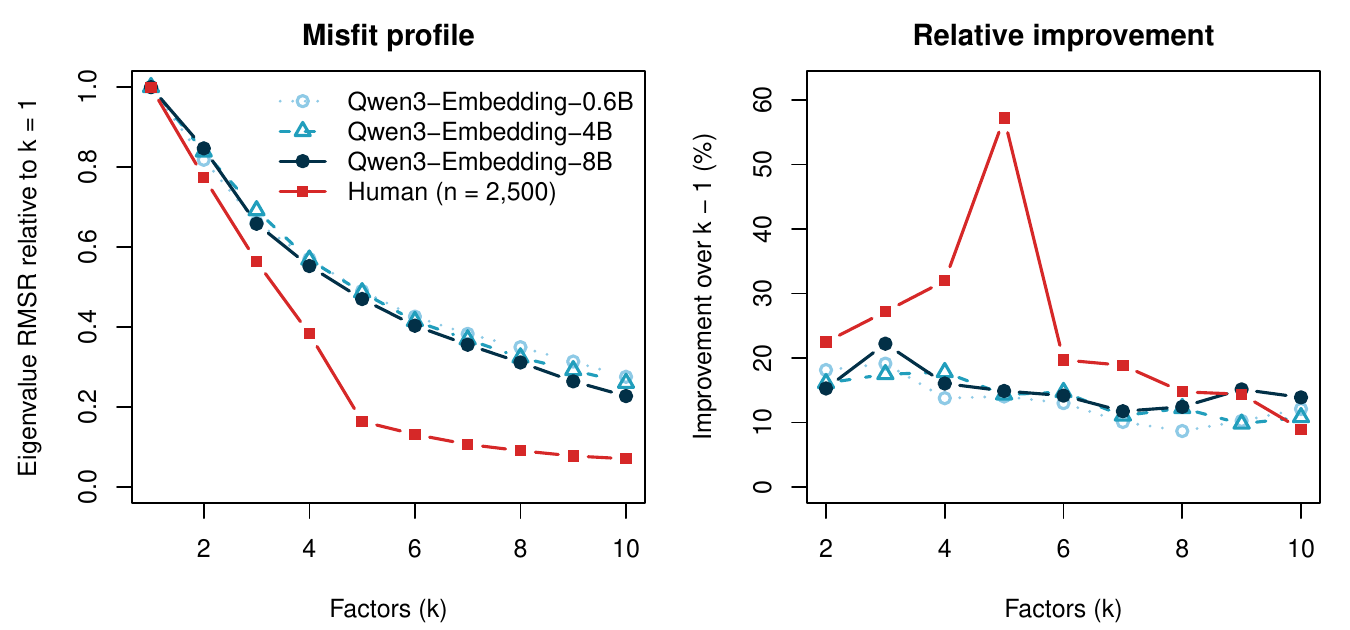}
\captionsetup{justification=raggedright,singlelinecheck=false}
\caption{Comparison-data misfit profiles for the similarity matrices of
three embedding models (Qwen3-Embedding-0.6B, -4B, and -8B) and for one
random subsample of $n = \valCdHumanN$ respondents' reverse-scored
responses, drawn with the seed 42.
The misfit at each candidate number of factors is the median, across 500
random samples from the comparison population, of the eigenvalue RMSR,
the root mean square difference between a sample's eigenvalues and the
observed eigenvalues. The eigenvalue RMSR is a different quantity from
the root mean square residual (RMSR) that \texttt{sfa()} reports as a
fit diagnostic. The embedding profiles treat the embedding dimensions
(1,024 to 4,096) as cases, and the human profile treats respondents as
cases. The three model sizes are drawn with different line types (0.6B
dotted, 4B dashed, 8B solid), and the human profile is drawn in red. Left:
misfit divided by its value at one factor.
Right: relative improvement, the percentage reduction in misfit from the
preceding number of factors. The human profile shows a sharp elbow at
five factors, where the improvement peaks at \valCdHumanImpFive\% and
then drops. The embedding profiles decline gradually. Their largest
improvement, \valCdEmbImpHigh\%, is less than half the human improvement
into five factors.}
\label{fig:cd-profiles}
\end{figure}

Figure~\ref{fig:cd-profiles} shows the profiles, and two quantities
summarize each one. The normalized misfit at $k$ is the median misfit at
$k$ divided by the median misfit at one factor, so it shows how much of
the one-factor misfit remains. The relative improvement at $k$ is the
percentage by which the median misfit falls from $k - 1$ to $k$ factors.
An elbow is a value of $k$ at which the relative improvement is large
and after which it drops sharply, and it marks the last factor that
substantially improves the fit.

For the human responses, the relative improvement rose with each added
factor up to the fifth, which reduced the misfit by
\valCdHumanImpFive\% relative to four factors, and then fell to
\valCdHumanImpSix\% at the sixth. At five factors only
\valCdHumanNormFive{} of the one-factor misfit remained. The eigenvalues
of the Big Five responses therefore show a sharp boundary after the
fifth factor, which is where the theory places it. This profile comes
from one random subsample of \valCdHumanN{} respondents, although the
elbow is large enough that another draw is unlikely to remove it. The three embedding
matrices showed no such elbow. Each added factor reduced their misfit by between \valCdEmbImpLow\% and
\valCdEmbImpHigh\%, with no spike comparable to the human one at any
number of factors (the largest embedding improvement was less than half
the human peak of \valCdHumanImpFive\%), and \valCdEmbNormFiveRange{} of the one-factor misfit
remained at five factors. Presumably, a factor model with one unique
variance per item cannot reproduce the many small eigenvalues of an embedding
similarity matrix, which decline gradually, so each added comparison
factor keeps improving the reproduction.

\citet{ruscio-roche-2012} also give a sequential stopping rule. It tests each $k$ against $k - 1$ with a one-tailed Mann--Whitney test
on the two sets of misfits, asking whether $k$ factors reduce the
misfit, and it keeps adding factors until the next one does not
reduce the misfit significantly. We ran it at the liberal significance level that they
used, $\alpha = .30$, which performed best in their preliminary tests. For all three embedding matrices it reached
\valCdEmbVerdictWord{} factors, the largest number tested.

The human responses show how the rule behaves on response data. Besides
the subsample of \valCdHumanN{} respondents behind the profile, we drew
subsamples of 250, 500, and 1,000 respondents. All four were drawn after
the same seed, 42, so each smaller subsample is the first part of the
larger ones, and the four are nested. Five further subsamples of 250 and
of 500 respondents were drawn with the seeds 1 to 5, and with each of
these seeds, too, the subsample of 250 is the first part of the
subsample of 500. These seeds governed only the choice of respondents,
because the random steps inside \texttt{sfa\_cd()} used the seed 42 in
every run. In the subsample of 250 drawn with the seed 42, near the lower
end of the sample sizes for which the method was validated, the rule
stopped at \valCdCrossTwoFiftyWord{} factors, and the five further
subsamples of 250 stopped at \valCdRepLowMin{} to \valCdRepLowMax. In
the subsamples drawn with the seed 42, the rule reached
\valCdCrossSatWord, the largest number tested, from 500 respondents
upward, although the five further subsamples of 500 stopped at
\valCdRepMidMin{} to \valCdRepMidMax. The sequential rule therefore did
not stop at five on the human data at any size tested. This is
consistent with the \valHumanPaKWord{} factors that conventional
parallel analysis indicated for the full benchmark of \valNHuman{}
respondents (see Retention Criteria above), although the cap prevents a
direct comparison.

These results are consistent with an account based on statistical
power. On this account, the test detects smaller reductions in misfit
as the number of cases grows, so with enough respondents the rule keeps adding factors
beyond the five interpretable ones. The embedding matrices, with 1,024
to 4,096 dimensions as cases, may reach the cap for the same reason
rather than because of a property peculiar to embeddings. The results do
not establish this account. The test's $p$ values also depend on the 500
simulated samples for each number of factors, and the embedding
dimensions only play the role of cases, so the comparison with the human
subsamples is an analogy. On the embedding matrices, moreover, the
profiles show no boundary for the rule to detect. Taken together, the profiles support the
practice recommended in the main text, which is to report retention as
the consensus of several criteria and to check agreement with theory
across several candidate numbers of factors.

\subsection{Model Size With a Second Scale}

In the main text, the separation between the Big Five domains did not
rise steadily from the 0.6B to the 8B model. Because that comparison
used a single inventory, it was repeated with a second one, the
Depression Anxiety Stress Scales (DASS). The DASS has 42 items, 14 for
each of three scales, and each item describes a negative emotional
symptom, such as ``I found myself getting upset by quite trivial
things'' \citep{lovibond-lovibond-1995}. The scale's history makes it a useful second case. Its Stress
scale grew out of items that did not separate anxiety from depression,
and in its authors' factor analysis of responses, the Anxiety and Stress
factors merged into one when only two factors were extracted
\citep{lovibond-lovibond-1995}.

The 42 items were embedded with the same three models and the same
settings as the Big Five items. Re-embedding them with the model
revisions given in the Open Practices Statement reproduced the stored
vectors, with a cosine of \valDassEmbCheckCos{} to six decimal places
between each new vector and the stored one. The comparison then used the
statistics of the main text. The first is the point-biserial correlation
between sharing a scale and a pair's similarity, over all
\valDassNPairs{} item pairs, with an item-level bootstrap 95\% interval.
The second is the gap between the mean similarity of pairs within a
scale and that of pairs between scales. Because the same resampled items
served all three sizes, the bootstrap also gave an interval for the
difference between two sizes. Finally, three factors were extracted at
each size, the theoretical number, and each item was assigned to the
factor on which it loaded most strongly. The agreement of this
assignment with the three scales was measured with the normalized mutual
information (NMI) and the adjusted Rand index (ARI), as in the main text,
and compared with a permutation baseline, the largest values reached
when the item assignment of the 8B solution was compared with 1,000
random permutations of the scale labels. For comparison,
Table~\ref{tab:s-modelsize} also gives the same agreement for the
five-factor Big Five solution at each size.

\begin{figure}[tbp]
\centering
\includegraphics[width=0.32\textwidth]{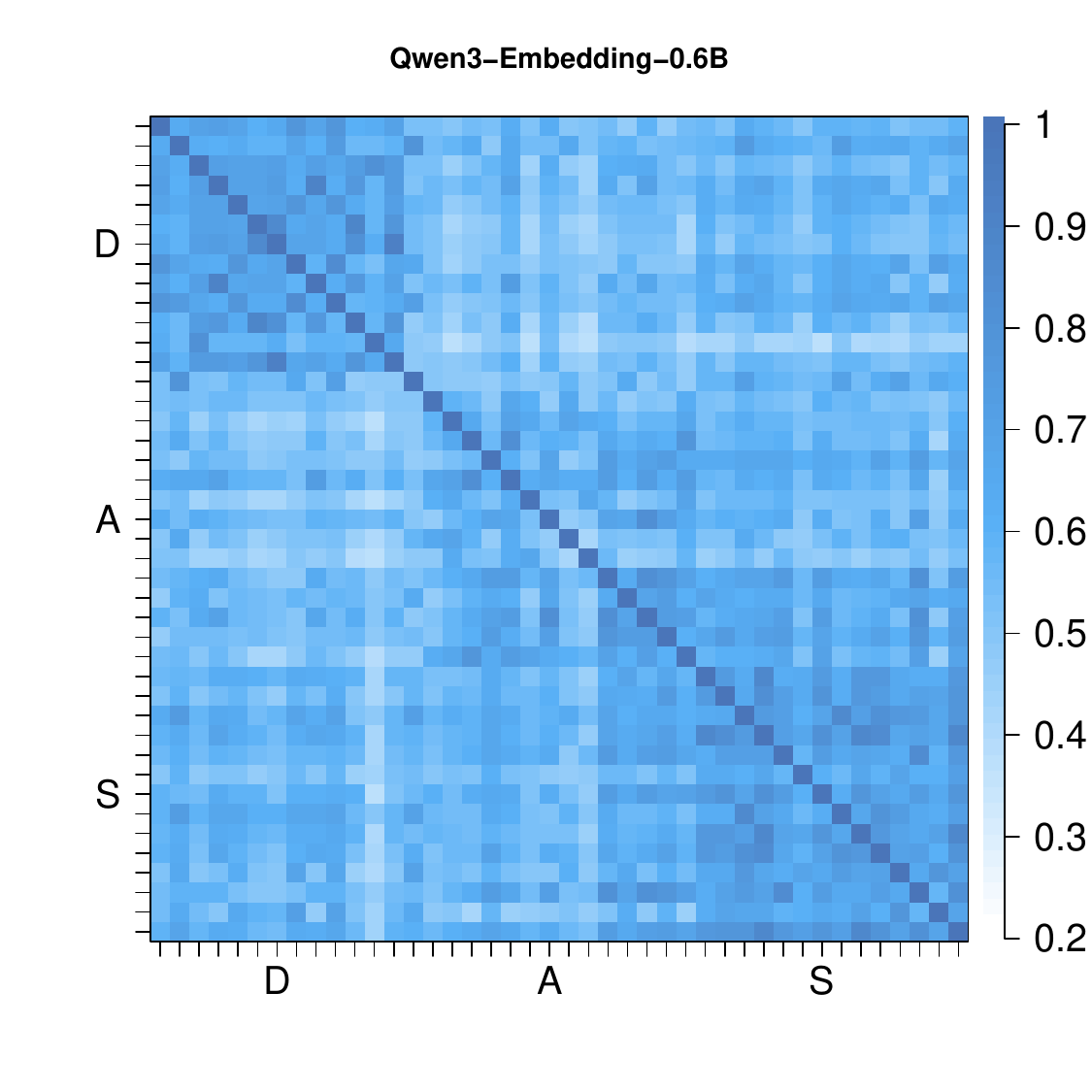}\hfill
\includegraphics[width=0.32\textwidth]{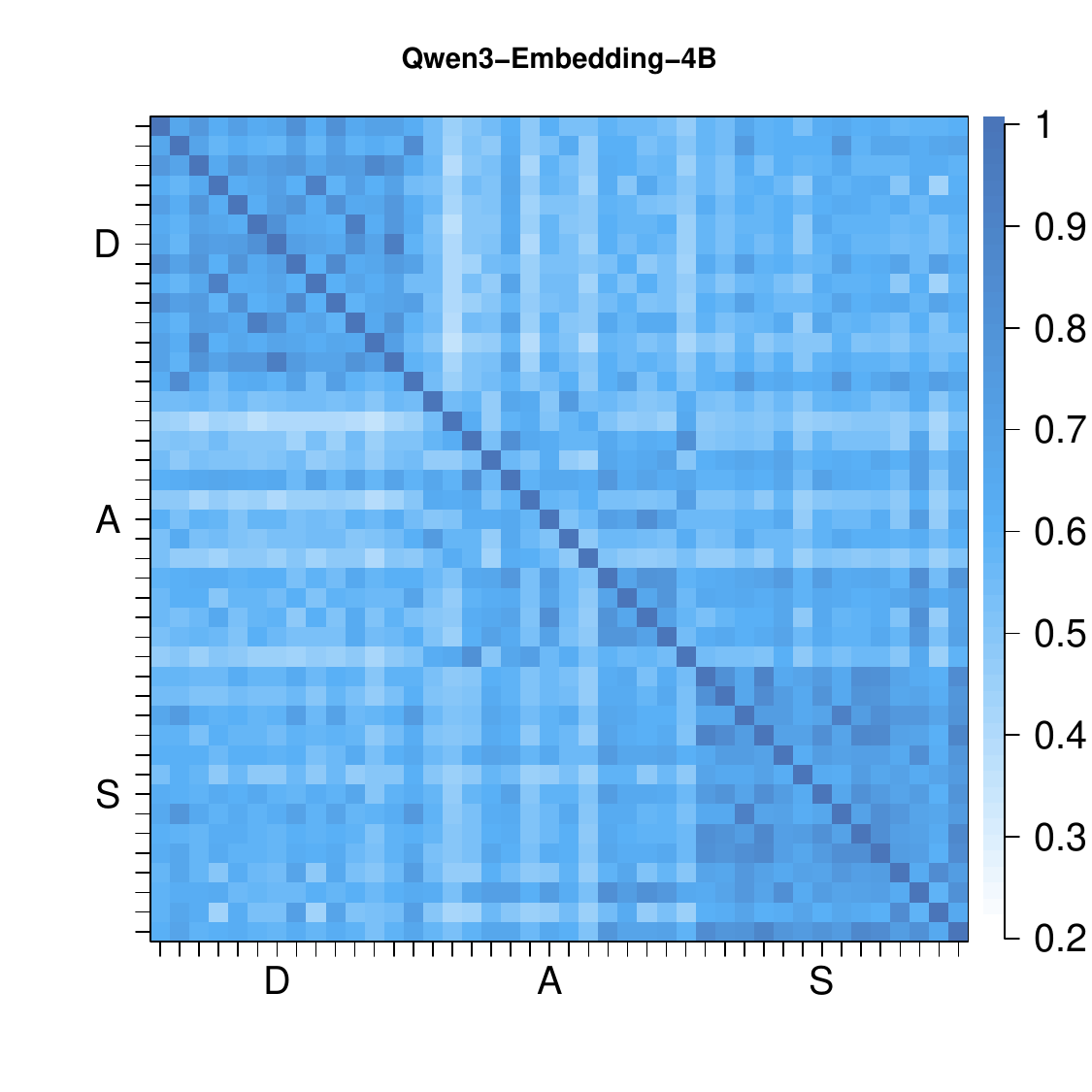}\hfill
\includegraphics[width=0.32\textwidth]{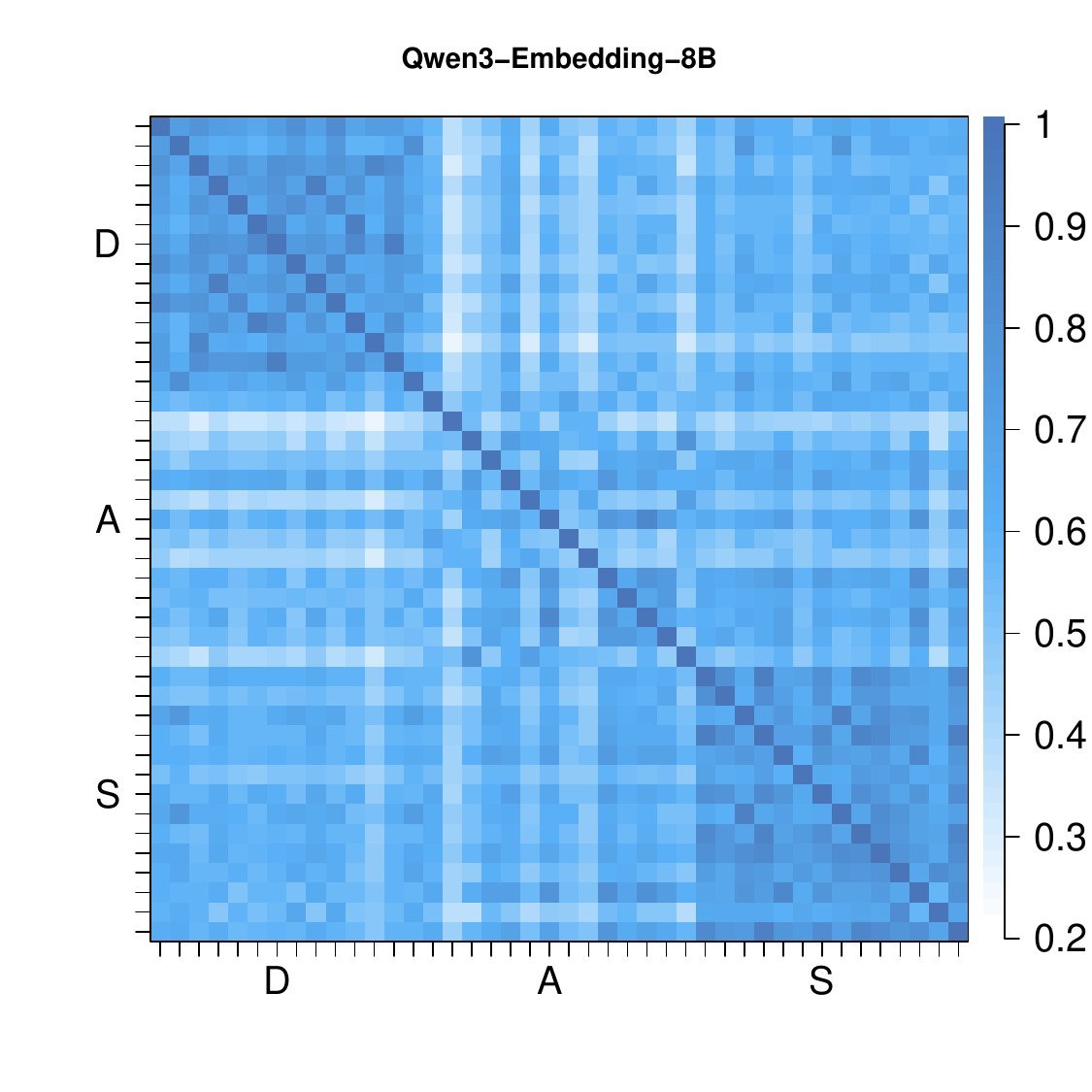}
\captionsetup{justification=raggedright,singlelinecheck=false}
\caption{Mean-centered Pearson similarity matrices of the 42 items of
the Depression Anxiety Stress Scales (DASS), grouped by scale in the
order Depression, Anxiety, and Stress, for Qwen3-Embedding-0.6B (left),
4B (center), and 8B (right). The axes mark each scale's block with its
initial letter (D, A, S) at the middle of the block. The three panels
share one color scale of mean-centered Pearson similarity, from white
at \valDassFloor, the lowest similarity in any of the three matrices
rounded down to the nearest tenth, to dark blue at 1. The diagonal is 1
by definition. The Depression and Stress blocks stand out more with each
larger model, and the Anxiety block less.}
\label{fig:s-dass}
\end{figure}

With the DASS, the gap rose with model size, from \valDassGapSmall{} for
0.6B to \valDassGapMedium{} for 4B and \valDassGapLarge{} for 8B
(Table~\ref{tab:s-modelsize}). With the Big Five it stayed nearly
constant. The paired 95\% interval for the DASS increase from 0.6B to 8B
ran from $\valDassGapDiffLo$ to \valDassGapDiffHi, so it only just
included zero, whereas the Big Five interval for the same increase ran
from $\valBigGapDiffLo$ to \valBigGapDiffHi. The point-biserial
correlation rose only slightly, from \valDassRpbSmall{} to
\valDassRpbMedium{} and \valDassRpbLarge, and its intervals overlapped
widely. The three scales did not change alike
(Figure~\ref{fig:s-dass}). The gaps of Depression (\valDassDepGapSmall,
\valDassDepGapMedium, and \valDassDepGapLarge) and Stress
(\valDassStrGapSmall, \valDassStrGapMedium, and \valDassStrGapLarge)
grew with size, whereas the gap of Anxiety shrank (\valDassAnxGapSmall,
\valDassAnxGapMedium, and \valDassAnxGapLarge).

The clearest change was in the three-factor solutions. Under the 0.6B
model, the factor with the most Anxiety items (\valDassSmallAnxModal{}
of 14) also took all \valDassSmallStrOnAnx{} Stress items, so
most Anxiety items and the Stress items formed a single factor, as
Anxiety and Stress did in the two-factor solution of responses that
\citet{lovibond-lovibond-1995} reported.
Under the 8B model, each scale had a factor of its own, and these
factors took \valDassLargeDepModal{} Depression, \valDassLargeStrModal{}
Stress, and \valDassLargeAnxModal{} Anxiety items, out of 14 each. Agreement with the
three scales rose accordingly, from NMI~=~\valDassNmiSmall{} and
ARI~=~\valDassAriSmall{} for 0.6B to \valDassNmiMedium{} and
\valDassAriMedium{} for 4B and \valDassNmiLarge{} and
\valDassAriLarge{} for 8B, all above the largest values reached with
permuted labels (\valDassPermNmi{} and \valDassPermAri). With the Big
Five, agreement did not rise with size. It was highest for 4B
(NMI~=~\valBigNmiMedium, ARI~=~\valBigAriMedium), against
\valBigNmiSmall{} and \valBigAriSmall{} for 0.6B and \valBigNmiLarge{}
and \valBigAriLarge{} for 8B. Neither parallel analysis nor the
consensus of the five retention criteria indicated three factors for the
DASS at any size. Parallel analysis retained
\valDassPaKLargeWord{} factors at every size, and the consensus of the
five criteria was \valDassConsSmallWord{} for 0.6B,
\valDassConsMediumWord{} for 4B, and \valDassConsLargeWord{} for 8B.

The 0.6B solution was not the only one. \texttt{psych}
\citep{revelle-2026} starts the oblique rotation from several
orientations, most of them random, and keeps the result with the largest
share of loadings close to zero, which is not necessarily the one with
the lowest value of the rotation criterion. The seed argument of
\texttt{sfa()} does not set these random orientations, so the
reproduction script sets the random seed before each DASS fit. To see how
far the results depend on the orientations drawn, each three-factor DASS
fit and each five-factor Big Five fit was repeated under
\valDassSeedN{} random seeds. Judged by their agreement with the scales,
every fit but one gave the same solution under every seed. The exception was the 0.6B DASS fit, which gave the
solution reported here under \valDassSmallSeedsRep{} seeds and a second
solution under the other \valDassSmallSeedsAlt. In the second solution,
all \valDassAltAnxModal{} Anxiety items and \valDassAltStrOnAnx{} of the
14 Stress items loaded most strongly on one factor, and agreement with the scales
was higher (NMI~=~\valDassAltNmi, ARI~=~\valDassAltAri), although still
lower than for 4B. The reported solution had the lower value of the
oblimin criterion, which the rotation minimizes (\valDassCritSmall{}
against \valDassCritAlt). Two of the refits ended with a warning that the
rotation had not converged, and neither gave a solution other than those
described here. Neither solution separated Anxiety from
Stress.

The DASS thus showed a clearer effect of model size than the Big Five.
The smallest model merged two scales that are also closely related in
responses, and the largest separated all three. The two inventories
differ in many ways besides the effect of size, including the number of
scales and the kind of content their items describe, so two inventories
cannot show which of these properties makes model size matter.

\begin{table}[tbp]
\centering
\captionsetup{justification=raggedright,singlelinecheck=false}
\caption{Separation of the theoretical scales and agreement of the
factor solutions with them, by model size, for the Big Five inventory
and the DASS}
\label{tab:s-modelsize}
\tighttable
{\footnotesize
\begin{tabular}{llrcrrrr}
\toprule
Scale & Model & $r_{\text{pb}}$ & 95\% interval & Gap & NMI & ARI & Consensus \\
\midrule
Big Five & 0.6B & .398 & [.286, .515] & .077 & .497 & .393 & 6 \\
 & 4B & .458 & [.372, .548] & .077 & .693 & .574 & 6 \\
 & 8B & .416 & [.316, .535] & .079 & .527 & .402 & 5 \\
\midrule
DASS & 0.6B & .481 & [.344, .612] & .099 & .412 & .271 & 5 \\
 & 4B & .505 & [.349, .649] & .106 & .700 & .650 & 4 \\
 & 8B & .510 & [.343, .664] & .124 & .834 & .801 & 4 \\
\bottomrule
\end{tabular}

}
\caption*{\textit{Note}. Each row is one inventory embedded with one
size of Qwen3-Embedding, under the mean-centered Pearson encoding. The
point-biserial correlation $r_{\text{pb}}$ is computed between sharing a
scale (for the Big Five, a domain) and a pair's similarity, over all
pairs of items, with its item-level bootstrap 95\% interval. Gap is the
mean similarity of pairs within a scale minus that of pairs between
scales. NMI (normalized mutual information) and ARI (adjusted Rand
index) measure the agreement of each item's dominant factor with its
scale, in the five-factor solution for the Big Five and the
three-factor solution for the DASS. For the 0.6B DASS model, the row
gives the solution with the lower oblimin criterion, found under
\valDassSmallSeedsRep{} of \valDassSeedN{} random seeds, and the text
describes a second solution, found under the other
\valDassSmallSeedsAlt{} (NMI = \valDassAltNmi, ARI = \valDassAltAri).
Consensus is the number
of factors chosen by more of the five retention criteria than any other
number, with ties going to the smaller number. DASS =
Depression Anxiety Stress Scales.}
\end{table}

\end{document}